\UseRawInputEncoding
\documentclass[11pt]{article}
	
	\newcommand{\blind}{0}
	
    \makeatletter
    \renewcommand\section{\@startsection {section}{1}{\z@}%
                                       {-3.5ex \@plus -1ex \@minus -.2ex}%
                                       {2.3ex \@plus.2ex}%
                                       {\normalfont\fontfamily{phv}\fontsize{16}{19}\bfseries}}
    \renewcommand\subsection{\@startsection{subsection}{2}{\z@}%
                                         {-3.25ex\@plus -1ex \@minus -.2ex}%
                                         {1.5ex \@plus .2ex}%
                                         {\normalfont\fontfamily{phv}\fontsize{14}{17}\bfseries}}
    \renewcommand\subsubsection{\@startsection{subsubsection}{3}{\z@}%
                                        {-3.25ex\@plus -1ex \@minus -.2ex}%
                                         {1.5ex \@plus .2ex}%
                                         {\normalfont\normalsize\fontfamily{phv}\fontsize{14}{17}\selectfont}}
    \makeatother
	
	\usepackage{amsmath}
	\usepackage{graphicx}
	\usepackage{enumerate}
	\usepackage{xcolor}
	\usepackage{natbib} 
	\usepackage{url} 
	
	\usepackage{hyperref}
        \usepackage{graphicx}
        \usepackage{tabularx}
        \usepackage{multirow}
        \usepackage{array}
        \usepackage{longtable}
        \usepackage{rotating}
        \usepackage{caption}
        \usepackage{tikz}
        \usepackage{comment}
        \usepackage{url}
        \usepackage{soul}
	
\begin{document}
		
		\def\spacingset#1{\renewcommand{\baselinestretch}%
			{#1}\small\normalsize} \spacingset{1}
		
		\if0\blind
		{
			\title{\bf{Advancing Additive Manufacturing through Deep Learning: A Comprehensive Review of Current Progress and Future Challenges}}
			\author{
   \normalsize Amirul Islam Saimon$^a$, Emmanuel Yangue$^b$, Xiaowei Yue$^c$, \\
   \normalsize Zhenyu (James) Kong$^a$, and Chenang Liu$^b$$^\ast$ (email: chenang.liu@okstate.edu) \\
   \text{ }\\
			\normalsize $^a$Grado Department of Industrial \& Systems Engineering, Virginia Tech, USA\\
            \normalsize $^b$School of Industrial Engineering \& Management, Oklahoma State University, USA\\
            \normalsize $^c$Department of Industrial Engineering, Tsinghua University, China}
			\date{}
			\maketitle
		} \fi
		
		\if1\blind
		{
            \title{\bf{Advancing Additive Manufacturing through Deep Learning: A Comprehensive Review of Current Progress and Future Challenges}}
            \author{\it Author information is purposely removed for double-blind review}
            \date{}
            \maketitle
		} \fi
		\bigskip
	\spacingset{1.00}
	\begin{abstract}
            \normalsize This paper presents the first comprehensive literature review of deep learning (DL) applications in additive manufacturing (AM). It addresses the need for a thorough analysis in this rapidly growing yet scattered field, aiming to bring together existing knowledge and encourage further development. Our research questions cover three major areas of AM: (\textit{i}) design for AM, (\textit{ii}) AM modeling, and (\textit{iii}) monitoring and control in AM. We use a step-by-step approach following the Preferred Reporting Items for Systematic Reviews and Meta-Analyses (PRISMA) guidelines to select papers from Scopus and Web of Science databases, aligning with our research questions. We include only those papers that implement DL across seven major AM categories - binder jetting, directed energy deposition, material extrusion, material jetting, powder bed fusion, sheet lamination, and vat photopolymerization. Our analysis reveals a trend towards using deep generative models, such as generative adversarial networks, for generative design in AM. It also highlights an increasing effort to incorporate process physics into DL models to improve AM process modeling and reduce data requirements. Additionally, there is growing interest in using 3D point cloud data for AM process monitoring, alongside traditional 1D and 2D formats. Finally, this paper summarizes the current challenges and recommends some of the promising opportunities in this domain for further investigation with a special focus on (\textit{i}) generalizing DL models for a wide range of geometry types, (\textit{ii}) managing uncertainties both in AM data and DL models, (\textit{iii}) overcoming limited, imbalanced, and noisy AM data issues by incorporating deep generative models, and (\textit{iv}) unveiling the potential of interpretable DL for AM. \\
	\end{abstract}
			
	\noindent%

	\spacingset{1.5} 

\section{Introduction} \label{s:intro}
The popularity of additive manufacturing (AM) over conventional subtractive manufacturing has been increasing due to the rising demand for customized products \citep{Qin_Hu_Liu__Tang2022_review, Mattera_Nele_Paolella2023}. This popularity can be easily realized from the rapidly expanding trend of the AM market size, from USD 20,670 million in 2023 to USD 98,310 million in 2032, as highlighted in \cite{AM_market_size_reference}. AM not only provides an excellent alternative to conventional manufacturing but also reduces material wastage during the process \citep{Gibson_Rosen_Stucker_2021}. Furthermore, it has the capability of printing parts having internal cavities in the design as well as parts having multiple components that are already assembled \citep{Despres_Cyr_Setoodeh_Mohammadi2020}. Despite these appealing features and potentials for industrial adoption, industries continue to encounter challenges in adopting AM as the primary manufacturing process for mass production, fundamentally due to inconsistency in part quality and properties as well as process inefficiencies \citep{Li_Jia_Yang_Lee2020, Li_Zhou_Huang_Li_Cao2023, Pandiyan_Cui__Shevchik2022}. Therefore, researchers are continuously analyzing AM processes from different perspectives and proposing methods to overcome these challenges.

Traditionally, AM has largely relied on pure physics-based models utilizing fundamental physical principles to predict behaviors such as thermal distribution during printing, with Finite Element Analysis (FEA) serving as a primary tool \citep{Gibson_Rosen_Stucker_2021}. However, despite their theoretical soundness, pure physics-based models encounter notable limitations such as significant computational expense and sensitivity to model assumptions \citep{huang2020_supp_ref_for_criticism_physics_based_model, Kouraytem2021modeling_supp_ref, Wang_Yang_Liu__Chen2022_Data_driven_AM_modeling_review}. Therefore, data-driven machine learning (ML) has emerged as a promising alternative to the conventional approaches in the domain of AM \citep{Wang_Tan_Tor_Lim_2020_review, Goh_Sing_Yeong_2021_review}.

\begin{table}[htbp]
    \centering
    \caption{List of \textit{review articles in the intersection of AM and ML}} \label{tab:review_AM_ML}
    \tiny
    \begin{tabularx}{1.0\textwidth}{p{2cm} p{1.5cm} p{10.4cm}}
         \hline
         \hline
         \textbf{ML techniques} & \textbf{AM process} & \textbf{Review articles}\\
         \hline
         \multirow{7}{*}{General ML} & General AM & \cite{Razvi_Feng_Narayanan_Lee_Witherrell2019_review}; \cite{Meng_McWilliams__Zhang2020_review}; \cite{Jin_Zhang_Demir_Gu_2020_review};  \cite{Wang_Tan_Tor_Lim_2020_review}; \cite{Goh_Sing_Yeong_2021_review}; \cite{Liu_Tian_Kan2022_when_ai_meets_AM_review};
         \cite{Qin_Hu_Liu__Tang2022_review}; \cite{Chinchanikar_Shaikh_2022_review}; \cite{Sarkon_Safaei__Zeeshan_2022_review}; \cite{Zhang_Safdar_Xie_Li_Sage_Zhao2022_review}; \cite{Xames_Torsha_Sarwar2022_review}; \cite{Kumar_Gopi__Wu2023_review}; \cite{Jiang2023_AM_ML_survey}; \cite{Yi_Xue_Cong__Guo2023_review_fatigue_prediction} \\
         \cline{2-3}
         & L-PBF & \cite{Sing_Kuo_Shih__Chua2021_ML_LPBF_review}; \cite{Mahmoud_Magolon__Mohammadi2021_review}; \cite{Fu_Downey_Yuan__Balogun2022_ML_General_LBAM_review}; \cite{Liu_Ye_Izquierdo__Shao2022_review}\\
         \cline{2-3}
         & WAAM & \cite{Hamrani_Agarwal__McDaniel2023_Ml_AM_review}\\
         \cline{2-3}
         & AJP & \cite{Guo_Ko_Wang2023_ML_AJP_review}\\
         \hline
         \multirow{2}{*}{NN-based} & General AM & \cite{Qi_Chen_Li_Cheng_Li2019_review}; \cite{Valizadeh_Wolff2022}\\
         \cline{2-3}
         & WAAM & \cite{He_Yuan_Mu_Ros__Li2023_review_AI_AM}; \cite{Mattera_Nele_Paolella2023}\\
         \hline
         \hline
    \end{tabularx}
\end{table}

There are many review papers in the literature, as listed in Table \ref{tab:review_AM_ML}, that summarized the implementation of ML techniques across various scenarios in the AM process. \textcolor{red}{Based on the approach to feature extraction, this paper categorizes \textit{general ML} techniques into two: \textit{classical ML} (methods that depend on manual selection and engineering of features) and \textit{deep learning} (methods that automatically learn and extract features from data). Deep learning (DL) methods are distinguished by their ability to learn hierarchical representations, where multiple levels of increasingly abstract features are extracted directly from raw data \citep{LeCun_Bengio_Hinton2015_DL_nature_review, Schmidhuber2015_DL_in_NN_ref_review, Prince_2023_Understanding_deep_learning_MIT_press}. Notably, the classification of DL in this paper includes both end-to-end learning and approaches that involve preprocessing steps (e.g., data normalization, augmentation, or transformation), as long as the central methodology focuses on automatic feature learning.} The majority of the review papers listed in Table~\ref{tab:review_AM_ML} primarily focused on the broader spectrum of general ML applications in AM. While DL applications in AM are naturally included as part of this broader domain, they have received limited or negligible attention in these reviews. However, as the complexity and scope of AM continue to expand, it becomes increasingly evident that DL has become more effective than classical ML in addressing AM challenges due to its ability to (\textit{i}) handle vast amounts of complex data, and (\textit{ii}) automatically capture complex and nonlinear relationships without manual feature extraction \citep{LeCun_Bengio_Hinton2015_DL_nature_review, Goodfellow_Bengio_Courville2016_DL_book}. Though efforts by \cite{Qi_Chen_Li_Cheng_Li2019_review}, \cite{Valizadeh_Wolff2022}, \cite{He_Yuan_Mu_Ros__Li2023_review_AI_AM}, and \cite{Mattera_Nele_Paolella2023} reviewed studies focused on automatic feature extraction using DL in AM, most of these review papers primarily focused on feedforward neural network applications. On the other hand, \cite{Valizadeh_Wolff2022} focused exclusively on articles using convolutional neural networks (CNNs) for AM, overlooking other DL techniques frequently used in the AM domain, such as recurrent neural networks (RNNs) and generative adversarial networks (GANs), to name a few. Therefore, a comprehensive review paper is a timely demand to keep track of the progress and be up to date with the state-of-the-art trends in the domain. To fill this gap, this paper reviews the studies that utilized various DL techniques, rather than classical ML, across diverse AM processes.   

The growing popularity of DL applications in AM stems from DL's ability to address some well-established and clearly defined research problems within the AM process, as summarized in Table \ref{tab:AM_research_problems}. DL techniques offer powerful solutions to these issues, making them a promising choice for advancing AM technology. On top of these frequently reported AM research problems listed in Table \ref{tab:AM_research_problems}, this paper defines the following three overarching research questions (RQs) to further clarify the scope of this review:
\begin{enumerate}
    \item How are DL techniques being utilized to advance the design phase of AM?
    \vspace{-10pt}
    \item How are DL techniques being utilized to advance the modeling of AM processes?
    \vspace{-10pt}
    \item How are DL techniques being utilized to advance the monitoring and control of AM processes?
\end{enumerate}

\begin{table}[htbp]
    \centering
    \caption{Frequently reported AM research problems and the rationale for employing DL solutions to address them} \label{tab:AM_research_problems}
    \tiny
    \begin{tabularx}{1.0\textwidth}{p{8.15cm} | p{6.15cm}}
        \hline
        \hline
        \textbf{Common AM research problems} & \textbf{DL methods' potential to address AM research problems} \\
        \hline
        Designing novel parts with minimum material requirements and desired properties \citep{Despres_Cyr_Setoodeh_Mohammadi2020, Guo_Lu_Fuh2021, Hertlein_Buskohl_Gillman_Vemaganti_Anand2021, Almasri_Danglade__Ababsa_2023, Almasri_Bettebghor__Ababsa2024_generation_mechanical_designs_ConvGANs} & Generative DL techniques such as GANs can be used for generating novel designs while adhering to specific constraints.\\
        \hline
        Predicting parts' geometric deviation for compensation modeling \citep{Shen_Shang_Zhao_Dong_Xiong_Wang2019_error_compensation, Zhao_Xiong_Shang__Wu2019nonlinear_error_prediction, Zhao_Xiong___Zhu2022point_deformation_prediction, Standfield_Wang__Kong2022_shape_deformation_prediction} & DL techniques such as CNNs can be trained on large available datasets of CAD models to predict geometric deviations for given part designs. \\
        \hline
        Optimizing the printing sequence for desired properties (i.e., minimum energy consumption, thermal homogeneity, uniform grain structure, etc.) \citep{Qin_Ding__Liao2024_DRL_toolpath_generation_thermal_uniformity_LPBF,Mozaffar_Ebrahimi_Cao2020toolpath} & DL models such as deep reinforcement learning can learn from feedback to optimize printing sequence during printing.\\
        \hline
        Understanding complex AM process dynamics \citep{Hemmasian_Ogoke__Beuth_Farimani2023, Nalajam_Varadarajan2021, Ren_Chew_Zhang_Fuh_Bi2020, Zhou_Shen_Liu_Du_Jin2021_Thermal_field_prediction, Mozaffar_Paul__Cao2018, Mozaffar_Liao_Lin_Ehmann_Cao2021, Pham_Hoang_Tran_Pham_Fetni_Duchene_Tran_Habraken2023} & DL techniques such as RNN, LSTM, and GRU can capture temporal dependencies to understand the dynamic nature of the AM process.\\
        \hline
        Approximating complex process-structure-property relationships \citep{Croom_Berkson_Mueller_Presley_Storck2022, Fang_Cheng_Glerum_Bennett_Cao_Wagner2022, Zhang_Wang_Gao2018, Lu_He_Shi_Bai_Zhao_Han2021, Maurizi_Gao_Berto2022_stress_strain_deformation_prediction, Qin_DeWitt_Radhakrishnan_Biros2023} & DL techniques such as DNN can automatically approximate highly nonlinear functions from complex data.\\
        \hline
        Extracting information from limited or noisy or imbalanced AM data \citep{Li_Cao_Liu__LI2023_Imbalanced_data_generation_in_situ_monitoring, Tan_Huang_Liu_Li_Wu2022, Kim_Lee__Yoo2023_time_series_augmentation_StyleGAN, Li_Zhang__Wen2023_LPBF_defect_classification_tansfer_learning_AE_Resnet50} & Generative DL techniques such as diffusion models or GANs can synthetically augment data to address limited or imbalanced data issues. \\
        \hline
        Extracting information from high-dimensional sensor data (i.e., image, point cloud, etc.) for \textit{in-situ} process monitoring and control \citep{Akhavan_Lyu_Manoochehri2024, Kaji_Nguyen-Huu_Toyserkani2022_surface_anomaly_detection_DED_point_cloud, Yangue2023_online_surface_prediction, Lee_Heogh_Yang__Lee2022_Powder_Stream_Fault} & Advanced DL techniques such as CNNs can analyze grid-like data such as images or voxels and point cloud neural networks such as PointNet can directly analyze point cloud data. \\
        \hline
        \hline
    \end{tabularx}
\end{table}

\begin{figure}[htbp]
    \centering
    \captionsetup{justification=centering}
    \includegraphics[width=1.0\textwidth]{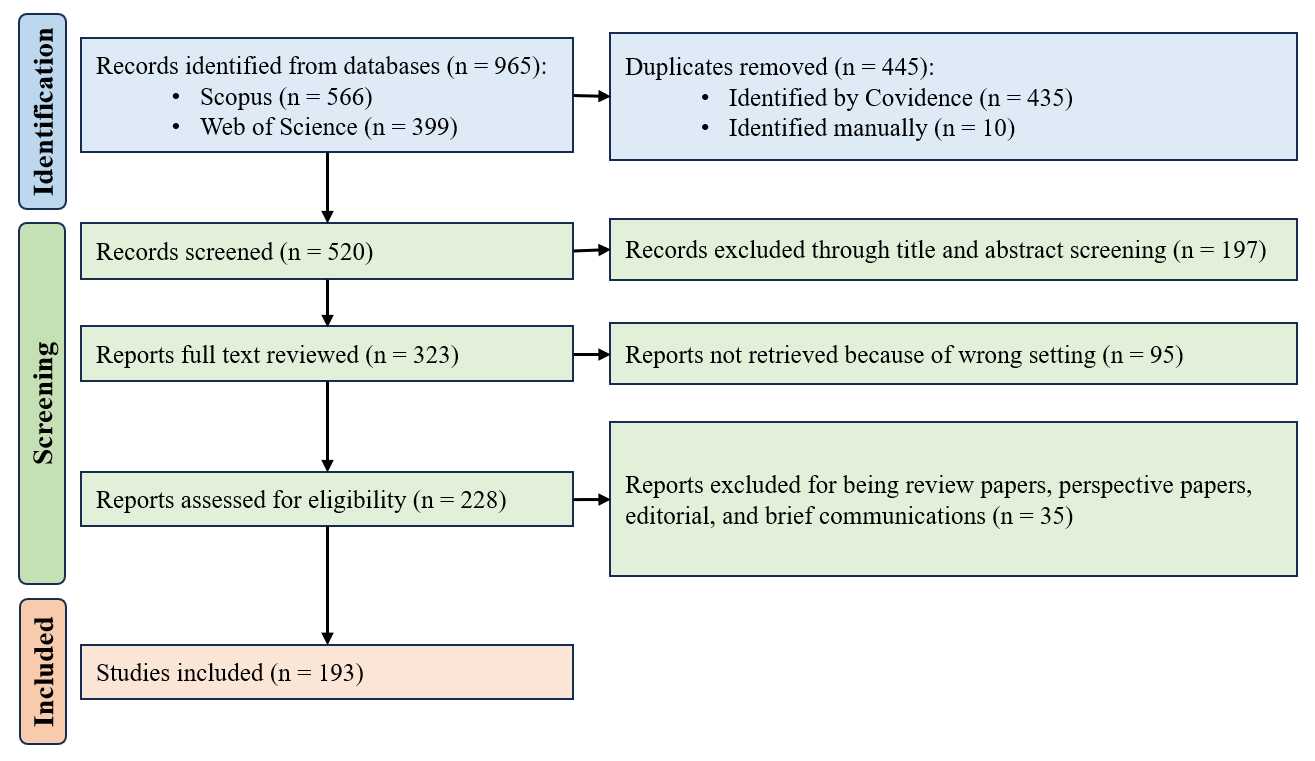}
    \caption{\textit{PRISMA flowchart} detailing the article selection process, adapted from the guidelines by \cite{page2021prisma_Prisma_Reference}} \label{fig: Prisma_DL_AM}
\end{figure}

These questions guide our exploration of how DL is transforming various stages of the AM process. After thorough trial and refinement, we crafted the following search string to cover a wide array of articles at the intersection of DL and AM: (``\textit{Additive Manufacturing}" \textbf{OR} ``\textit{Binder Jetting}" \textbf{OR} ``\textit{Directed Energy Deposition}" \textbf{OR} ``\textit{Material Extrusion}" \textbf{OR} ``\textit{Material Jetting}" \textbf{OR} ``\textit{Powder Bed Fusion}" \textbf{OR} ``\textit{Sheet Lamination}" \textbf{OR} ``\textit{Vat Photopolymerization}") \textbf{AND} (``\textit{Deep Learning}"). The search string includes all seven AM categories as per ISO/ASTM 52900:2021 \citep{ISOASTM52900}, along with the generic term "Additive Manufacturing." To systematically retrieve relevant literature, we conducted a comprehensive search on April 20, 2024, across two well-regarded databases: \textit{Scopus} and \textit{Web of Science}. We finalized the articles through a three-step methodology: identification, screening, and inclusion with the help of the systematic review tool \textit{Covidence} \citep{covidence}. Details of the search process are depicted in the \textit{Preferred Reporting Items for Systematic Reviews and Meta-Analyses} (\textit{PRISMA}) flowchart provided in Figure \ref{fig: Prisma_DL_AM}. All the finalized articles are listed in Table \ref{tab:big_paper_summary_list} to provide a high-level summary of the included papers identified through the systematic process. This table represents our final list of papers that applied DL to AM and are included in our analysis. All other papers listed in the references serve as supplementary or complementary resources and are not considered as part of the reviewed papers.

The papers in Table \ref{tab:big_paper_summary_list} are categorized into seven major AM categories, with an additional \textit{general AM} category for papers addressing AM without specifying a particular technology. The two categories with the most papers are powder bed fusion (PBF) and directed energy deposition (DED). It is worth mentioning that there has been a significant amount of research focused on laser powder bed fusion (L-PBF) technology within the PBF category, a long-standing trend. More recently, there has been a considerable surge in the number of papers utilizing DL for wire arc additive manufacturing (WAAM) technology within the DED category. Moreover, papers within each AM category are further grouped based on specific DL techniques used. Deep neural networks (DNN) and convolutional neural networks (CNN) are fundamental techniques commonly used within larger networks. Therefore, if a paper specifically utilized any DL technique alongside DNN or CNN, it is categorized within the cluster corresponding to that specific DL technique. Otherwise, if DNN or CNN was the sole technique utilized without any others, the paper is categorized as DNN or CNN. Papers utilizing TL are grouped into the CNN cluster.

\begin{table}[htbp]
    \centering
    \caption{List of articles applying DL for AM applications, selected through step-by-step PRISMA methodology and included in the review analysis of this paper} \label{tab:big_paper_summary_list}
    \tiny
    \begin{tabularx}{1.0\textwidth}{p{2cm} p{12.3cm}}
        \hline
        \hline
        \textbf{AM process categories} & \textbf{Articles applying DL for AM with specific DL techniques} \\
        \hline
        \textit{General AM} & \textbf{DNN} \citep{Zhao_Xiong___Zhu2022point_deformation_prediction, Ha_Yao__Zheng2023rapid_inverse_design, Lee_Zhang_Gu2022_NN_GO_lattice_structures_superior_mechanical_properties}; \textbf{CNN} \citep{Peng_Liu_Huang__Lu2022, Jiang_Xiong_Zhang_Rosen2022_DNN_DfAM, Huang_Sun_Kwok_Zhou_Xu2020, Williams_Meisel_Simpson_McComb2019design_repository_manufacturability, Shen_Shang_Zhao_Dong_Xiong_Wang2019_error_compensation, Senanayaka_Tian_Falls_Bian2023_Porosity_prediction, Zhao_Xiong_Shang__Wu2019nonlinear_error_prediction, Shen_Shang_Li__Wang2019_error_prediction_compensation, Standfield_Wang__Kong2022_shape_deformation_prediction, Liu_Wang_Ho_Kong__Chase2022, Garland_White__Boyce2020, Jiang_Smith__Rollett2024_laser_absorptance_prediction_CNN_AM, Hilbig_Vogt_Holtzhausen_Paetzold2023_geometric_feature_recognition}; \textbf{GAN} \citep{Almasri_Danglade__Ababsa_2023,Almasri_Danglade__Ababsa2022, Hertlein_Buskohl_Gillman_Vemaganti_Anand2021, Wang_Wang__Li2023_suface_Defect_Detection_AM_YOLOv8, Almasri_Bettebghor__Ababsa2024_generation_mechanical_designs_ConvGANs}; \textbf{GCN} \citep{Despres_Cyr_Setoodeh_Mohammadi2020}; \textbf{GNN} \citep{Maurizi_Gao_Berto2022_stress_strain_deformation_prediction}; \textbf{LSTM} \citep{Qin_DeWitt_Radhakrishnan_Biros2023, Wang_Shraida_Jin2023predictive_cloud} \\
        \hline
        \textit{Binder Jetting} & \textbf{CNN} \citep{Mehta_Shao2022_Federated_defect_detection}; \textbf{DRL} \citep{Li_Segura__Sun2023_Droplet_Pinch-Off_Behaviors_Identification_Inkjet_DRL_GCN}; \textbf{GCN} \citep{Li_Segura__Sun2023_Droplet_Pinch-Off_Behaviors_Identification_Inkjet_DRL_GCN}; \textbf{RNN} \citep{Inyang-Udoh_Chen_Mishra2022, Huang_Segura__Zhou2020_droplet_evolution_prediction_inkjet_DRNN, Inyang-Udoh_Mishra2021_PINN}; \textbf{LSTM} \citep{Kwon_Kim__Lee2023_PIDL_inkjet_ConvLSTM_next_image_prediction_from_sequential_images} \\
        \hline
        \textit{Directed Energy Deposition} & 
        \textbf{DNN} \citep{Biehler_Shi2024retrofit, Pham_Hoang_Tran_Pham_Fetni_Duchene_Tran_Habraken2023, Chen_Wong__LI2022_Bead_geometry_prediction_DED_DNN, Pham_Hoang__Habraken2022_uncertainties_DED};
        \textbf{CNN} \citep{Mi_Zhang_Li_Shen__Mai2023, Chen_Yao_Tan___Moon2023_pore_detection,Pandiyan_Cui__Shevchik2022, Jamnikar_Liu_Brice_Zhang2023_process-optimization, Chen_Yao_Moon2022, Lee_Heogh_Yang__Lee2022_Powder_Stream_Fault, Davtalab_Kazemian_Yuan_Khoshnevis2022, Chen_Yao_Feng_Chew_Moon2023_multimodal_fusion_in_situ_defect_detection, Fang_Cheng_Glerum_Bennett_Cao_Wagner2022, shi2022hybrid, Zhu_Jiang_Guo_Xu__Jiang2023_surface_morphology_inspection_TL, Xia_Pan_Li_Chen_Li2022, Wang_Lu_Zhao_Deng__Yao2021, Kim_Lee_Seo_Kim_Shin2023_in_Situ_monitoring, Surovi_Soh2023_monitoring_Defect_identification, Lu_He_Shi_Bai_Zhao_Han2021, Shin_Hong_Jadhav_KIm2023_defect_detection_transfer_learning, Perani_Baraldo__Paoli2023, Kim_Chong__Shin2024_WAAM_defect_detection_CNN_transfer_learning, Glasder_Fabbri__Wegener2023_footprint_weld_bead_prediction_using_transfer_learning_WAMM_CNN, Li_Zhang__Li2023_surface_defect_detection_WAAM, Wang_Xu__Yao2021_deposited_layer_width_and_reinforcement_prediction_WAAM, Pandiyan_Cui__Leparoux2023_LDED_meltpool_image_and_AE_CNN_ViT, Liu_Yuan__Weiwei2022_meltpool_states_classification_DED_ResNet, Kim_Oh__Kim2022_pore_detection_YOLOv5_pore_image_DED, Kaji_Nguyen-Huu_Toyserkani2022_surface_anomaly_detection_DED_point_cloud, Zhang_Wen_Chen2019_Weld_image_defect_detection_CNN_Robotic_arc_welding, Li_Siahpour__Shi2020_CNN_DED_thermal_images, Herbeaux_Aboleinein__Klocker2024_microstructure_modeling_from_SEM_EBSD_images_CNN_WAAM_or_CMT};
        \textbf{GAN} \citep{Kim_Lee__Yoo2023_time_series_augmentation_StyleGAN, Liu_Wang_Tian_Lauria_Liu2021, Chen_Guo2023_DCGAN-CNN_porosity_prediction_LMD_unbalanced, Guo_Guo_Bian_Guo2022, Guo_Tian_Guo_Guo2020, Pandiyan_Cui__Wasmer2022_DED_meltpool_state_prediction_meltpool_image_GAN_CNN};
        \textbf{GNN} \citep{Mozaffar_Liao_Lin_Ehmann_Cao2021};
        \textbf{RNN} \citep{Mozaffar_Liao_Lin_Ehmann_Cao2021, Mozaffar_Paul__Cao2018, Zhou_Shen_Liu_Du_Jin2021_Thermal_field_prediction};
        \textbf{LSTM} \citep{Ren_Wen_Zhang_Mazumder2022_quality_monitoring, Hu_Wang_Li_Wang2022, Nalajam_Varadarajan2021, Perani_Jandl_Baraldo_Valente_Paoli2023_modeling_track_geometry, Wang_Zhang__Han2020_Weld_reinforcement_prediction_molten_pool_image_WAAM_LSTM, Abranovic_Sarkar__Beuth2024_predicting_next_frame_meltpool_for_flaw_detection_in_video_data_DED_ConvLSTM, Perumal_Abueidda__Kontsos2023_TCN_data-driven_thermal_prediction_DED};
        \textbf{DRL} \citep{Petrik_Bambach2023_path_planning_reinforcement};
        \textbf{TCN} \citep{Wang_Hu_Li_Wang2023_meltpool_width_layerHeight_prediction, Perumal_Abueidda__Kontsos2023_TCN_data-driven_thermal_prediction_DED};
        \textbf{Transformer} \citep{Zhang_Xu__Wang2023_WAAM_detect_surface_oxidation_defects_Transformer_time_series_voltage_data, Chen_Yang__Rong2023_Physics-Informed_Attention_Network_Condition_Monitoring_WAAM, Pandiyan_Cui__Leparoux2023_LDED_meltpool_image_and_AE_CNN_ViT}\\
        \hline
        \textit{Material Extrusion} & 
        \textbf{CNN} \citep{Akhavan_Lyu_Manoochehri2024, Brion_Shen_Pattinson2022, Jin_Zhang_Gu2019, Li_Zhang_Zhou__Zhang2023, Kim_Lee_Ahn2022, Lyu_Akhavan__Monoochehri2021_monitoring_anomaly, Banadaki_Razaviarab_Fekrmandi_Sharifi2020, Saluja_Xie_Fayazbakhsh2020, Wright_Darville_Celik__Celik2022, Kim_Lee__Ahn2020_failure_detection_ME_VGGNet_transfer_learning};
        \textbf{GNN} \citep{Yang_Kan2023_3D_Geometry_Representation_AM};
        \textbf{GAN} \citep{Tan_Huang_Liu_Li_Wu2022, Petros_Siegkas2022, Li_Shi__Williams2021_ATR_GAN};
        \textbf{GCN} \citep{Li_Segura__Sun2023_Droplet_Pinch-Off_Behaviors_Identification_Inkjet_DRL_GCN};
        \textbf{RNN} \citep{Inyang-Udoh_Chen_Mishra2022, Huang_Segura__Zhou2020_droplet_evolution_prediction_inkjet_DRNN, Inyang-Udoh_Mishra2021_PINN};
        \textbf{DRL} \citep{Li_Segura__Sun2023_Droplet_Pinch-Off_Behaviors_Identification_Inkjet_DRL_GCN};
        \textbf{LSTM} \citep{Yangue2023_online_surface_prediction, Zhang_Wang_Gao2019, shi_Mamun__Liu2022_LSTM_autoencoder, Zhang_Wang_Gao2018, Castro_Pathinettampadian__Subramaniyan2023_Prediction_compressive_strength_LSTM_FDM, Khusheef_Shahbazi_Hashemi2023_copare_three_levels_of_fusion_FDM_LSTM_transfer_learning, Kwon_Kim__Lee2023_PIDL_inkjet_ConvLSTM_next_image_prediction_from_sequential_images};
        \textbf{DF} \citep{Ye_Liu__Tian_Kan2020_monitoring_point_cloud};
        \textbf{Transformer} \citep{Li_Huang__Tan_2024)_defect_classification_ViT_FDM}\\
        \hline
        \textit{Material Jetting} & 
        \textbf{DRL} \citep{Li_Segura__Sun2023_Droplet_Pinch-Off_Behaviors_Identification_Inkjet_DRL_GCN};
        \textbf{LSTM} \citep{Kwon_Kim__Lee2023_PIDL_inkjet_ConvLSTM_next_image_prediction_from_sequential_images};
        \textbf{RNN} \citep{Segura_Li__Sun2023_Tensor_time_series_TGCN_TRNN_Material_jetting, Inyang-Udoh_Chen_Mishra2022, Huang_Segura__Zhou2020_droplet_evolution_prediction_inkjet_DRNN, Inyang-Udoh_Mishra2021_PINN};
        \textbf{GCN} \citep{Li_Segura__Sun2023_Droplet_Pinch-Off_Behaviors_Identification_Inkjet_DRL_GCN, Segura_Li__Sun2023_Tensor_time_series_TGCN_TRNN_Material_jetting}\\
        \hline
        \textit{Powder Bed Fusion} & 
        \textbf{DNN} \citep{Zhao_Wei_Mao__Liao2023_prediction_params_melt_pool_dimensions_PIDL, Ghungrad_Faegh_Gould_Wolff_Haghighi2023_PIDL, Tu_Liu_Carneiro__Harrison2022, Pandita_Ghosh_Gupta__Wang2022_process_modeling, Mohammadi_Mahmoud_Elbestawi2021, Park_Choi_Jhang2021, Ghungrad_Gould_Soltanalian_Wolff_Haghighi2021, Koc_Zeybek_Kisasoz__Bulduk2022, Ghungrad_Gould_Wolff_Haghighi2022_physics_informed_AI, Kwon_Kim_Ham__Kim2020, Mohammed_Almutahhar__Ali2023_porosity_prediction_LPBF_DNN, Sharma_Raissi_Guo2023_PIDL_multi-physical_PBF, Fotovvati_Chou2022_SLM_surface_roughness_prediction_ANN};
        \textbf{CNN} \citep{Scime_Joslin__Paquit2023_Tensile_Property_Prediction_LPBF, Imani_Chen_Diewald__Yang2019_defect_detection, Surana_Lynch_Nassar__Overdorff2023, Hemmasian_Ogoke__Beuth_Farimani2023, Kim_Yang_Ko_Cho_Lu2023, Pandiyan_Drissi-Daoudi_Wasmer2022, Johnson_Maestas_Martinez2022, Croom_Berkson_Mueller_Presley_Storck2022, Schmid_Krabusch_Schromm__Schleifenbaum2021, Baumgartl_Tomas_Buettner_Merkel2020, Fathizadan_Ju_Lu2021, Scime_Beuth2018, Williams_Dryburgh_Clare_Rao_Samal2018, Zhang_Liu_Shin2019, Zhang_Yang_Dong_Zhao2021_predictive_manufacturability, Drissi-Daoudi_Pandiyan_Wasmer2022, Zhang_Zhao2022, Scime_Siddel_Baird_Paquit2020, Tan_Fang_Li__Yang2020, Mehta_Shao2022_Federated_defect_detection, Ansari_Crampton_Garrard_Cai_Atallah2022_porosity_preidction, Snow_Diehl__Nassar2021_in_situ_flaw_detection, Snow_Reutzel_Petrich2022_Correlating_monitoring_data_to_fatigue_performance, Sofi_Ravani2023, Kim_Zohdi2022, Westphal_Seitz2021, Jin_Tan__Sangiovanni-Vincentelli2019, Zhu_Ferreira_Anwar__Qiao2020, Yuan_Giera_Guss__Mcmains2019, Li_Zhou_Huang_Li_Cao2023, Wang_Cheung2022, Oster_Breese__Altenburg2023_defect_prediction_LPBF, Yang_Qiu__Bai2023_Defect_classification_LPBF_simulated_meltpool_images_and_thermal_images_CNN, Pandiyan_Wrobel__Shevchik_2024_domainAdaptation_LPBF_CNN_AE, Li_Zhang__Wen2023_LPBF_defect_classification_tansfer_learning_AE_Resnet50, Desrosiers_Letenneur__Brailovski2024_porosity_segmentation_LPBF_computed_tomography_CNN, Jiang_Zhang__ZHang2023_CNN_layerwise_imagesLPBF_surface_defect_classification, Yang_Wang__Fang2022_UNet_XCT_image_reconstruction_prediction_compression_lattice_structures, Zhang_Fu__Schleifenbaum2023_image_quality_enhancement_Unet_LPBF, Shevchik_Masinelli__WAsmer2019_SCNN_quality_level_classification_PBF, Manivannan2023_powder_bed_defect_detection_SLS_CNN, Anidjar_Lang_Mega2024)_Transfer_Learning_Detection_Fatigue_Crack_Initiation_SLM_YOLOv5, Rezasefat_Hogan2024_Prediction_4D_stress_evolution_SLM_CNN_voxel, Bellens_Probst__Dewulf2022_U_Net_porosity_segmentation_XCT_scan_SLS, Wang_Cheung_wang2023_surface_Defect_Segmentation_SLM_CNN_with_attention};
        \textbf{GAN} \citep{Chen_Liu__Witherell2024_MeltpoolGAN_Melt_pool_prediction_from_path-level_thermal_history, Li_Cao_Liu__LI2023_Imbalanced_data_generation_in_situ_monitoring, Pandiyan_Drissi-Daoudi_Shevchik__Wasmer2021_semi_supervised_monitoring, Zhang_Sahu_Singh__Lu2023_meltpool_morphology_prediction, Song_Wang_Gao_Son_Wu2023_pore_prediction, Ramlatchan_Li2022, Guo_Lu_Fuh2021, Li_Cao__Zhang2023_Imbalanced_SLM_AE};
        \textbf{DBN} (\citealp{Ye_Fuh_Zhang_Hong_Zhu2018}, \citealp{Ye_Hong_Zhang_Zhu_Fuh2018});
        \textbf{RNN} \citep{Voigt_Moeckel2022, Larsen_Hooper2022, Ho_Zhang_Young__Mozumder2021, Williams_Sing2024_Spatiotemporal_PBF_meltpool_monitoring_videos_convRNN, Luo_Ma__Cao2021_1DCNN_LSTM_RNN_GRU_spatter_defect_classification_SLM};
        \textbf{LSTM} \citep{Mao_Lin_Yu_Frye_Beckett_Anderson___Agarwal2023, Ko_Kim_Lu_Shin_Yang_Oh2022_spatial, Pandiyan_Masinelli_Claire__Wasmer2022, Ko_Kim_Lu_Shin_Yang_Oh2022_spatial, Zhang_Sahu_Singh__Lu2023_meltpool_morphology_prediction, Park_Choi_Um2024_ConvLSTM_meltpool_prediction_from_images_of_laser_tool_path_strategy_LPBF, Fathizadan_Ju__Yang2023_LPBF_ConvLSTM_Anomaly_Detection, Zhang_Vallabh__Zhao2022_meltpool_temperature_and_morphology_monitoring_LSTM_LPBF, Luo_Ma__Cao2021_1DCNN_LSTM_RNN_GRU_spatter_defect_classification_SLM};
        \textbf{GRU} \citep{Luo_Ma__Cao2021_1DCNN_LSTM_RNN_GRU_spatter_defect_classification_SLM};
        \textbf{DRL} \citep{Ogoke_Farimani2021, Qin_Ding__Liao2024_DRL_toolpath_generation_thermal_uniformity_LPBF}\\
        \hline
        \textit{Sheet Lamination} & N/A \\
        \hline
        \textit{Vat Photopolymerization} & \textbf{CNN} \citep{Khadilkar_Wang_Rai2019_stress_prediction_SLA_CNN}; \textbf{LSTM} \citep{Lee_Saha_Sarkar_Giera2020} \\
        \hline
        \hline
    \end{tabularx}
\end{table}

This paper assumes readers have a preliminary understanding of both AM processes and DL techniques, targeting individuals interested in a brief overview of how DL is propelling advancements in AM. Considering the scope and space limitations, the focus is on providing a summary of the intersection of these two vibrant domains - DL and AM. The high-level contributions of this paper can be summarized as follows:
\begin{enumerate}
    \item This is the first paper, to the best of our knowledge, that comprehensively reviews most of the existing literature within the systematically defined scope applying DL to diverse aspects of AM life-cycle including design, process modeling and optimization, as well as process monitoring and control. Furthermore, for readers seeking a quick overview, this paper provides a big-picture summary of all reviewed papers that used DL for AM in a single table (Table \ref{tab:big_paper_summary_list}).
    
    \item Compared to the existing review papers (Table \ref{tab:review_AM_ML}), this paper specifically focuses only on the research applying DL in AM. All other papers reviewed studies applying general ML techniques (both classical ML and DL) in AM where DL applications are reviewed with partial attention.
    
    \item In addition to the state-of-the-art literature review, this paper also offers recommendations and guidelines for further research in this direction. Therefore, it would be a great source for the new researchers interested in the intersection of AM and DL, to capture existing progress within a short time and focus on the potential gaps.   
\end{enumerate}

\begin{table}[htbp]
    \centering
    \caption{\textit{List of abbreviations and their meaning used in this paper}} \label{tab:abbreviations_list}
    \tiny
    \begin{tabularx}{1.0\textwidth}{p{0.5cm} p{13.80cm}}
        \hline
        \hline
        \textbf{Set} & \textbf{\textit{List of abbreviations and their meaning}} \\
        \hline
        A-N & AE: acoustic emissions; AJP: aerosol jet printing; AM: additive manufacturing; CNN: convolutional neural networks; DDDL: data-driven deep learning; DBN: deep belief network; DED: directed energy deposition; DL: deep learning; DNN: deep neural networks; DRL: deep reinforcement learning; DTL: deep transfer learning; DfAM: design for additive manufacturing; FDM: fused deposition modeling; FE: finite element; FFF: fused filament fabrication; GAN: generative adversarial network; FL: federated learning ; GCN: graph Convolutional network; GNN: graph neural network; GRU: gated recurrent units; LSTM: long short-term memory; LMD: laser metal deposition; L-PBF: laser powder bed fusion; ML: machine learning; MAM: metal additive manufacturing; MJ: material jetting; NN: neural network \\
        \hline
        O-Z & PIDL: physics-Informed Deep Learning; PRISMA: preferred reporting Items for systematic reviews and meta-analyses; PSP: process-structure-property; PBF: powder bed fusion; SLA: stereolithography; SLS: selective laser sintering; SLM: selective laser melting; RNN: recurrent neural networks; RQ: research questions; TCN: temporal convolutional networks; TL: transfer learning; TO: topology optimization; WAAM: wire arc additive manufacturing; XCT: X-ray computed tomography \\
        \hline
        \hline
    \end{tabularx}
\end{table}

Throughout this paper, we use various abbreviations; a comprehensive list can be found in Table \ref{tab:abbreviations_list}. The rest of the paper is organized as follows. Section \ref{DL_for_AM} systematically reviews the implementations of DL in AM by clustering articles in different subsections - design for AM (Subsection \ref{DL_for_AM_DfAM}), modeling of AM processes (Subsection \ref{DL_for_AM_data_driven_AM_modeling}), and process monitoring and control (Subsection \ref{DL_for_AM_monitoring_control}), according to a proposed methodology framework. Section \ref{challenges_future_directions} lists the challenges and potential research scopes in the intersection of DL and AM based on the analysis and findings in Section \ref{DL_for_AM}, and finally, Section \ref{conclusion} completes the paper with relevant concluding remarks.

\section{Deep learning techniques in additive manufacturing applications} \label{DL_for_AM}

This section aims to systematically review the implementation of DL in AM in detail. Before proceeding, we summarize AM data. The successful implementation of DL for AM relies heavily on the quality and availability of AM data \citep{Zhang_Safdar_Xie_Li_Sage_Zhao2022_review}. AM Data comes from diverse sources and can be classified in various ways. One possible approach is the classification based on data dimensions - (\textit{i}) \textit{1D AM data} consists of time series and sensor readings, capturing dynamic changes during the process \citep{Ye_Hong_Zhang_Zhu_Fuh2018, Mozaffar_Liao_Lin_Ehmann_Cao2021, Li_Zhang_Zhou__Zhang2023}, (\textit{ii}) \textit{2D AM data} includes images and layer-wise process information, providing insights into the surface and structural features of the manufactured objects \citep{Despres_Cyr_Setoodeh_Mohammadi2020, Westphal_Seitz2021, Wang_Chandra_Huang_Tor_Tan2023}, and (\textit{iii}) \textit{3D AM data} involves volumetric representations and complex geometries, allowing for a comprehensive understanding of the spatial characteristics of the printed parts \citep{Huang_Sun_Kwok_Zhou_Xu2020, Zhu_Ferreira_Anwar__Qiao2020, Ye_Liu__Kan2021_point_cloud_fusion}. Considering the scope and space limitations, we do not delve into various AM data types, referring readers to Figure \ref{fig:AM_Data_classification}. However, as examples, we provide insights into video data and sequences of point cloud data. Video data captures the entire AM process in real-time, providing a continuous stream of information that can be analyzed by DL models to detect anomalies \citep{Kim_Chong__Shin2024_WAAM_defect_detection_CNN_transfer_learning, Abranovic_Sarkar__Beuth2024_predicting_next_frame_meltpool_for_flaw_detection_in_video_data_DED_ConvLSTM}, monitor process stability \citep{Yuan_Giera_Guss__Mcmains2019, Williams_Sing2024_Spatiotemporal_PBF_meltpool_monitoring_videos_convRNN}, and ensure quality control \citep{Lee_Saha_Sarkar_Giera2020}. Point cloud sequences capture the 3D spatial coordinates of points on the surface of the printed part over time. DL techniques can be used to process these sequences to detect geometric inconsistencies \citep{Yang_Kan2023_3D_Geometry_Representation_AM}, surface morphology \citep{Yangue2023_online_surface_prediction}, and process shift \citep{Ye_Liu__Tian_Kan2020_monitoring_point_cloud}.

\begin{figure}[htbp]
    \centering
    \captionsetup{justification=centering}
    \includegraphics[width=1.0\textwidth]{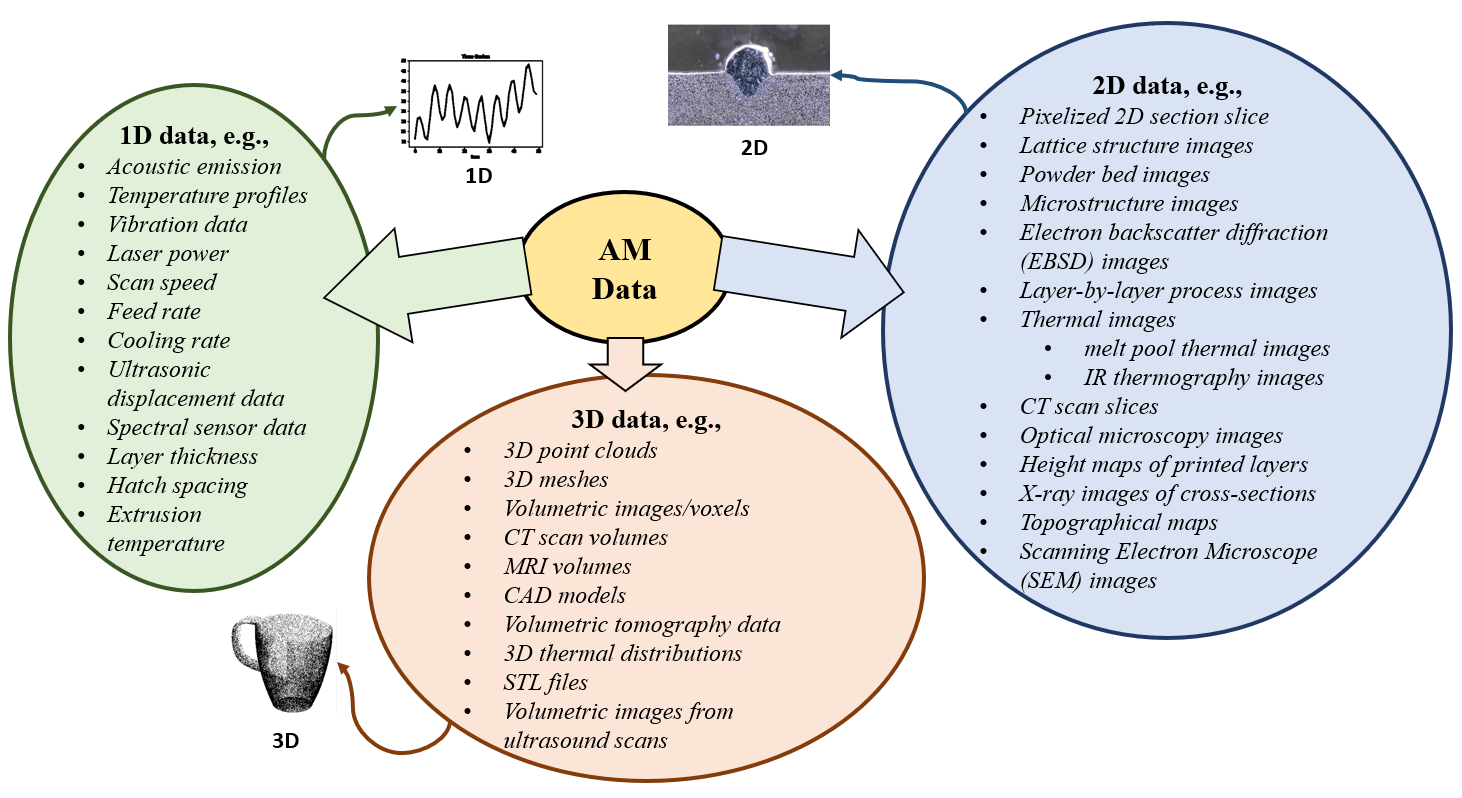}
    \caption{Summary of \textit{AM data types} extracted from reviewed literature. These data types serve as inputs for DL models which leverage the inherent temporal, spatial, and structural information to enhance AM performance and quality.} \label{fig:AM_Data_Classification}
\end{figure}

\begin{figure}[htbp]
    \centering
    \captionsetup{justification=centering}
    \includegraphics[width=1.0\textwidth]{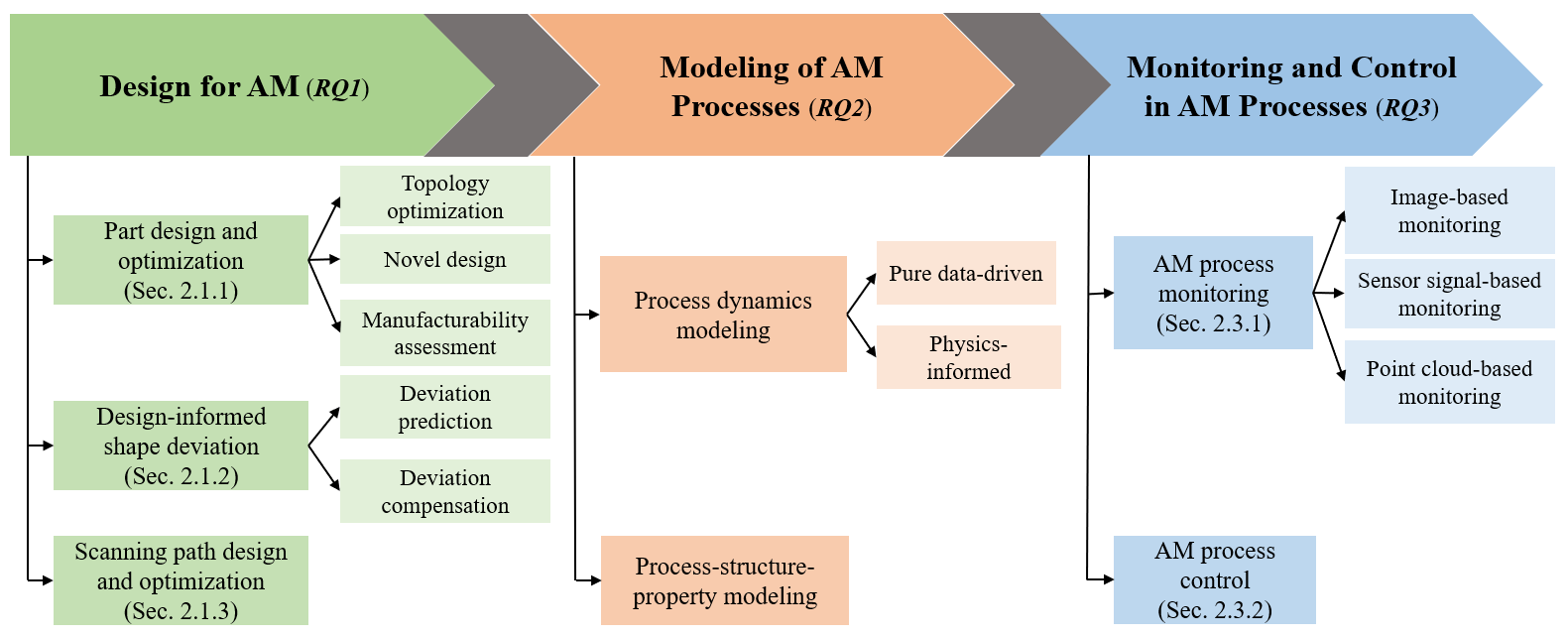}
    \caption{Proposed \textit{methodology framework} for deep learning applications in various aspects of additive manufacturing}
    \label{fig:Methodology_framework_DL_AM}
\end{figure}

In alignment with the overarching three research questions mentioned in Section \ref{s:intro} and the scope of the paper, we have proposed a methodology framework (Figure \ref{fig:Methodology_framework_DL_AM}) to delve into the applications of DL in three key areas of AM: design (Subsection \ref{DL_for_AM_DfAM}), modeling (Subsection \ref{DL_for_AM_data_driven_AM_modeling}), and monitoring and control (Subsection \ref{DL_for_AM_monitoring_control}). This framework serves as the basis for structuring the current section.

\subsection{\emph{Design for AM (DfAM)}} \label{DL_for_AM_DfAM}
\begin{figure}[htbp]
    \centering
    \captionsetup{justification=centering}
    \includegraphics[width=1.0\textwidth]{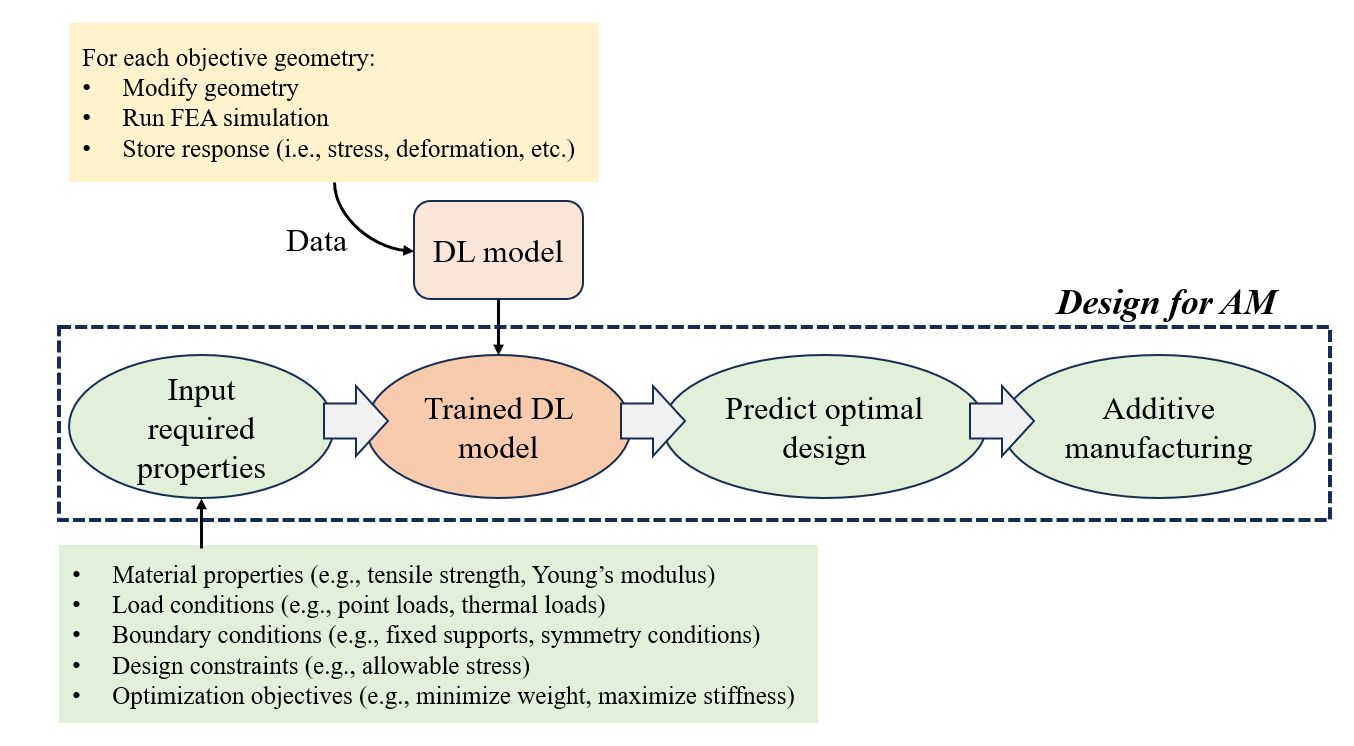}
    \caption{Proposed framework of \textit{DL-enabled design for AM}}
    \label{fig:Design_for_AM}
    \vspace{-10pt}
\end{figure}

The design phase significantly impacts the overall product cost, accounting for over 70\% according to \cite{Almasri_Danglade__Ababsa_2023}. This impact is even greater in AM due to its design freedom, allowing the creation of complex parts in almost (theoretically) any shape \citep{Jiang_Xiong_Zhang_Rosen2022_DNN_DfAM}. Design for additive manufacturing (DfAM) aims to enhance the quality and performance of AM products by optimizing design in a cost-effective and time-efficient manner \citep{Gibson_Rosen_Stucker_2021}. \cite{Jiang_Xiong_Zhang_Rosen2022_DNN_DfAM} proposed an ML-integrated DfAM framework. Motivated by this, we propose a customized and more detailed framework of DL-enabled DfAM, as shown in Figure \ref{fig:Design_for_AM}. The framework illustrates how DL enhances DfAM by optimizing geometries through an iterative process. Initial geometries undergo modification and simulation after inputting the required properties and constraints. Subsequently, the simulation results train a DL model to predict and refine optimal designs, which are then manufactured using AM processes.

\subsubsection{\emph{Part design and design optimization}}

\textbf{Topology optimization (TO)} aims to determine the optimal material distribution within a predefined design domain, traditionally requiring computationally expensive FE simulations \citep{Ibhadode_Zhang__Toyserkani2023topology_Supplemenatary_article}. DL techniques are explored to circumvent this computational burden by identifying hidden patterns from multiple TO iterations \citep{Hertlein_Buskohl_Gillman_Vemaganti_Anand2021}, computing complex and computationally intensive indexes \citep{Iyer_Mirzendehdel__Robinson2021pato}, and incorporating both mechanical and geometrical constraints into TO formulations \citep{Hertlein_Buskohl_Gillman_Vemaganti_Anand2021, Almasri_Danglade__Ababsa2022, Almasri_Danglade__Ababsa_2023}. Among these, \cite{Hertlein_Buskohl_Gillman_Vemaganti_Anand2021} claimed to be the first article implementing DL that outperforms traditional TO in AM applications even with limited test cases. \cite{Hertlein_Buskohl_Gillman_Vemaganti_Anand2021} utilized a conditional generative adversarial network (cGAN) to capture latent patterns from multiple similar TO runs which helps in predicting near-optimal designs without running from scratch each time. They included both mechanical and AM constraints into TO for the first time, ensuring that designed shapes encountered no difficulties during manufacturing \citep{Almasri_Danglade__Ababsa2022, Almasri_Danglade__Ababsa_2023}. Motivated by the work of \cite{Hertlein_Buskohl_Gillman_Vemaganti_Anand2021}, \cite{Almasri_Danglade__Ababsa2022} proposed a DL-AM-TO model where they integrated both mechanical and manufacturing constraints. Though both \cite{Hertlein_Buskohl_Gillman_Vemaganti_Anand2021} and \cite{Almasri_Danglade__Ababsa2022} employed GAN-based generative methods, their architectures differed. Subsequently, \cite{Almasri_Danglade__Ababsa_2023} further improved the DL-AM-TO architecture by converting it from a regression-type to a classification-type problem, resulting in better performance. Additionally, \cite{Iyer_Mirzendehdel__Robinson2021pato} integrated producibility constraints into standard TO using an attention-aware deep CNN model to mitigate cracking caused by steep thermal gradients.

\begin{table}[htbp]
    \centering
    \tiny
    \caption{Articles applying DL techniques for \textit{part design and design optimization}}
    \label{tab:DfAM}
    \begin{tabularx}{1.0\textwidth}{p{2.6cm} p{1.1cm} p{2.6cm} p{1.5cm} p{1.1cm} p{3.65cm}}
        \hline
        \hline
        \textbf{Applications} & \textbf{AM Processes} & \textbf{Articles} & \textbf{DL Techniques} & \textbf{Input Data} & \textbf{Evaluation Metrics}\\
        \hline
        \multirow{3}{*}{Topology optimization} & \multirow{2}{*}{General AM} & \cite{Hertlein_Buskohl_Gillman_Vemaganti_Anand2021} & cGAN & Image & MSE\\
        \cline{3-6}
        & & \cite{Almasri_Danglade__Ababsa2022} & cGAN & Image & Relative error\\
        \cline{2-6}
        & L-PBF & \cite{Iyer_Mirzendehdel__Robinson2021pato} & 3D CNN & Voxel & Relative error and Accuracy\\
        \hline
        \multirow{3}{*}{Novel micro-lattice design} & \multirow{3}{*}{General AM} & \cite{Despres_Cyr_Setoodeh_Mohammadi2020} & GCN & 2D lattice & AUC\\
        \cline{3-6}
        & & \cite{Lee_Zhang_Gu2022_NN_GO_lattice_structures_superior_mechanical_properties} & DNN & Coordinates & RMSE \\
        \cline{3-6}
        & & \cite{Almasri_Bettebghor__Ababsa2024_generation_mechanical_designs_ConvGANs} & ConvGAN & 2d lattice & MSE\\
        \hline
        \multirow{3}{2.5cm}{Manufacturability assessment} & SLS & \cite{Guo_Lu_Fuh2021} & GAN & Voxel & Accuracy, AUC-ROC, and F1-score\\
        \cline{2-6}
        & \multirow{2}{*}{L-PBF} & \cite{Zhang_Yang_Dong_Zhao2021_predictive_manufacturability} & CNN & Voxel & Accuracy and IoU\\
        \cline{3-6}
        & & \cite{Zhang_Zhao2022} & Sparse CNN & Voxel & Accuracy and IoU\\
        \hline
        \hline
    \end{tabularx}
\end{table}

\textbf{Novel micro-lattice design} is gradually becoming important due to the attractive properties of the micro-lattices including a high strength-to-weight ratio and excellent energy absorption capabilities, to name a few. AM can effectively print lattice structures. However, it still requires substantial research for the analysis and design \citep{Rashed_Ashraf_Mines_Hazell2016_Supp, Ha_Yao__Zheng2023rapid_inverse_design}. Therefore, \cite{Despres_Cyr_Setoodeh_Mohammadi2020} proposed a deep autoencoder model for creating novel lattice structures where they used 2D lattice structures as the input. Similarly, \cite{Almasri_Bettebghor__Ababsa2024_generation_mechanical_designs_ConvGANs} presented a geometrically-driven generation of mechanical designs through a triple-discriminator GAN to integrate layout and mechanical constraints. This model generates mechanically valid 2D designs quickly and demonstrates creativity by proposing multiple shapes for the same mechanical constraints.

\textbf{Manufacturability assessment} (also known as printability assessment) can address the potential challenges in printing complex geometric parts (e.g., metal cellular structures) early in the design phase. \cite{Guo_Lu_Fuh2021} proposed a semi-supervised deep autoencoder GAN to assess the manufacturability of cellular structures in a direct metal laser sintering process. \cite{Zhang_Yang_Dong_Zhao2021_predictive_manufacturability} proposed another manufacturability assessment model where they combined CNNs for design considerations and NNs for process aspects. They only considered the low-resolution cases as their pre-processing step involved computationally expensive voxelization. To potentially resolve the issue of computation, \cite{Zhang_Zhao2022} proposed to use the concept of sparsity in the design which helped the model to be more generalized as well as capacitated high-resolution voxelization. 

A high-level summary of \textit{part design and design optimization} subsection is provided in Table \ref{tab:DfAM}.


\subsubsection{\emph{Geometric shape deviation: prediction and compensation}} \label{DL_for_AM_geometric_shape_deviation}

Geometric shape deviation refers to the discrepancies between the designed model and the final printed part. Early work on deviation prediction and compensation, including \cite{Qiang_huang2014statistical_modeling_in_plane1_supp} and \cite{Qiang_huang2015optimal_shrinkage_in_plane2_supp} for in-plane deviation, \cite{Qiang_huang2016analytical_out_of_plane_supp} for out-of-plane deviation, and the recent \cite{ruiz2022prediction_decompose_deviation_into_global_local_roughness_supp} which decomposes deviation into global deviation and local roughness patterns, have demonstrated significant effectiveness due to their interpretability and lower computational requirements. However, these models often rely on assumptions like \textit{independent and identically distributed} deviations and are validated primarily on simplified, mostly \textit{cylindrical} shapes. While simpler models are still valuable in many contexts, DL excels at handling complex patterns and large datasets. As discussed in the introduction, this capability makes DL particularly effective for nuanced control in AM processes, which justifies its growing adoption alongside traditional methods. 

To the best of our knowledge, \cite{Shen_Shang_Zhao_Dong_Xiong_Wang2019_error_compensation} were the first to use DL to model AM deviations. They developed a convolutional autoencoder with a specially designed cross-entropy loss function, demonstrating the model in low resolution ($32 \times 32 \times 32$) and linear deformation settings. Later, \cite{Zhao_Xiong_Shang__Wu2019nonlinear_error_prediction} verified this method in higher resolution ($64 \times 64 \times 64$) and nonlinear settings. \cite{Shen_Shang_Li__Wang2019_error_prediction_compensation} further improved the model's accuracy by modifying the architecture, using a U-Net-based encoder-decoder framework instead of a 3D CNN-based autoencoder. \cite{Zhao_Xiong___Zhu2022point_deformation_prediction} improved the previous models \citep{Shen_Shang_Zhao_Dong_Xiong_Wang2019_error_compensation, Zhao_Xiong_Shang__Wu2019nonlinear_error_prediction} by proposing a point-wise error prediction framework based on PointNet++ architecture \citep{qi2017pointnet++}. Both \cite{Shen_Shang_Zhao_Dong_Xiong_Wang2019_error_compensation} and \cite{Zhao_Xiong_Shang__Wu2019nonlinear_error_prediction} considered the 3D object as a voxel grid, whereas \cite{Zhao_Xiong___Zhu2022point_deformation_prediction} focused on point-wise deviations.

\begin{table}[htbp]
    \centering
    \tiny
    \caption{Articles applying DL for \textit{geometric shape deviation prediction and compensation}}
    \label{tab:geometric_deviation_prediction_compensation}
    \begin{tabularx}{1.0\textwidth}{p{2.2cm} p{1.9cm} p{2.7cm} p{1.0cm} p{2.3cm} p{2.46cm} }
        \hline
        \hline
        \textbf{Applications} & \textbf{AM Processes} & \textbf{Articles} & \textbf{DL Techniques} & \textbf{Input Data} & \textbf{Evaluation Metrics}\\
        \hline
        \multirow{5}{*}{Deviation prediction} & \multirow{3}{*}{General AM} & \cite{Zhao_Xiong___Zhu2022point_deformation_prediction} & DNN & Point cloud & Euclidean distance\\
        \cline{3-6}
        & & \cite{Standfield_Wang__Kong2022_shape_deformation_prediction} & CNN & Voxel & F1 score\\
        \cline{3-6}
        & & \cite{Wang_Shraida_Jin2023predictive_cloud} & LSTM & Point cloud & $r^2$ and MSE\\
        \cline{2-6}
        & Laser-based AM & \cite{Francis_Bian2019_distortion_prediction} & CNN & Thermal image & RMSE\\
        \cline{2-6}        
        & SLM & \cite{Zhu_Ferreira_Anwar__Qiao2020} & CNN & Voxel & RMSE\\
        \hline
        \multirow{3}{2.6cm}{Deviation prediction and compensation} & \multirow{3}{*}{General AM} & \cite{Shen_Shang_Zhao_Dong_Xiong_Wang2019_error_compensation} & CNN & Voxel & F1 score\\
        \cline{3-6}        
        & & \cite{Zhao_Xiong_Shang__Wu2019nonlinear_error_prediction} & CNN & Voxel & F1 score\\
        \cline{3-6}        
        & & \cite{Shen_Shang_Li__Wang2019_error_prediction_compensation} & CNN & Pixelized 2D section & F1 score\\
        \hline        
        Shape correspondence identification & General AM & \cite{Huang_Sun_Kwok_Zhou_Xu2020} & CNN & 3D mesh & Accuracy\\
        \hline
        \hline
    \end{tabularx}
\end{table}

\cite{Standfield_Wang__Kong2022_shape_deformation_prediction} addressed many of the issues faced by \cite{Shen_Shang_Zhao_Dong_Xiong_Wang2019_error_compensation}, \cite{Zhao_Xiong_Shang__Wu2019nonlinear_error_prediction}, and \cite{Shen_Shang_Li__Wang2019_error_prediction_compensation} by (\textit{i}) breaking the 3D sample into smaller parts, (\textit{ii}) making individual predictions, and then (\textit{iii}) aggregating the results. They validated their approach on a real AM dataset, achieving higher resolution than previous research. Similarly, \cite{Wang_Shraida_Jin2023predictive_cloud} used a real AM dataset to predict z-directional geometric deviations with an encoder-decoder LSTM network and an attention mechanism to maintain spatial and temporal dependencies. Some studies approached shape deviation prediction from distinct perspectives. \cite{Francis_Bian2019_distortion_prediction} assumed that deviations result from thermal gradients in the surrounding area, using thermal images of neighboring points as input for a CNN-based prediction model. \cite{Huang_Sun_Kwok_Zhou_Xu2020} treated deviation prediction as a shape correspondence problem and proposed a CNN-based DL framework, which effectively learned and predicted the shape correspondence of deformed shapes. A brief summary of these articles employing DL to predict and/or compensate for geometric shape deviation is also provided in Table \ref{tab:geometric_deviation_prediction_compensation}.

\subsubsection{\emph{Scanning path design and optimization}}
Within the broader domain of DfAM, efficient design and optimization of the scanning toolpath, representing the sequential movements of the printing head or nozzle, are crucial for achieving optimal geometric accuracy, minimizing defects, and maximizing production throughput \citep{bhardwaj2018_importance_of_toolpath_design_SUPP, Mozaffar_Ebrahimi_Cao2020toolpath, Park_Choi_Um2024_ConvLSTM_meltpool_prediction_from_images_of_laser_tool_path_strategy_LPBF}.

\cite{Kim_Zohdi2022} used a deep CNN model to predict the optimal scanning path for the SLS process where they considered the thermal gradient of laser paths as the optimization cost function. Similarly, \cite{Ren_Chew_Liu__Bi2021_toolpath_planning} used heat accumulation as the criteria for selecting the optimal laser path in their work. Their model integrated deep RNN into FE simulation to predict the probable temperature fields of the next deposition layer for various toolpath strategies. Furthermore, \cite{Glasder_Fabbri__Wegener2023_footprint_weld_bead_prediction_using_transfer_learning_WAMM_CNN} optimized the scanning path by minimizing surface energy and predicted weld bead geometry using deep transfer learning (DTL) in the WAAM process. \cite{Zhou_Shen__Sheng2022_tool-path_planning_optimizing_thermo-mechanical_properties_WAAM_DNN_RNN_LSTM} proposed a genetic algorithm-based method to optimize continuous toolpaths in WAAM for enhanced thermomechanical properties. They utilized a rapid prediction model integrating RNN, LSTM, and DNN to efficiently evaluate toolpath fitness. Experimental results showcased substantial improvements, with over 87\% enhancement in thermomechanical properties and a reduction of over 32 MPa in average residual Von Mises stresses. \cite{Perani_Jandl_Baraldo_Valente_Paoli2023_modeling_track_geometry} introduced a LSTM neural network to model over-deposition in LMD. The model, trained on simple geometries such as straight tracks, spirals, and V-tracks, successfully generalized to more complex shapes like random tracks. Adding a small amount of data from random tracks to the training dataset further improved the model's performance, making this approach viable for broader applications in optimizing scanning strategies.

Different from these studies, \cite{Mozaffar_Ebrahimi_Cao2020toolpath} proposed a novel toolpath design framework for metal AM using model-free deep reinforcement learning (DRL) where an RL agent was asked to design a toolpath for a given 2D section sliced from a 3D object. They demonstrated that while DRL struggled with sparse reward structures, it effectively optimized toolpaths in dense reward systems. \cite{Petrik_Bambach2023_path_planning_reinforcement} introduced \textit{RLPlanner}, a reinforcement learning framework for automatic path planning in WAAM, utilizing the Proximal Policy Optimization (PPO) algorithm. Integrated with CNN and DNN, RLPlanner dynamically adjusted welding parameters, enhancing adaptability to different geometries. However, it was limited to thin-walled structures and did not support multi-bead deposition paths. \cite{Qin_Ding__Liao2024_DRL_toolpath_generation_thermal_uniformity_LPBF} developed a DRL-based toolpath generation framework specifically for L-PBF processes. Their approach aimed to achieve uniform heat distribution and reduce distortion, showcasing significant improvements over traditional scan patterns like zigzag and chessboard through numerical simulations and experimental validations.

\subsection{\emph{DL-driven AM modeling}} \label{DL_for_AM_data_driven_AM_modeling}
This section explores AM process dynamics modeling, primarily focusing on thermal-based modeling and process-structure-property (PSP) relationships in AM processes. Thermal modeling addresses processes involving thermal phenomena. However, some AM processes, such as stereolithography (SLA) and material jetting (MJ), do not rely on thermal phenomena to form parts. Our systematic review found insufficient research papers utilizing DL for non-thermal phenomena-based modeling approaches, as reflected in Table \ref{tab:big_paper_summary_list}. Consequently, we chose to focus more on thermal-based modeling and include a brief discussion on non-thermal approaches alongside it. Additionally, thermal phenomena affect PSP modeling, influencing printing patterns, grain structures, defects, stress levels, and mechanical properties of the printed part. This underscores the importance of employing DL to optimize and model the PSP relationship. Overall, DL-driven AM modeling aligns with a generalized framework, depicted in Figure \ref{fig:data_driven_AM_modeling_schematic}, illustrating the flow from input features through DL-driven modeling to output features.

\begin{figure}[htbp]
    \centering
    \captionsetup{justification=centering}
    \includegraphics[width=1.0\textwidth]{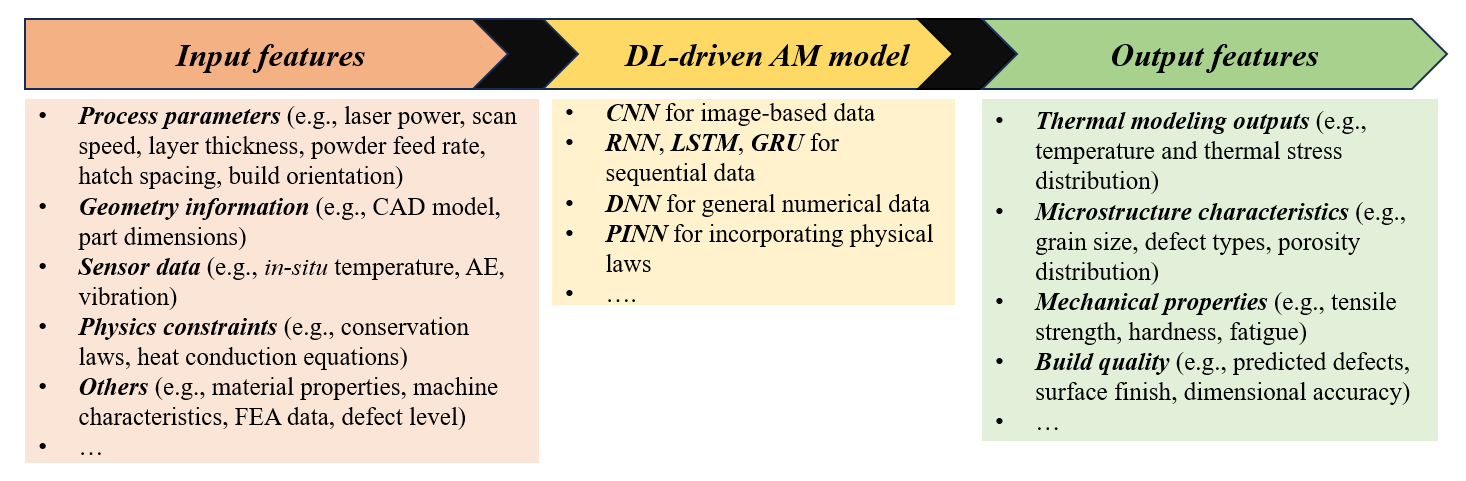}
    \caption{Generalized high-level \textit{DL-driven AM modeling framework}} \label{fig:Data_driven_AM}
\end{figure}


\textbf{Thermal profile modeling} is the process of predicting and controlling the temperature distribution and variations during the AM process. The thermal profile includes thermal gradients, heating and cooling rate, and thermal cycles, all of which significantly influence the quality and microstructure of the AM products. Data-Driven Deep Learning (DDDL) models and Physics-Informed Deep Learning (PIDL) models are commonly used techniques for thermal profile modeling, with DDDL models relying solely on data and PIDL models incorporating process physics information to enhance accuracy and reliability.

Data-driven DL (DDDL) models for thermal profile modeling involve using DL techniques to analyze patterns and relationships in AM data to model thermal profiles. For instance, \cite{Hemmasian_Ogoke__Beuth_Farimani2023} proposed a CNN-based DL model to predict thermal fields and improve parts quality. Their proposed model can take process parameters and time steps as inputs and emulate the 3D thermal geometry of a melt pool in just a few seconds compared to computational fluid dynamics (CFD) or finite element (FE). Similarly, \cite{Nalajam_Varadarajan2021} used a CNN-LSTM model to capture both spatial and temporal features for thermal forecasting in WAAM. However, the major limitation of DDDL modeling approach is the requirement for a large volume of input data to train the DL models effectively. 

\begin{table}[t]
    \centering
    \tiny
    \caption{Articles applying DL for \textit{AM thermal modeling}}
    \label{tab:thermal_profile_modeling}
    \begin{tabularx}{1.0\textwidth}{p{0.9cm} p{0.8cm} p{2.7cm} p{1.0cm} p{2.75cm} p{4.43cm}}
        \hline
        \hline
        \textbf{DL Frameworks} & \textbf{AM Processes} & \textbf{Articles} & \textbf{DL Techniques} & \textbf{Input Data} & \textbf{Process physics}\\
        \hline
        \multirow{7}{*}{DDDL} & WAAM & \cite{Nalajam_Varadarajan2021} & CNN-LSTM & Layer-wise temperature & N/A\\
        \cline{2-6}
        & SLS & \cite{Sofi_Ravani2023} & CNN & Process parameters & N/A\\
        \cline{2-6}
        & \multirow{2}{*}{L-PBF} & \cite{Hemmasian_Ogoke__Beuth_Farimani2023} & CNN & Process parameters & N/A\\
        \cline{3-6}
        & & \cite{Chen_Liu__Witherell2024_MeltpoolGAN_Melt_pool_prediction_from_path-level_thermal_history} & GAN & Thermal history, meltpool images & N/A\\
        \cline{2-6}
        & \multirow{3}{*}{DED} & \cite{Mozaffar_Paul__Cao2018} & RNN & Time series & N/A\\
        \cline{3-6}
        & & \cite{Mozaffar_Liao_Lin_Ehmann_Cao2021} & R-GNN & Time series & N/A\\
        \cline{3-6}
        & & \cite{Pham_Hoang_Tran_Pham_Fetni_Duchene_Tran_Habraken2023} & FFNN & High-fidelity data & N/A\\
        \hline
        \multirow{9}{*}{PIDL} & SLM & \cite{Ghungrad_Gould_Wolff_Haghighi2022_physics_informed_AI} & DNN & Layer-wise temperature & Heat transfer equation\\
        \cline{2-6}
        & DED & \cite{Perumal_Abueidda__Kontsos2023_TCN_data-driven_thermal_prediction_DED} & TCN, LSTM, GRU & Process parameters and geometry & Heat transfer equation\\
        \cline{2-6}
        & Metal AM & \cite{Zhu_Liu_Yan2021} & FCNN & Process parameters and geometry & Energy, momentum, and mass conservation equations\\
        \cline{2-6}
        & \multirow{2}{*}{LMD} & \cite{Ren_Chew_Zhang_Fuh_Bi2020} & RNN & Thermal field & Heat transfer equation\\
        \cline{3-6}
        &  & \cite{Guo_Guo_Bian_Guo2022} & cGAN & Melt pool thermal image & Heat transfer equation\\
        \cline{2-6}        
        & WAAM & \cite{Zhou_Shen_Liu_Du_Jin2021_Thermal_field_prediction} & RNN & Thermal field data & Tailored threshold (transition states, min. and max. temperatures)\\
        \cline{2-6}
         & \multirow{3}{*}{L-PBF} & \cite{Ghungrad_Faegh_Gould_Wolff_Haghighi2023_PIDL} & DNN & Layer-wise temperature & Heat transfer equation, density, viscosity, surface emissivity\\
        \cline{3-6}        
        & & \cite{Sharma_Raissi_Guo2023_PIDL_multi-physical_PBF} & DNN & Thermal data & Energy, momentum, and mass conservation\\
        \cline{3-6}
        & & \cite{Zhao_Wei_Mao__Liao2023_prediction_params_melt_pool_dimensions_PIDL} & DNN & Process parameters and melt pool dimensions & Mass, momentum and energy conservation, heat flux and heat transfer\\
        \hline
        \hline
    \end{tabularx}
\end{table}

Physics-informed DL (PIDL) models for thermal profile modeling involve the incorporation of process physics information in the DL models to develop more efficient techniques for predicting temperature distribution. These models integrate physical equations and domain knowledge to enhance prediction accuracy, transparency, and interpretability \citep{Guo_Tian_Guo_Guo2020, Kwon_Kim__Lee2023_PIDL_inkjet_ConvLSTM_next_image_prediction_from_sequential_images}. For instance, \cite{Zhu_Liu_Yan2021} proposed an integration of physical laws (e.g., mass, momentum, and energy conservation) with DNNs to improve the learning process in metal AM. The efficacy of the PIDL has also been demonstrated in predicting porosity via thermal images \citep{Guo_Tian_Guo_Guo2020}. \cite{Ghungrad_Gould_Wolff_Haghighi2022_physics_informed_AI} demonstrated the effectiveness of their PIDL model compared to a purely DDDL LSTM model in the SLM process for thermal prediction. Their PIDL model showed superior performance in terms of time and accuracy, even with limited data. \cite{Ghungrad_Faegh_Gould_Wolff_Haghighi2023_PIDL} further enhanced the performance of their previous PIDL model proposed in \cite{Ghungrad_Gould_Wolff_Haghighi2022_physics_informed_AI} by introducing an architecture-driven PIDL (APIDL) to predict thermal history in the L-PBF process. Despite the promise of PIDL, due to the requirement of having accurate physics-based parameters, some researchers continue to rely on advanced DDDL models (e.g., GNN) that can capture dependencies typically encompassed by physics-based models \citep{Mozaffar_Paul__Cao2018, Mozaffar_Liao_Lin_Ehmann_Cao2021, Pham_Hoang_Tran_Pham_Fetni_Duchene_Tran_Habraken2023}. Furthermore, given the computational complexity of both DDDL and PIDL models, there is an increasing trend toward employing DTL techniques to achieve more efficient modeling with smaller datasets \citep{Chen_Wong__LI2022_Bead_geometry_prediction_DED_DNN}. Table \ref{tab:thermal_profile_modeling} provides a high-level summary of thermal profile modeling using DL techniques.

As mentioned earlier, \textbf{non-thermal-based AM processes} like SLA and MJ either do not depend on thermal phenomena or are minimally influenced by them to form parts. SLA prints parts from liquid resin using photopolymerization, while MJ uses inkjet printing. These processes are utilized less frequently as they are susceptible to quality issues \citep{Khadilkar_Wang_Rai2019_stress_prediction_SLA_CNN, Segura_Li__Sun2023_Tensor_time_series_TGCN_TRNN_Material_jetting}. Presently, modeling strategies predominantly rely on experimental and simulation methodologies. However, DL models offer a promising solution for modeling the mechanical properties of such processes. Studies by \cite{Khadilkar_Wang_Rai2019_stress_prediction_SLA_CNN}, \cite{Segura_Li__Sun2023_Tensor_time_series_TGCN_TRNN_Material_jetting}, and \cite{Huang_Segura__Zhou2020_droplet_evolution_prediction_inkjet_DRNN} have demonstrated the effectiveness of DL-based modeling for predicting properties such as stress, tensile strength, and droplet evolution.

\textbf{Process-structure-property (PSP) modeling} aims to understand how AM processes influence material microstructure and properties \citep{Wang_Chandra_Huang_Tor_Tan2023}. DL enhances PSP modeling by extracting intricate patterns from large datasets, enabling accurate predictions of material behavior based on process parameters. Several studies have employed DL-based PSP modeling to predict properties in AM. For instance, \cite{Croom_Berkson_Mueller_Presley_Storck2022} used a modified U-Net to predict stress in defective porous parts produced through the L-PBF process. As it has been demonstrated that thermal properties, like cooling rate, alone are insufficient predictors of mechanical properties, DL models present a promising alternative for improving performance \citep{Fang_Cheng_Glerum_Bennett_Cao_Wagner2022, Pham_Hoang__Habraken2022_uncertainties_DED}. \cite{Zhang_Wang_Gao2018} proposed an LSTM-based tool to quantify the nonlinear relationship between the printing process and tensile strength. They further improved this approach in a subsequent study \citep{Zhang_Wang_Gao2019} to enhance tensile strength prediction for the FDM process. However, due to the multitude of features that could be used to predict tensile strength, many DL models lack transferability to different applications. Therefore, \cite{Scime_Joslin__Paquit2023_Tensile_Property_Prediction_LPBF} built a CNN-based tensile property prediction model generalizable for L-PBF parts. In another study, \cite{Herriott_Spear2020_property_prediction} performed a comparative study to assess the capability of various classical ML and CNN models for predicting microstructure properties in a metal AM process. Other notable studies that used DL for PSP modeling in general AM applications are \cite{Lu_He_Shi_Bai_Zhao_Han2021}, \cite{Tu_Liu_Carneiro__Harrison2022}, \cite{Maurizi_Gao_Berto2022_stress_strain_deformation_prediction}, \cite{Peng_Liu_Huang__Lu2022}, \cite{Koc_Zeybek_Kisasoz__Bulduk2022}, \cite{Fotovvati_Chou2022_SLM_surface_roughness_prediction_ANN}, \cite{Qin_DeWitt_Radhakrishnan_Biros2023}, \cite{Wang_Chandra_Huang_Tor_Tan2023}, and \cite{Herbeaux_Aboleinein__Klocker2024_microstructure_modeling_from_SEM_EBSD_images_CNN_WAAM_or_CMT}, among others. Although DL models offer powerful tools for PSP modeling, further investigation is still needed to achieve a reliable shift from research tools to practical engineering applications. Present limitations of DL applications to PSP modeling in AM include data size, interpretability, generalization, and transferability \citep{Castro_Pathinettampadian__Subramaniyan2023_Prediction_compressive_strength_LSTM_FDM, Rezasefat_Hogan2024_Prediction_4D_stress_evolution_SLM_CNN_voxel}.


\subsection{\emph{DL-driven process monitoring and control}} \label{DL_for_AM_monitoring_control}
\begin{figure}[htbp]
    \centering
    \captionsetup{justification=centering}
    \includegraphics[width=0.8\textwidth]{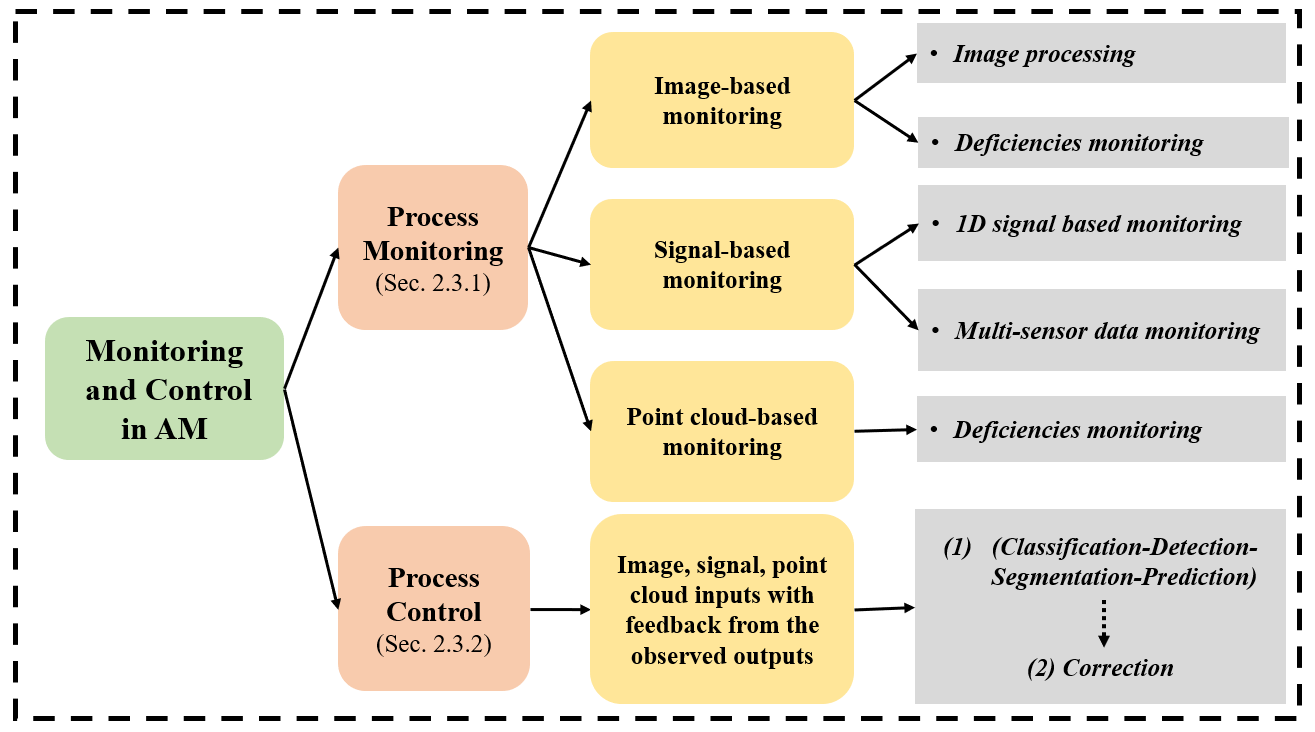}
    \caption{Proposed high-level summary of \textit{process monitoring and control} literature}
    \label{fig:process_monitoring_control_summary_figure}
\end{figure}

Process monitoring and control in AM plays a vital role in detecting and correcting irregularities during the printing process. DL techniques are increasingly utilized in this area, comprising a significant portion of the entire DL for AM literature. To facilitate a cohesive demonstration, we have proposed a high-level summary of process monitoring and control literature, as illustrated in Figure \ref{fig:process_monitoring_control_summary_figure}.

\subsubsection{\emph{AM Process monitoring}}

\subsubsection*{\emph{Image-based monitoring}}
Image-based monitoring is widely used in AM due to DL models like CNNs, which can extract spatial features from grid-like data, and the availability of high-resolution cameras. Studies in this area can be summarized into two categories: DL-based image processing and deficiency monitoring for quality assessment.

\paragraph{DL-based image processing} involves using DL techniques, such as image augmentation, segmentation, transfer learning, and feature extraction, to enhance AM process monitoring by manipulating, transforming, or extracting features from images. For instance, \cite{Kim_Yang_Ko_Cho_Lu2023} used a stacked CNN denoising autoencoder to transform melt pool images, which were then processed by pretrained CNN models with self-supervised learning. Due to the large number of AM process parameters, \cite{Tan_Fang_Li__Yang2020} and \cite{Xie_Jiang_Chen2022} focused on feature extraction instead of using pretrained models. \cite{Tan_Fang_Li__Yang2020} used a CNN-based model to segment spatter signatures, while \cite{Xie_Jiang_Chen2022} developed a data labeling framework for accurate 2D image labeling. Additionally, \cite{Ramlatchan_Li2022} proposed using a cGAN model to generate synthetic datasets rich in features customized for real-world applications. Furthermore, imaging techniques such as X-ray computed tomography (XCT) are widely used for AM monitoring due to their ability to capture internal and external features non-destructively \citep{Bellens_Probst__Dewulf2022_U_Net_porosity_segmentation_XCT_scan_SLS}. However, XCT's effectiveness is limited by slow scanning speeds and high costs \citep{Ziabari_Venkatakrishnan__Paquit2023_part_reconstruction_from_XCT_images_MAM_CNN}. To address these issues, DL-based segmentation models, primarily CNNs, have been employed for fast XCT characterization and reconstruction \citep{Li_Lambert-Garcia__Leung2024_AM-SegNet, Yang_Wang__Fang2022_UNet_XCT_image_reconstruction_prediction_compression_lattice_structures}. Additionally, DL models can enhance image resolution, augmentation, and characterization, which are critical for effective image-based monitoring in AM processes \citep{Zhang_Fu__Schleifenbaum2023_image_quality_enhancement_Unet_LPBF, Kim_Lee__Yoo2023_time_series_augmentation_StyleGAN}.

\paragraph{Deficiencies monitoring for quality assessment} involves assessing and detecting imperfections such as defects, anomalies, deformations, faults, or cracks within AM to ensure the quality of the printed products. Table \ref{tab:CNN_and_image_based_deficiencies_monitoring} provides an overview of deficiencies monitored in various AM processes, along with relevant articles and DL techniques employed for classification, detection, segmentation, and prediction. DL-based approaches, such as CNNs, have been extensively utilized to address a wide range of deficiencies, including track geometry defects, porosity, warp deformation, surface defects, powder stream faults, meltpool spatter, recoater streaking, lack of fusion, delamination, and balling. Among these, porosity is the most extensively monitored deficiency in the literature, as indicated by Table \ref{tab:CNN_and_image_based_deficiencies_monitoring}. DL techniques offer a promising solution for accurate porosity prediction models. For instance, \cite{Tian_Guo_Melder_Bian_Guo2021_data_fusion_porosity_detection} proposed a novel approach by fusing two established DL techniques through TL. The PyroNet \citep{Simonyan_Zisserman2014_VGG16} and IRNet \citep{Donahue_Hendricks__Darrell2015_LRCN}, respectively correlate pyrometer images and infrared (IR) images to porosity. Similarly, \cite{Ho_Zhang_Young__Mozumder2021}, \cite{Mao_Lin_Yu_Frye_Beckett_Anderson___Agarwal2023}, and \cite{Zamiela_Jiang__Bian2023_porosity_detection} all used thermal images for porosity prediction because thermal history can capture the relationship between the AM processes and porosity.

In anomaly detection studies, DL models are employed to classify between abnormal and normal conditions with high accuracy \citep{Kwon_Kim_Ham__Kim2020, Lee_Heogh_Yang__Lee2022_Powder_Stream_Fault}. For example, \cite{Tan_Huang_Liu_Li_Wu2022} utilized a conditional GAN to generate minority sample images for a fault group, while the domain adversarial neural network (DANN) was used to identify faulty parameters in FDM due to process drift. CNN-based models are commonly employed for image-based deficiencies monitoring in AM, such as CenterNet-CL for defect detection \citep{Wang_Cheung2022}. TL from effective networks has also been widely utilized to address AM defect detection challenges \citep{Scime_Beuth2018, Scime_Siddel_Baird_Paquit2020, Xia_Pan_Li_Chen_Li2022, Kim_Lee_Seo_Kim_Shin2023_in_Situ_monitoring, Zhu_Jiang_Guo_Xu__Jiang2023_surface_morphology_inspection_TL}. However, challenges remain in transferring models across materials, limiting the widespread generalization of DL tools \citep{Banadaki_Razaviarab_Fekrmandi_Sharifi2020, Li_Jia_Yang_Lee2020, Fischer_Zimmermann_Praetzsch_Knaak2022, Pandiyan_Cui__Shevchik2022, Mi_Zhang_Li_Shen__Mai2023, Li_Zhou_Huang_Li_Cao2023}. Additionally, other DL techniques such as federated learning (FL) \citep{Mehta_Shao2022_Federated_defect_detection}, deep belief networks (DBN) \citep{Ye_Fuh_Zhang_Hong_Zhu2018}, and spatial-temporal monitoring models like convLSTM \citep{Ko_Kim_Lu_Shin_Yang_Oh2022_spatial} and autoencoders \citep{Tan_Le2019_Efficientnet_supp_paper, Zhao_Xiong_Shang__Wu2019nonlinear_error_prediction, Larsen_Hooper2022} have also been employed for direct deficiencies monitoring in AM.

\begin{table}[htbp]
    \centering
    \tiny
    \caption{Overview of \textit{deficiencies monitored} in AM processes, with relevant articles and DL techniques} \label{tab:CNN_and_image_based_deficiencies_monitoring}
    \begin{tabularx}{1.0\textwidth}{p{2.5cm} p{1.0cm} p{5.2cm} p{4.8cm}}
        \hline
        \hline
        \textbf{Deficiencies monitored} & \textbf{AM Processes} & \textbf{Articles} & \textbf{DL Techniques (task)}\\
        \hline
        Track geometry defect & PBF & \cite{Yuan_Giera_Guss__Mcmains2019} & CNN (pixel-wise segmentation) \\
        \hline
        \multirow{12}{*}{Porosity} & \multirow{6}{*}{PBF} & \cite{Wang_Cheung2022}; \cite{Ansari_Crampton_Garrard_Cai_Atallah2022_porosity_preidction}; \cite{Zhang_Liu_Shin2019}; \cite{Mao_Lin_Yu_Frye_Beckett_Anderson___Agarwal2023}; \cite{Ho_Zhang_Young__Mozumder2021}; \cite{Oster_Breese__Altenburg2023_defect_prediction_LPBF} & CNN (classification)\\
        \cline{3-4}
         &  & \cite{Ogoke_Johnson_Glinsky__Farimani2022} & DC-GAN (generation and prediction)\\
         \cline{3-4}
         & & \cite{Desrosiers_Letenneur__Brailovski2024_porosity_segmentation_LPBF_computed_tomography_CNN} & CNN (segmentation)\\
         \cline{3-4}
         & & \cite{Mohammed_Almutahhar__Ali2023_porosity_prediction_LPBF_DNN} & DNN (prediction)\\
        \cline{3-4}
         &  & \cite{Song_Wang_Gao_Son_Wu2023_pore_prediction} & GAN (pore morphology prediction)\\
        \cline{2-4}
         & \multirow{2}{*}{DED} & \cite{Pandiyan_Cui__Shevchik2022} & CNN (classification)\\
        \cline{3-4}
        & & \cite{Kim_Oh__Kim2022_pore_detection_YOLOv5_pore_image_DED} & CNN (pore detection)\\
        \cline{2-4}
         & AM & \cite{Senanayaka_Tian_Falls_Bian2023_Porosity_prediction} & CNN (classification)\\
        \cline{2-4}
         & \multirow{2}{*}{LMD} & \cite{Guo_Tian_Guo_Guo2020}; \cite{McGowan_Gawade_Guo2022PICNN_porosity_prediction_LMD} & CNN (classification)\\
         \cline{3-4}
         & & \cite{Chen_Guo2023_DCGAN-CNN_porosity_prediction_LMD_unbalanced} & DCGAN-CNN (generation and prediction)\\
        \cline{2-4}
         & ME & \cite{Petros_Siegkas2022} & GAN (generation and classification)\\
        \hline
        \multirow{2}{*}{Warp deformation} & \multirow{2}{*}{ME} & \cite{Saluja_Xie_Fayazbakhsh2020} & CNN (classification)\\
        \cline{3-4}
         &  & \cite{Brion_Shen_Pattinson2022} & CNN (detection)\\
        \hline
        Spaghetti-shape defect & ME & \cite{Kim_Lee_Ahn2022} & CNN (classification)\\
        \hline
        \multirow{2}{*}{Weld defects} & MAM & \cite{Zhang_Wen_Chen2019_Weld_image_defect_detection_CNN_Robotic_arc_welding} & CNN (classification and detection)\\
        \cline{2-4}
        & DED & \cite{Wang_Zhang__Han2020_Weld_reinforcement_prediction_molten_pool_image_WAAM_LSTM} & LSTM (prediction)\\
        \hline
        \multirow{2}{*}{Powder bed defects} & \multirow{2}{*}{PBF} & \cite{Imani_Chen_Diewald__Yang2019_defect_detection}; \cite{Westphal_Seitz2021} & CNN (classification)\\
        \cline{3-4}
        & & \cite{Jiang_Zhang__ZHang2023_CNN_layerwise_imagesLPBF_surface_defect_classification} & CNN (detection and classification)\\
        \hline
        \multirow{7}{*}{Surface defect} & \multirow{4}{*}{DED} & \cite{Davtalab_Kazemian_Yuan_Khoshnevis2022} & CNN (pixel-wise segmentation)\\
        \cline{3-4}
        & & \cite{Zhang_Xu__Wang2023_WAAM_detect_surface_oxidation_defects_Transformer_time_series_voltage_data} & Transformer (classification)\\
        \cline{3-4}
        & & \cite{Li_Zhang__Li2023_surface_defect_detection_WAAM} & CNN (classification and detection)\\
        \cline{2-4}
        & PBF & \cite{Wang_Cheung2022} & CNN (segmentation)\\
        \cline{3-4}
         &  & \cite{Zhu_Jiang_Guo_Xu__Jiang2023_surface_morphology_inspection_TL} & CNN (detection)\\
         \cline{2-4}
         & \multirow{2}{*}{ME} & \cite{Banadaki_Razaviarab_Fekrmandi_Sharifi2020} & CNN (classification)\\
         \cline{3-4}
         & & \cite{Li_Huang__Tan_2024)_defect_classification_ViT_FDM} & Transformer (classification)\\
        \hline
        Powder stream fault & DED & \cite{Lee_Heogh_Yang__Lee2022_Powder_Stream_Fault} & CNN (classification)\\
        \hline
        \multirow{9}{*}{Meltpool spatter} & \multirow{5}{*}{DED} & \cite{Mi_Zhang_Li_Shen__Mai2023_DED_CNN_spatter_detection} & CNN (detection)\\
        \cline{3-4}
        & & \cite{Abranovic_Sarkar__Beuth2024_predicting_next_frame_meltpool_for_flaw_detection_in_video_data_DED_ConvLSTM} & ConvLSTM (detection)\\
        \cline{3-4}
        & & \cite{Liu_Yuan__Weiwei2022_meltpool_states_classification_DED_ResNet} & CNN (classification)\\
        \cline{3-4}
        & & \cite{Pandiyan_Cui__Wasmer2022_DED_meltpool_state_prediction_meltpool_image_GAN_CNN} & GAN (generation and prediction)\\
        \cline{2-4}
         & \multirow{5}{*}{PBF} & \cite{Baumgartl_Tomas_Buettner_Merkel2020} & CNN (detection)\\
         \cline{3-4}
         & & \cite{Luo_Ma__Cao2021_1DCNN_LSTM_RNN_GRU_spatter_defect_classification_SLM}; \cite{Manivannan2023_powder_bed_defect_detection_SLS_CNN}; \cite{Yang_Qiu__Bai2023_Defect_classification_LPBF_simulated_meltpool_images_and_thermal_images_CNN} & CNN (classification)\\
         \cline{3-4}
         & & \cite{Zhang_Vallabh__Zhao2022_meltpool_temperature_and_morphology_monitoring_LSTM_LPBF} & LSTM (prediction)\\
         \cline{3-4}
         & & \cite{Fathizadan_Ju__Yang2023_LPBF_ConvLSTM_Anomaly_Detection} & ConvLSTM (classification and prediction)\\
         \cline{3-4}
         & & \cite{Williams_Sing2024_Spatiotemporal_PBF_meltpool_monitoring_videos_convRNN} & ConvRNN (classification)\\
        \hline
        \multirow{2}{*}{Recoater streaking} & \multirow{2}{*}{PBF} & \cite{Scime_Beuth2018}; \cite{Mehta_Shao2022_Federated_defect_detection} & CNN (classification)\\
        \cline{3-4}
         &  & \cite{Scime_Siddel_Baird_Paquit2020} & CNN (detection)\\
        \hline
        Lack of fusion & AM & \cite{Senanayaka_Tian_Falls_Bian2023_Porosity_prediction} & CNN (classification)\\
        \hline
        Delamination & PBF & \cite{Baumgartl_Tomas_Buettner_Merkel2020} & CNN (detection)\\
        \hline
        Balling & PBF & \cite{Fischer_Zimmermann_Praetzsch_Knaak2022} & CNN (classification)\\
        \hline
        Fatigue crack & PBF & \cite{Anidjar_Lang_Mega2024)_Transfer_Learning_Detection_Fatigue_Crack_Initiation_SLM_YOLOv5} & CNN (classification)\\
        \hline
        process condition & DED & \cite{Li_Siahpour__Shi2020_CNN_DED_thermal_images} & CNN (classification)\\
        \hline
        \hline
    \end{tabularx}
\end{table}

\subsubsection*{\emph{Sensor signal-based monitoring}}

Sensors are widely used in AM process monitoring. This section categorizes the sensor signal-based monitoring into time-series signal-based monitoring and multi-sensor data-based monitoring.

\paragraph{Time-series signal-based monitoring} encompasses various types of 1D data including acoustic emissions (AE) and spectral emissions. \textit{Acoustic emissions} (AE) is a powerful waveform signal used to monitor and record transient elastic waves \citep{Pandiyan_Wrobel__Shevchik_2024_domainAdaptation_LPBF_CNN_AE, Li_Zhang__Wen2023_LPBF_defect_classification_tansfer_learning_AE_Resnet50}. This non-destructive monitoring method is particularly useful for capturing time-related defects and monitoring processes where visual inspection is not feasible. Researchers have leveraged AE inputs for defect detection, such as cracks, and porosity \citep{Shevchik_Masinelli__WAsmer2019_SCNN_quality_level_classification_PBF, Li_Cao__Zhang2023_Imbalanced_SLM_AE}. For example, \cite{Mohammadi_Mahmoud_Elbestawi2021} developed an AE-based defect detection model for L-PBF, employing DL to match signals with defects. Similarly, \cite{Ye_Hong_Zhang_Zhu_Fuh2018} utilized AE for defect detection in SLM, achieving superior performance compared to classical methods. \cite{Chen_Yao_Tan___Moon2023_pore_detection} proposed a CNN-based model for porosity and crack detection, demonstrating high prediction accuracy. Several studies have explored AE-based monitoring for the DED process as well. \cite{Hespeler_Dehghan-Niri__Halliday2022_MDPI_paper} employed a CNN-based model for quality assessment, highlighting the challenge of overfitting and model generalization. In contrast, \cite{Chen_Yao_Moon2022} used TL from two established DL models (R-CED) \citep{Park_Lee2016_supp} and (F-DNN) \citep{Liu_Smaragdis_Kim2014_supp_paper} to bypass the issues faced by \cite{Hespeler_Dehghan-Niri__Halliday2022_MDPI_paper}. All those combinations helped AE monitoring to achieve a better performance.

\textit{Spectral emissions} provide another avenue for signal-based monitoring in AM \citep{Chen_Yang__Rong2023_Physics-Informed_Attention_Network_Condition_Monitoring_WAAM}. \cite{Pandiyan_Drissi-Daoudi_Wasmer2022} and \cite{Ren_Wen_Zhang_Mazumder2022_quality_monitoring} developed quality assessment monitoring for AM via spectral-based data. \cite{Pandiyan_Drissi-Daoudi_Wasmer2022} employed DTL like \cite{Chen_Yao_Moon2022} from VGG-16 \citep{Simonyan_Zisserman2014_VGG16} and ResNet \citep{He_Zhang_Ren_Sun2016_ResNet_Supp_paper}. \cite{Ren_Wen_Zhang_Mazumder2022_quality_monitoring} used an LSTM autoencoder for quality assessment, demonstrating its effectiveness in reconstructing signals and detecting defects. Moreover, \cite{Williams_Dryburgh_Clare_Rao_Samal2018} utilized \textit{spatial resolved acoustic spectroscopy signals} within the L-PBF process to develop defect monitoring and detection using a CNN-based model, showing improvements in defect detection. Considering the potential challenges of extensive data processing and data size limitations in image-based and signal-based monitoring, various studies are integrating imaging and signals with DL models to offer faster and more efficient monitoring tools for AM processes \citep{Luo_Ma__Cao2021_1DCNN_LSTM_RNN_GRU_spatter_defect_classification_SLM, Khusheef_Shahbazi_Hashemi2023_copare_three_levels_of_fusion_FDM_LSTM_transfer_learning, Pandiyan_Cui__Leparoux2023_LDED_meltpool_image_and_AE_CNN_ViT, Kim_Chong__Shin2024_WAAM_defect_detection_CNN_transfer_learning}.

\paragraph{Multi-sensor signal-based monitoring} is essential for complex AM processes like \textit{multi-laser powder bed fusion} (M-LPBF), where multiple movable sources are involved. \cite{Voigt_Moeckel2022} conducted a benchmark study to evaluate DL models for defect detection in M-LPBF using multi-sensor data. Their approach utilized CNNs for spatial relationships and RNNs for detecting patterns like pores over time, emphasizing the significance of multi-layer classification with time series characteristics. \cite{Surana_Lynch_Nassar__Overdorff2023} proposed a novel method for Lack of fusion (LoF) flaw detection in M-LPBF using multi-spectral sensors and lasers. They integrated sensor data into rasterized images and employed a convolutional autoencoder for analysis. \cite{Pandiyan_Masinelli_Claire__Wasmer2022} focused on monitoring L-PBF using various sensors, proposing a CNN-LSTM approach to analyze signals from AE, infrared, back reflection, and visible sensors. This method enabled the detection of process zones lacking fusion, keyhole, and conduction modes. \cite{Li_Cao_Liu__LI2023_Imbalanced_data_generation_in_situ_monitoring} addressed imbalanced data issues in L-PBF monitoring by employing a GAN-based approach to generate balanced data from layer-wise images, photodiode signals, and AE signals. Their study demonstrated the superiority of multi-sensor data fusion over single-sensor data for quality monitoring in L-PBF. 

\subsubsection*{\emph{Point cloud-based monitoring}}
Point clouds, which represent objects in 3D space, provide detailed geometric information that is critical for monitoring AM process \citep{Ye_Liu__Tian_Kan2020_monitoring_point_cloud, Kaji_Nguyen-Huu_Toyserkani2022_surface_anomaly_detection_DED_point_cloud, Liu_Wang_Ho_Kong__Chase2022}. \textcolor{red}{Approaches to analyzing point cloud data generally fall into three categories: image-based, voxel-based, and direct point cloud-based techniques. \textit{Image-based} approaches convert 3D point cloud data into 2D images through projection so that traditional image analysis models (e.g., CNNs) can be used for further processing.} For example, \cite{Lyu_Akhavan__Monoochehri2021_monitoring_anomaly} introduced a hybrid convolutional autoencoder to detect surface anomalies in AM by processing 2D images transformed from point clouds. Similarly, \cite{Yangue2023_online_surface_prediction} utilized 3D point clouds from FFF-printed objects for surface morphology prediction by combining a CNN encoder for spatial feature extraction with an LSTM for defect prediction. \textit{Voxel-based} approaches, on the other hand, transform point clouds of AM parts into structured 3D grids that are processed using 3D CNNs \citep{George2024_voxelization_refererence_MDPI}. However, both approaches involve preprocessing steps, projection in image-based methods and voxelization in voxel-based methods, that often lead to information loss and increased computational overhead. \textcolor{red}{\textit{Direct point cloud-based} approaches, such as PointNet \citep{qi2017_pointnet_supp}, PointNet++ \citep{qi2017pointnet++}, Point transformer \citep{Zhao2021_Point_Transformer_ref_paper}, and DGCNN \citep{Wang_Sun__Solomon2019_DGCNN_ref}, bypass these preprocessing steps and process directly on unstructured point cloud data.} For instance, \cite{Wang_Sun_Jin_Kong_Yue2022_MVGCN_defect_identification} proposed a graph convolutional network for surface defect monitoring using raw point cloud data. Although their case study focused on concrete surfaces, this approach is generalized for surface defect detection using raw point clouds, including those of AM parts. This exemplifies the advantages of direct point cloud-based monitoring for AM applications.

\subsubsection{\emph{AM process control}}

Ensuring consistent quality in final AM parts is always challenging due to the dynamic and complex nature of the process. Through DL techniques, researchers have developed models that control parameters such as welding current, print speed, and nozzle height by closely monitoring outcomes like welding width, layer deposition, and print quality. Table \ref{tab:AM_process_control} provides a brief overview of relevant DL applications in AM process control, highlighting the controlled inputs and observed outputs across different AM processes.

\begin{table}[htbp]
    \centering
    \tiny
    \caption{Articles applying DL techniques in AM \textit{process control}}
    \label{tab:AM_process_control}
    \begin{tabularx}{1.0\textwidth}{p{3cm} p{3.3cm} p{1.8cm} p{2.9cm} p{2.0cm}}
        \hline
        \hline
        \textbf{Controlled inputs} & \textbf{Observed outputs} & \textbf{AM Processes} & \textbf{Articles} & \textbf{DL Techniques}\\
        \hline
        Welding current &  welding width and reinforcement of the deposited layer & WAAM & \cite{Wang_Xu__Yao2021_deposited_layer_width_and_reinforcement_prediction_WAAM} & CNN (ResNet-34)\\
        \hline
        Welding current & welding width & WAAM & \cite{Wang_Lu_Zhao_Deng__Yao2021} & CNN\\
        \hline
        Print speed, feed rate & layer deposition & FDM & \cite{Akhavan_Lyu_Manoochehri2024} & CNN\\
        \hline
        Print speed, flow rate, nozzle height & over/under extrusion quality & FDM & \cite{Jin_Zhang_Gu2019} & CNN\\
        \hline
        Signal to turn off printing & print quality level & FDM & \cite{Saluja_Xie_Fayazbakhsh2020} & CNN\\
        \hline
        Feedback signal from the process video & flow rate & FDM & \cite{Brion_Pattinson2022_quantitative_real_time_control} & CNN\\
        \hline
        Bed temperature, fan speed, lateral speed & warp deformation lavel & ME & \cite{Brion_Shen_Pattinson2022} & CNN\\
        \hline
        Reference profile geometry & predicted profile & Jet-based AM & \cite{Inyang-Udoh_Chen_Mishra2022} & RNN\\
        \hline
        Reference profile geometry and process parameters & predicted profile & WAAM & \cite{Biehler_Shi2024retrofit} & point-wise MLP\\
        \hline
        \hline
    \end{tabularx}
\end{table}

CNN-based models have been utilized in several process control applications for detection and correction via image input data analysis \citep{Wang_Lu_Zhao_Deng__Yao2021}. \cite{Brion_Pattinson2022_quantitative_real_time_control} used 3D metadata and a real-time video to produce labeled images fed to a RegNet DL model \citep{Radosavovic_Kosaraju_Girshick_He_Dollar2020_designing_network_design_spaces}, enabling the prediction of material flow rate alteration in the FDM process. The model can therefore correct errors in due time. Addressing the lack of correction tools, \cite{Brion_Shen_Pattinson2022} proposed a warp detection and correction model for FDM via TL. \cite{Akhavan_Lyu_Manoochehri2024} developed a closed-loop system for quality assessment and control of AM using a hybrid convolutional autoencoder-decoder model to statistically characterize top surface roughness. Furthermore, \cite{Inyang-Udoh_Chen_Mishra2022} developed a ConvRNN model for predicting part height evolution, building on previous predictive control work \citep{Inyang-Udoh_Guo_Peters_Oomen_Mishra2020_layer_to_layer_predictive_control, Inyang-Udoh_Mishra2021_PINN}. Recently, \cite{Biehler_Shi2024retrofit} introduced \textit{RETROFIT}, a 3D profile-based control framework for enhancing shape accuracy in WAAM, employing PointNet architecture \citep{qi2017_pointnet_supp} for point-wise feature extraction from 3D point cloud data, and leveraging \textit{Koopman operator theory} for model-based control.

Reinforcement learning (RL) and deep learning (DL) have also been used for process control in AM \citep{dharmawan2020model_control_reinforcement, Piovarci2022closedLoop_Control_reinforcement, Chung2022_Reinforcement, Yue_Chen_LI_Yin2023_control_droplet_volume_inkjet_printing_DRL}. In a study by \cite{Ogoke_Farimani2021}, deep reinforcement learning (DRL) was employed to control thermal features in the L-PBF process. They proposed a policy-based model where the agent learns parameters to control the environment. Another significant application was highlighted in the work by \cite{Li_Segura__Sun2023_Droplet_Pinch-Off_Behaviors_Identification_Inkjet_DRL_GCN}, which focused on optimizing droplet pinch-off behaviors in the inkjet printing AM process to improve product quality. By using a Graph Convolutional Network for feature extraction from droplet images, and a \textit{Multiclass Reinforced Active Learning} framework for dynamic annotation, the system effectively classified pinch-off behaviors with minimal manual effort.


\section{Challenges and future directions} \label{challenges_future_directions}

\subsection{\emph{Generalizing the AM-oriented DL model to better handle the complex AM part geometry}}
Current literature often focuses on simple geometric shapes (e.g., cubic, cylindrical) as a proof of concept to demonstrate the efficacy of proposed DL methods. To advance DL techniques for complex and customized AM products, it is necessary to improve model generalization for a wider range of geometries. Future research could focus on improving model transferability and sharing knowledge among models to build geometry-invariant DL models for AM applications.
\begin{enumerate}
    
    \item \textbf{Enhancing model transferability:} TL enhances transferability by first training a model on a wide range of simple geometric shapes and then fine-tuning it for more complex and customized products \citep{Tang_Dehaghani_Wang_2023_Review_transfer_learning}. (\textit{i}) Most of the current TL applications for AM overlook the \textit{relevance between source and target domains,} relying on \emph{qualitative} measures for domain similarity assessment. Therefore, future research could focus on developing a generalized framework to \emph{quantitatively} evaluate the similarity among different AM domains. \textit{ii}) \textit{Integrating domain-specific AM process knowledge into the transferable information} might also help in enhancing the transferability of models, since traditional TL for AM mostly focuses on the transfer of information as a form of data, model parameters, and weights. (\textit{iii}) Developing \textit{novel domain adaptation methods} might also be considered to transition from simple geometric shapes to complex ones. Including non-black-box methods such as statistical modeling \citep{liu2019layer} and domain-aware approaches \citep{bappy2022morphological} would also be beneficial. Although this section primarily emphasizes the potential improvement of TL in AM applications to generalize DL models for complex geometries, there is increasing attention to TL across the literature for various AM applications. Therefore, we include a high-level summary of articles utilizing deep TL for AM, as listed in Table \ref{tab:transer_learning}. These papers have been previously discussed in relevant sections.

\begin{table}[htbp]
    \centering
    \caption{Articles utilizing \textit{transfer learning} with various pretrained networks for diverse AM applications} \label{tab:transer_learning}
    \tiny
    \begin{tabularx}{1.0\textwidth}{p{3.8cm}| p{10.53cm}}
        \hline
        \hline
        \textbf{Pretrained Networks (authors)} & \textbf{Articles using transfer learning for AM applications (task)} \\
        \hline
        ResNet \citep{He_Zhang_Ren_Sun2016_ResNet_Supp_paper} & \cite{Xia_Pan_Li_Chen_Li2022} (defect detection); \cite{Pandiyan_Drissi-Daoudi_Wasmer2022} (build quality classification across materials); \cite{Kim_Yang_Ko_Cho_Lu2023} (melt pool image monitoring); \cite{Xie_Jiang_Chen2022} (process zone image segmentation)\\
        \hline
        VGGNet \citep{Simonyan_Zisserman2014_VGG16} & \citep{Westphal_Seitz2021, Xia_Pan_Li_Chen_Li2022, Kim_Lee_Ahn2022} (defect detection); \cite{Li_Zhou_Huang_Li_Cao2023} (quality assessment through layer-wise image classification); \cite{Pandiyan_Drissi-Daoudi_Wasmer2022} (build quality classification across materials); \cite{Kim_Yang_Ko_Cho_Lu2023} (melt pool image monitoring)\\
        \hline
        Xception \citep{Chollet2017_Xception} & \cite{Westphal_Seitz2021} (defect detection); \cite{Fischer_Zimmermann_Praetzsch_Knaak2022} (powder bed quality monitoring)\\
        \hline
        EfficientNet \citep{Tan_Le2019_Efficientnet_supp_paper} & \cite{Xia_Pan_Li_Chen_Li2022} (defect detection)\\
        \hline
        GoogLeNet \citep{Szegedy_Liu_Jia__Rabinovich2015_GoogLeNet_Inception} & \citep{Xia_Pan_Li_Chen_Li2022, Kim_Lee_Ahn2022} (defect detection)\\
        \hline
        ErfNet \citep{Romera_Alvarez_Bergasa_Arroyo2017_ERFNet} & \cite{Wang_Lu_Zhao_Deng__Yao2021} (process control) \\
        \hline
        MobileNet \citep{Howard_Zhu_Chen__Adam2017Mobilenets} & \cite{Xie_Jiang_Chen2022} (process zone image segmentation)\\
        \hline
        AlexNet \citep{Krizhevsky_Sutskever_Hinton2012_ImageNet_Classification_with_DCNN} & \cite{Scime_Beuth2018} (defect detection); \cite{Imani_Chen_Diewald__Yang2019_defect_detection} (powder bed image classification)\\
        \hline
        YOLO \citep{Redmon_Divvala_Girshick_Farhadi2016_YOLO_supp_paper} & \cite{Zhu_Jiang_Guo_Xu__Jiang2023_surface_morphology_inspection_TL} (surface morphology inspection); \cite{Anidjar_Lang_Mega2024)_Transfer_Learning_Detection_Fatigue_Crack_Initiation_SLM_YOLOv5} (localization of fatigue crack initiation); \cite{Wang_Wang__Li2023_suface_Defect_Detection_AM_YOLOv8} (defect classification); \cite{Li_Zhang__Li2023_surface_defect_detection_WAAM} (surface defect detection); \cite{Kim_Oh__Kim2022_pore_detection_YOLOv5_pore_image_DED} (pore detection from images); \cite{Brion_Shen_Pattinson2022} (warp deformation detection)\\
        \hline
        \hline
    \end{tabularx}
\end{table}
    
    \item \textbf{Facilitating knowledge sharing among DL models:} Encouraging knowledge sharing among DL models, using techniques like ensemble, federated, and incremental learning, improves their capability in handling complex AM geometries. \textit{Ensemble models} combine multiple DL models, each focusing on specific features of part geometry (e.g., internal structure, layered structure, fillets and chamfers, overhangs and undercuts, surface textures), improving understanding of complex customized shapes. \textit{Federated learning (FL)} allows the model to learn from different data sources and, as a consequence, be robust to highly customized geometries. \cite{Mehta_Shao2022_Federated_defect_detection} followed an FL approach for pixel-wise defect detection in AM for only single class defect; their work might be considered as a starting point for an extension to multiple defect types. \textit{Incremental transfer learning} with new AM data might be explored to gradually improve the DL model by incorporating time-dependent geometric shapes and varied complexities from the new unseen data and at the same time sharing knowledge from the already trained model. 

\end{enumerate}

\subsection{\emph{Data landscape: availability and quality of AM data}}
DL models usually require a large amount of data. However, data generated from AM are not always readily available due to the issues of availability and quality.

\begin{enumerate}
    \item \textbf{AM Data availability:} (\textit{i}) \textit{Imbalanced data:} Imbalanced data is a common issue in AM processes, where producing defective parts is less frequent than defect-free ones. Current literature, such as \cite{Chen_Guo2023_DCGAN-CNN_porosity_prediction_LMD_unbalanced}, \cite{Tan_Huang_Liu_Li_Wu2022}, and \cite{Kim_Lee__Yoo2023_time_series_augmentation_StyleGAN}, have utilized GAN-based models to augment minority class samples to tackle the problem of class imbalance. There is room for further improvement to enhance the performance of these models. Recently, \cite{Jihoon_chung_Shen_Kong2023_three_player_GAN} proposed a DL-based three-player GAN to solve the issue of class imbalance which shows better performance than most of the existing methods. Their method was demonstrated using both AM surface defect data \citep{Jihoon_chung_Shen_Kong2023_three_player_GAN} and generalized spectral sensor data \citep{chung_saimon2024imbalanced_glyco_data}. (\textit{ii}) \textit{Incomplete data}: Due to technical difficulties or sensor failures, it is possible that some of the sensor readings may be missing in a specific crucial time interval during AM process monitoring which might affect the decisions of process optimization. In these scenarios, DL-based data imputation methods, such as autoencoders or generative models, could be explored to find patterns from the existing sensor measurements and intelligently fill in the missing measurements \citep{Li_Shi_Liu2023_data_imputation}. Again, to address the incomplete data issues, \cite{Williams_Meisel_Simpson_McComb2019design_repository_manufacturability} initialized the effort of utilizing readily-available online design repositories (e.g., GrabCAD, Thingiverse) which could be a good starting point for further exploration. 
        
    \item \textbf{AM Data quality:} (\textit{i}) \textit{Noisy data}: Noise can arise in AM data from several sources, such as environmental factors and sensor inaccuracies \citep{Li_Jia_Yang_Lee2020, Davtalab_Kazemian_Yuan_Khoshnevis2022}. The DL model should be checked for robustness to noisy data. The test performance on the synthetically generated noisy datasets might be an indicator of model robustness. Regularization techniques could also be explored to make the model more robust. The robust model remains less sensitive to minor variations arising from sensor-related difficulties in the AM data \citep{Li_Jia_Yang_Lee2020}. (\textit{ii}) \textit{Low accuracy data}: If sensors are not well-calibrated then it is possible to find less accurate data which are anomalies or outliers. Outliers might lead to inaccurate prediction of the DL model. To exclude outliers from the dataset, novel DL-based outlier detection tools are required. DL-based anomaly detection models should be integrated with the AM process in such a way that it can detect anomalies in real-time and help take appropriate feedback actions. (\textit{iii}) \textit{Data security}: AM data, both process and design, might be confidential or partially confidential to share across the manufacturing industries \citep{Shi_Kan_Tian_Liu2021_Blockchain_security, fullington2023design}. On the other hand, DL models are data-hungry since more training data helps the model be more confident on the unseen data. Therefore, deep FL can be an effective solution to use data from many sources or clients without sharing mutual information among themselves to ensure data security and at the same time utilize the collaborative understanding to train the model. Additionally, \cite{Mehta_Shao2022_Federated_defect_detection} suggested exploring various data protection approaches including blockchain technology, differential privacy, secure multiparty computation, and homomorphic encryption to enhance the security of AM client data. (\textit{iv}) \textit{Data heterogeneity:} AM processes involve multiple sources of data including temperature sensors and camera images \citep{Khusheef_Shahbazi_Hashemi2023_copare_three_levels_of_fusion_FDM_LSTM_transfer_learning}. Sometimes it is required to integrate them for comprehensive analysis. Specific DL architectures such as multi-modal architectures can jointly analyze these different types of data. Generalized DL-based multi-modal data fusion methods can be tailored for specific data sources of the AM process, such as integrating \textit{in-situ} and \textit{ex-situ} images for porosity detection in metal AM \citep{Zamiela_Jiang__Bian2023_porosity_detection}. However, these types of works still suffer from adjusting complex geometries and sensor placement-related issues, which might be a direction of future research. 
\end{enumerate}

\subsection{\emph{Uncertainty management in DL for AM applications}}

\begin{figure}
    \centering
    \captionsetup{justification=centering}
   \includegraphics[width=1.0\textwidth]{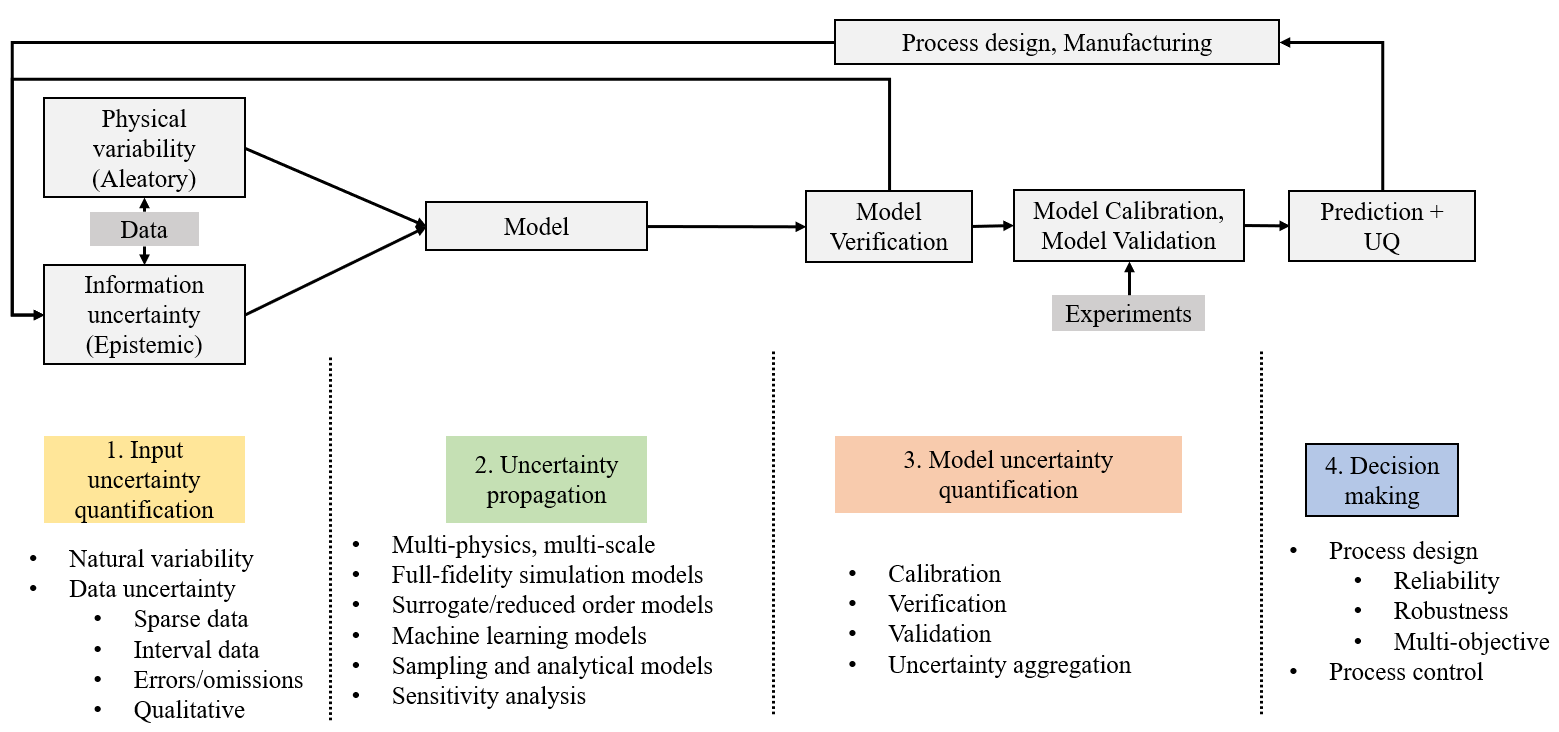}
   \vspace{-10pt}
   \caption{\textit{Uncertainty quantification flowchart in AM} process, reproduced from \cite{Mahadevan_Nath_Hu2022_UQ_AM_Review}}
   \label{fig:UQ_flowchart_Maha2022}
   \vspace{-10pt}
\end{figure}

Uncertainty is everywhere and being a practical process AM is more prone to uncertainty \citep{Pham_Hoang__Habraken2022_uncertainties_DED, Pandita_Ghosh_Gupta__Wang2022_process_modeling}. AM process-oriented uncertainty can be involved in DL models in different formats including input data uncertainty and model uncertainty. If input data uncertainty cannot be quantified before feeding into the DL model, it will propagate and might mislead model outputs and interpretations. Therefore, effective uncertainty quantification (UQ) in the input data and propagation through DL models specific to AM applications might be explored. A general UQ flowchart tailored for AM is shown in Figure \ref{fig:UQ_flowchart_Maha2022} which was proposed by \cite{Mahadevan_Nath_Hu2022_UQ_AM_Review}. \cite{Mahadevan_Nath_Hu2022_UQ_AM_Review} also reviewed the literature on UQ for the improvement of AM process, however, it was a generalized review of UQ for AM where DL was not the primary focus. In contrast, \cite{Abdar_Pourpanah__Nahavandi2021_UQ_DL_Review} provided a comprehensive survey of existing methods to handle uncertainty in DL models, but it did not focus on the AM process. Therefore, these two review papers \citep{Abdar_Pourpanah__Nahavandi2021_UQ_DL_Review, Mahadevan_Nath_Hu2022_UQ_AM_Review} provided strong support to the UQ for DL research in AM applications.

\subsection{\emph{Balancing model complexity and interpretability}}

Finding an appropriate balance between model complexity and interpretability is one of the challenges in the implementation of DL for AM. This section highlights some of the potential \textit{issues related to highly complex DL models}, and \textit{techniques to reduce model complexity}. Finally, it focuses on the potential of \textit{leveraging interpretable DL in AM}.

Highly complex DL models are naturally related to some of the inherent challenges of overfitting, requirements of large amounts of AM data, and computational resources. However, as discussed earlier, accurate and noise-free data is still both expensive and time-consuming in AM \citep{Zhang_Safdar_Xie_Li_Sage_Zhao2022_review}. Furthermore, complex models with larger datasets require high computational resources. All of these challenges need to be kept in mind during the implementation of DL for AM applications. \textit{Data dimension reduction} and \textit{model hyper-parameter tuning} techniques can be utilized to reduce the complexity of the DL models. Dimension reduction techniques could simplify the raw data and, thereby, help the model to capture informative features \citep{liu2021integrated, shi2022hybrid}. Regarding hyper-parameter tuning, it is worth mentioning that most of the existing DL implementations for AM follow random or trial-and-error basis search. It would be a good research direction to explore advanced optimization-based hyper-parameter tuning methods tailored to AM.

It is still difficult to clearly interpret the prediction results of the DL models due to the over-parameterized complex black-box nature. Moreover, \emph{no significant studies} have been noticed in the literature on interpretable DL dedicated to AM applications. Therefore, interpretable DL can be a potential research direction in the field of AM. It can contribute by making DL models' decisions easily interpretable for the AM community, identifying critical features responsible for defective parts, enhancing trust in the capacity of DL models, and eventually helping in the widespread adoption of DL in AM industries. A recent survey paper on generalized interpretable DL by \cite{Li_Xiong__Dou2022_interpretable_DL_review} could be a starting point for further potential exploration tailored to AM applications.

\section{Conclusion} \label{conclusion}
This paper provides a comprehensive review of the studies focused on the implementation of DL, rather than classical ML, to improve the performance of AM, showing the most recent trends at this intersection. The studies were gathered using a step-by-step PRISMA methodology, and the review was conducted using a proposed methodological framework based on three defined research questions. The review starts with the identification of the frequent research problems in the AM domain and then refers to the capacity of DL to address these issues. Following this, a comprehensive review of the studies applying DL in different aspects of the AM life cycle is provided which includes design for AM, both pure data-driven and physics-informed AM process modeling, and AM process monitoring and control. The findings and future guidelines are also listed generously.

This work also identifies that leveraging DL to advance AM still requires significant attention because of (\textit{i}) the rapid individual growth of DL and AM domains, (\textit{ii}) the availability of an enormous amount of AM data as a result of the development of sensor data acquisition and fusion techniques, and (\textit{iii}) the advancement of computational resources.

\bibliographystyle{chicago}
\spacingset{1}
\bibliography{IISE-Trans}

\begin{thebibliography}{}

\bibitem[\protect\citeauthoryear{Abdar, Pourpanah, Hussain, Rezazadegan, Liu, Ghavamzadeh, Fieguth, Cao, Khosravi, Acharya, et~al.}{Abdar et~al.}{2021}]{Abdar_Pourpanah__Nahavandi2021_UQ_DL_Review}
Abdar, M., F.~Pourpanah, S.~Hussain, D.~Rezazadegan, L.~Liu, M.~Ghavamzadeh, P.~Fieguth, X.~Cao, A.~Khosravi, U.~R. Acharya, et~al. (2021).
\newblock A review of uncertainty quantification in deep learning: Techniques, applications and challenges.
\newblock {\em Information fusion\/}~{\em 76}, 243--297.

\bibitem[\protect\citeauthoryear{Abranovic, Sarkar, Chang-Davidson, and Beuth}{Abranovic et~al.}{2024}]{Abranovic_Sarkar__Beuth2024_predicting_next_frame_meltpool_for_flaw_detection_in_video_data_DED_ConvLSTM}
Abranovic, B., S.~Sarkar, E.~Chang-Davidson, and J.~Beuth (2024).
\newblock Melt pool level flaw detection in laser hot wire directed energy deposition using a convolutional long short-term memory autoencoder.
\newblock {\em Additive Manufacturing\/}~{\em 79}, 103843.

\bibitem[\protect\citeauthoryear{Akhavan, Lyu, and Manoochehri}{Akhavan et~al.}{2024}]{Akhavan_Lyu_Manoochehri2024}
Akhavan, J., J.~Lyu, and S.~Manoochehri (2024).
\newblock A deep learning solution for real-time quality assessment and control in additive manufacturing using point cloud data.
\newblock {\em Journal of Intelligent Manufacturing\/}~{\em 35\/}(3), 1389--1406.

\bibitem[\protect\citeauthoryear{Almasri, Bettebghor, Adjed, Danglade, and Ababsa}{Almasri et~al.}{2024}]{Almasri_Bettebghor__Ababsa2024_generation_mechanical_designs_ConvGANs}
Almasri, W., D.~Bettebghor, F.~Adjed, F.~Danglade, and F.~Ababsa (2024).
\newblock Geometrically-driven generation of mechanical designs through deep convolutional gans.
\newblock {\em Engineering Optimization\/}~{\em 56\/}(1), 18--35.

\bibitem[\protect\citeauthoryear{Almasri, Danglade, Bettebghor, Adjed, and Ababsa}{Almasri et~al.}{2022}]{Almasri_Danglade__Ababsa2022}
Almasri, W., F.~Danglade, D.~Bettebghor, F.~Adjed, and F.~Ababsa (2022).
\newblock Deep learning for additive manufacturing-driven topology optimization.
\newblock {\em Procedia CIRP\/}~{\em 109}, 49--54.

\bibitem[\protect\citeauthoryear{Almasri, Danglade, Bettebghor, Adjed, and Ababsa}{Almasri et~al.}{2023}]{Almasri_Danglade__Ababsa_2023}
Almasri, W., F.~Danglade, D.~Bettebghor, F.~Adjed, and F.~Ababsa (2023).
\newblock A data-driven topology optimization approach to handle geometrical manufacturing constraints in the earlier steps of the design phase.
\newblock {\em Procedia CIRP\/}~{\em 119}, 377--383.

\bibitem[\protect\citeauthoryear{Anidjar, Mega, et~al.}{Anidjar et~al.}{2024}]{Anidjar_Lang_Mega2024)_Transfer_Learning_Detection_Fatigue_Crack_Initiation_SLM_YOLOv5}
Anidjar, O.~H., M.~Mega, et~al. (2024).
\newblock Transfer learning methods for fractographic detection of fatigue crack initiation in additive manufacturing.
\newblock {\em IEEE Access\/}.

\bibitem[\protect\citeauthoryear{Ansari, Crampton, Garrard, Cai, and Attallah}{Ansari et~al.}{2022}]{Ansari_Crampton_Garrard_Cai_Atallah2022_porosity_preidction}
Ansari, M.~A., A.~Crampton, R.~Garrard, B.~Cai, and M.~Attallah (2022).
\newblock A convolutional neural network (cnn) classification to identify the presence of pores in powder bed fusion images.
\newblock {\em The International Journal of Advanced Manufacturing Technology\/}~{\em 120\/}(7-8), 5133--5150.

\bibitem[\protect\citeauthoryear{Banadaki, Razaviarab, Fekrmandi, and Sharifi}{Banadaki et~al.}{2020}]{Banadaki_Razaviarab_Fekrmandi_Sharifi2020}
Banadaki, Y., N.~Razaviarab, H.~Fekrmandi, and S.~Sharifi (2020).
\newblock Toward enabling a reliable quality monitoring system for additive manufacturing process using deep convolutional neural networks.
\newblock {\em arXiv preprint arXiv:2003.08749\/}.

\bibitem[\protect\citeauthoryear{Bappy, Liu, Bian, and Tian}{Bappy et~al.}{2022}]{bappy2022morphological}
Bappy, M.~M., C.~Liu, L.~Bian, and W.~Tian (2022).
\newblock Morphological dynamics-based anomaly detection towards in situ layer-wise certification for directed energy deposition processes.
\newblock {\em Journal of Manufacturing Science and Engineering\/}~{\em 144\/}(11), 111007.

\bibitem[\protect\citeauthoryear{Baumgartl, Tomas, Buettner, and Merkel}{Baumgartl et~al.}{2020}]{Baumgartl_Tomas_Buettner_Merkel2020}
Baumgartl, H., J.~Tomas, R.~Buettner, and M.~Merkel (2020).
\newblock A deep learning-based model for defect detection in laser-powder bed fusion using in-situ thermographic monitoring.
\newblock {\em Progress in Additive Manufacturing\/}~{\em 5\/}(3), 277--285.

\bibitem[\protect\citeauthoryear{Bellens, Probst, Janssens, Vandewalle, and Dewulf}{Bellens et~al.}{2022}]{Bellens_Probst__Dewulf2022_U_Net_porosity_segmentation_XCT_scan_SLS}
Bellens, S., G.~M. Probst, M.~Janssens, P.~Vandewalle, and W.~Dewulf (2022).
\newblock Evaluating conventional and deep learning segmentation for fast x-ray ct porosity measurements of polymer laser sintered am parts.
\newblock {\em Polymer Testing\/}~{\em 110}, 107540.

\bibitem[\protect\citeauthoryear{Bhardwaj and Shukla}{Bhardwaj and Shukla}{2018}]{bhardwaj2018_importance_of_toolpath_design_SUPP}
Bhardwaj, T. and M.~Shukla (2018).
\newblock Effect of laser scanning strategies on texture, physical and mechanical properties of laser sintered maraging steel.
\newblock {\em Materials Science and Engineering: A\/}~{\em 734}, 102--109.

\bibitem[\protect\citeauthoryear{Biehler and Shi}{Biehler and Shi}{2024}]{Biehler_Shi2024retrofit}
Biehler, M. and J.~Shi (2024).
\newblock Retrofit: Real-time control of time-dependent 3d point cloud profiles.
\newblock {\em Journal of Manufacturing Science and Engineering\/}~{\em 146\/}(6).

\bibitem[\protect\citeauthoryear{Brion and Pattinson}{Brion and Pattinson}{2022}]{Brion_Pattinson2022_quantitative_real_time_control}
Brion, D.~A. and S.~W. Pattinson (2022).
\newblock Quantitative and real-time control of 3d printing material flow through deep learning.
\newblock {\em Advanced Intelligent Systems\/}~{\em 4\/}(11), 2200153.

\bibitem[\protect\citeauthoryear{Brion, Shen, and Pattinson}{Brion et~al.}{2022}]{Brion_Shen_Pattinson2022}
Brion, D.~A., M.~Shen, and S.~W. Pattinson (2022).
\newblock Automated recognition and correction of warp deformation in extrusion additive manufacturing.
\newblock {\em Additive Manufacturing\/}~{\em 56}, 102838.

\bibitem[\protect\citeauthoryear{Castro, Pathinettampadian, Ravi, and Subramaniyan}{Castro et~al.}{2023}]{Castro_Pathinettampadian__Subramaniyan2023_Prediction_compressive_strength_LSTM_FDM}
Castro, P., G.~Pathinettampadian, C.~S.~D. Ravi, and M.~K. Subramaniyan (2023).
\newblock Prediction of compressive strength in additively fabricated part using long short term memory based neural network.
\newblock {\em Materials Today Communications\/}~{\em 37}, 107139.

\bibitem[\protect\citeauthoryear{Chen, Wong, Raghavan, and Li}{Chen et~al.}{2022}]{Chen_Wong__LI2022_Bead_geometry_prediction_DED_DNN}
Chen, C., S.~J.~L. Wong, S.~Raghavan, and H.~Li (2022).
\newblock Design of experiments informed deep learning for modeling of directed energy deposition process with a small-size experimental dataset.
\newblock {\em Materials \& Design\/}~{\em 222}, 111098.

\bibitem[\protect\citeauthoryear{Chen, Liu, Liu, and Witherell}{Chen et~al.}{2024}]{Chen_Liu__Witherell2024_MeltpoolGAN_Melt_pool_prediction_from_path-level_thermal_history}
Chen, H., X.~Liu, X.~Liu, and P.~Witherell (2024).
\newblock Meltpoolgan: Melt pool prediction from path-level thermal history.
\newblock {\em Additive Manufacturing\/}~{\em 84}, 104095.

\bibitem[\protect\citeauthoryear{Chen, Yang, Wang, Zhang, Diao, and Rong}{Chen et~al.}{2023}]{Chen_Yang__Rong2023_Physics-Informed_Attention_Network_Condition_Monitoring_WAAM}
Chen, L., F.~Yang, R.~Wang, Y.~Zhang, Z.~Diao, and M.~Rong (2023).
\newblock Optical spectral physics-informed attention network for condition monitoring in waam.
\newblock {\em IEEE Transactions on Industrial Electronics\/}.

\bibitem[\protect\citeauthoryear{Chen, Yao, Feng, Chew, and Moon}{Chen et~al.}{2023}]{Chen_Yao_Feng_Chew_Moon2023_multimodal_fusion_in_situ_defect_detection}
Chen, L., X.~Yao, W.~Feng, Y.~Chew, and S.~K. Moon (2023).
\newblock Multimodal sensor fusion for real-time location-dependent defect detection in laser-directed energy deposition.
\newblock In {\em International Design Engineering Technical Conferences and Computers and Information in Engineering Conference}, Volume 87295, pp.\  V002T02A069. American Society of Mechanical Engineers.

\bibitem[\protect\citeauthoryear{Chen, Yao, and Moon}{Chen et~al.}{2022}]{Chen_Yao_Moon2022}
Chen, L., X.~Yao, and S.~K. Moon (2022).
\newblock In-situ acoustic monitoring of direct energy deposition process with deep learning-assisted signal denoising.
\newblock {\em Materials Today: Proceedings\/}~{\em 70}, 136--142.

\bibitem[\protect\citeauthoryear{Chen, Yao, Tan, He, Su, Weng, Chew, Ng, and Moon}{Chen et~al.}{2023}]{Chen_Yao_Tan___Moon2023_pore_detection}
Chen, L., X.~Yao, C.~Tan, W.~He, J.~Su, F.~Weng, Y.~Chew, N.~P.~H. Ng, and S.~K. Moon (2023).
\newblock In-situ crack and keyhole pore detection in laser directed energy deposition through acoustic signal and deep learning.
\newblock {\em Additive Manufacturing\/}~{\em 69}, 103547.

\bibitem[\protect\citeauthoryear{Chen et~al.}{Chen et~al.}{2023}]{Chen_Guo2023_DCGAN-CNN_porosity_prediction_LMD_unbalanced}
Chen, M. et~al. (2023).
\newblock Dcgan-cnn with physical constraints for porosity prediction in laser metal deposition with unbalanced data.
\newblock {\em Manufacturing Letters\/}~{\em 35}, 1146--1154.

\bibitem[\protect\citeauthoryear{Chinchanikar and Shaikh}{Chinchanikar and Shaikh}{2022}]{Chinchanikar_Shaikh_2022_review}
Chinchanikar, S. and A.~A. Shaikh (2022).
\newblock A review on machine learning, big data analytics, and design for additive manufacturing for aerospace applications.
\newblock {\em Journal of Materials Engineering and Performance\/}~{\em 31\/}(8), 6112--6130.

\bibitem[\protect\citeauthoryear{Chollet}{Chollet}{2017}]{Chollet2017_Xception}
Chollet, F. (2017).
\newblock Xception: Deep learning with depthwise separable convolutions.
\newblock In {\em Proceedings of the IEEE conference on computer vision and pattern recognition}, pp.\  1251--1258.

\bibitem[\protect\citeauthoryear{Chung, Shen, and Kong}{Chung et~al.}{2023}]{Jihoon_chung_Shen_Kong2023_three_player_GAN}
Chung, J., B.~Shen, and Z.~J. Kong (2023).
\newblock Anomaly detection in additive manufacturing processes using supervised classification with imbalanced sensor data based on generative adversarial network.
\newblock {\em Journal of Intelligent Manufacturing\/}, 1--20.

\bibitem[\protect\citeauthoryear{Chung, Shen, Law, and Kong}{Chung et~al.}{2022}]{Chung2022_Reinforcement}
Chung, J., B.~Shen, A.~C.~C. Law, and Z.~J. Kong (2022).
\newblock Reinforcement learning-based defect mitigation for quality assurance of additive manufacturing.
\newblock {\em Journal of Manufacturing Systems\/}~{\em 65}, 822--835.

\bibitem[\protect\citeauthoryear{Chung, Zhang, Saimon, Liu, Johnson, and Kong}{Chung et~al.}{2024}]{chung_saimon2024imbalanced_glyco_data}
Chung, J., J.~Zhang, A.~I. Saimon, Y.~Liu, B.~N. Johnson, and Z.~J. Kong (2024).
\newblock Imbalanced spectral data analysis using data augmentation based on the generative adversarial network.

\bibitem[\protect\citeauthoryear{Covidence}{Covidence}{2024}]{covidence}
Covidence (2024).
\newblock Covidence: Systematic review software.
\newblock Available at : \url{https://www.covidence.org/} (accessed on June 1, 2024).

\bibitem[\protect\citeauthoryear{Croom, Berkson, Mueller, Presley, and Storck}{Croom et~al.}{2022}]{Croom_Berkson_Mueller_Presley_Storck2022}
Croom, B.~P., M.~Berkson, R.~K. Mueller, M.~Presley, and S.~Storck (2022).
\newblock Deep learning prediction of stress fields in additively manufactured metals with intricate defect networks.
\newblock {\em Mechanics of Materials\/}~{\em 165}, 104191.

\bibitem[\protect\citeauthoryear{Davtalab, Kazemian, Yuan, and Khoshnevis}{Davtalab et~al.}{2022}]{Davtalab_Kazemian_Yuan_Khoshnevis2022}
Davtalab, O., A.~Kazemian, X.~Yuan, and B.~Khoshnevis (2022).
\newblock Automated inspection in robotic additive manufacturing using deep learning for layer deformation detection.
\newblock {\em Journal of Intelligent Manufacturing\/}~{\em 33\/}(3), 771--784.

\bibitem[\protect\citeauthoryear{Despr{\'e}s, Cyr, Setoodeh, and Mohammadi}{Despr{\'e}s et~al.}{2020}]{Despres_Cyr_Setoodeh_Mohammadi2020}
Despr{\'e}s, N., E.~Cyr, P.~Setoodeh, and M.~Mohammadi (2020).
\newblock Deep learning and design for additive manufacturing: a framework for microlattice architecture.
\newblock {\em Jom\/}~{\em 72}, 2408--2418.

\bibitem[\protect\citeauthoryear{Desrosiers, Letenneur, Bernier, Pich{\'e}, Provencher, Cheriet, Guibault, and Brailovski}{Desrosiers et~al.}{2024}]{Desrosiers_Letenneur__Brailovski2024_porosity_segmentation_LPBF_computed_tomography_CNN}
Desrosiers, C., M.~Letenneur, F.~Bernier, N.~Pich{\'e}, B.~Provencher, F.~Cheriet, F.~Guibault, and V.~Brailovski (2024).
\newblock Automated porosity segmentation in laser powder bed fusion part using computed tomography: a validity study.
\newblock {\em Journal of Intelligent Manufacturing\/}, 1--21.

\bibitem[\protect\citeauthoryear{Dharmawan, Xiong, Foong, and Soh}{Dharmawan et~al.}{2020}]{dharmawan2020model_control_reinforcement}
Dharmawan, A.~G., Y.~Xiong, S.~Foong, and G.~S. Soh (2020).
\newblock A model-based reinforcement learning and correction framework for process control of robotic wire arc additive manufacturing.
\newblock In {\em 2020 IEEE International Conference on Robotics and Automation (ICRA)}, pp.\  4030--4036. IEEE.

\bibitem[\protect\citeauthoryear{Donahue, Anne~Hendricks, Guadarrama, Rohrbach, Venugopalan, Saenko, and Darrell}{Donahue et~al.}{2015}]{Donahue_Hendricks__Darrell2015_LRCN}
Donahue, J., L.~Anne~Hendricks, S.~Guadarrama, M.~Rohrbach, S.~Venugopalan, K.~Saenko, and T.~Darrell (2015).
\newblock Long-term recurrent convolutional networks for visual recognition and description.
\newblock In {\em Proceedings of the IEEE conference on computer vision and pattern recognition}, pp.\  2625--2634.

\bibitem[\protect\citeauthoryear{Drissi-Daoudi, Pandiyan, Log{\'e}, Shevchik, Masinelli, Ghasemi-Tabasi, Parrilli, and Wasmer}{Drissi-Daoudi et~al.}{2022}]{Drissi-Daoudi_Pandiyan_Wasmer2022}
Drissi-Daoudi, R., V.~Pandiyan, R.~Log{\'e}, S.~Shevchik, G.~Masinelli, H.~Ghasemi-Tabasi, A.~Parrilli, and K.~Wasmer (2022).
\newblock Differentiation of materials and laser powder bed fusion processing regimes from airborne acoustic emission combined with machine learning.
\newblock {\em Virtual and Physical Prototyping\/}~{\em 17\/}(2), 181--204.

\bibitem[\protect\citeauthoryear{Fang, Cheng, Glerum, Bennett, Cao, and Wagner}{Fang et~al.}{2022}]{Fang_Cheng_Glerum_Bennett_Cao_Wagner2022}
Fang, L., L.~Cheng, J.~A. Glerum, J.~Bennett, J.~Cao, and G.~J. Wagner (2022).
\newblock Data-driven analysis of process, structure, and properties of additively manufactured inconel 718 thin walls.
\newblock {\em npj Computational Materials\/}~{\em 8\/}(1), 126.

\bibitem[\protect\citeauthoryear{Fathizadan, Ju, and Lu}{Fathizadan et~al.}{2021}]{Fathizadan_Ju_Lu2021}
Fathizadan, S., F.~Ju, and Y.~Lu (2021).
\newblock Deep representation learning for process variation management in laser powder bed fusion.
\newblock {\em Additive Manufacturing\/}~{\em 42}, 101961.

\bibitem[\protect\citeauthoryear{Fathizadan, Ju, Lu, and Yang}{Fathizadan et~al.}{2023}]{Fathizadan_Ju__Yang2023_LPBF_ConvLSTM_Anomaly_Detection}
Fathizadan, S., F.~Ju, Y.~Lu, and Z.~Yang (2023).
\newblock Deep spatio-temporal anomaly detection in laser powder bed fusion.
\newblock {\em IEEE Transactions on Automation Science and Engineering\/}.

\bibitem[\protect\citeauthoryear{Fischer, Zimmermann, Praetzsch, and Knaak}{Fischer et~al.}{2022}]{Fischer_Zimmermann_Praetzsch_Knaak2022}
Fischer, F.~G., M.~G. Zimmermann, N.~Praetzsch, and C.~Knaak (2022).
\newblock Monitoring of the powder bed quality in metal additive manufacturing using deep transfer learning.
\newblock {\em Materials \& Design\/}~{\em 222}, 111029.

\bibitem[\protect\citeauthoryear{Fotovvati and Chou}{Fotovvati and Chou}{2022}]{Fotovvati_Chou2022_SLM_surface_roughness_prediction_ANN}
Fotovvati, B. and K.~Chou (2022).
\newblock Build surface study of single-layer raster scanning in selective laser melting: surface roughness prediction using deep learning.
\newblock {\em Manufacturing Letters\/}~{\em 33\/}(S).

\bibitem[\protect\citeauthoryear{Francis and Bian}{Francis and Bian}{2019}]{Francis_Bian2019_distortion_prediction}
Francis, J. and L.~Bian (2019).
\newblock Deep learning for distortion prediction in laser-based additive manufacturing using big data.
\newblock {\em Manufacturing Letters\/}~{\em 20}, 10--14.

\bibitem[\protect\citeauthoryear{Fu, Downey, Yuan, Zhang, Pratt, and Balogun}{Fu et~al.}{2022}]{Fu_Downey_Yuan__Balogun2022_ML_General_LBAM_review}
Fu, Y., A.~R. Downey, L.~Yuan, T.~Zhang, A.~Pratt, and Y.~Balogun (2022).
\newblock Machine learning algorithms for defect detection in metal laser-based additive manufacturing: A review.
\newblock {\em Journal of Manufacturing Processes\/}~{\em 75}, 693--710.

\bibitem[\protect\citeauthoryear{Fullington, Bian, and Tian}{Fullington et~al.}{2023}]{fullington2023design}
Fullington, D., L.~Bian, and W.~Tian (2023).
\newblock Design de-identification of thermal history for collaborative process-defect modeling of directed energy deposition processes.
\newblock {\em Journal of Manufacturing Science and Engineering\/}~{\em 145\/}(5), 051004.

\bibitem[\protect\citeauthoryear{Garland, White, Jared, Heiden, Donahue, and Boyce}{Garland et~al.}{2020}]{Garland_White__Boyce2020}
Garland, A.~P., B.~C. White, B.~H. Jared, M.~Heiden, E.~Donahue, and B.~L. Boyce (2020).
\newblock Deep convolutional neural networks as a rapid screening tool for complex additively manufactured structures.
\newblock {\em Additive Manufacturing\/}~{\em 35}, 101217.

\bibitem[\protect\citeauthoryear{George, Trevisan~Mota, Maguire, O'Callaghan, Roche, and Papakostas}{George et~al.}{2024}]{George2024_voxelization_refererence_MDPI}
George, A., M.~Trevisan~Mota, C.~Maguire, C.~O'Callaghan, K.~Roche, and N.~Papakostas (2024).
\newblock Using voxelisation-based data analysis techniques for porosity prediction in metal additive manufacturing.
\newblock {\em Applied Sciences\/}~{\em 14\/}(11), 4367.

\bibitem[\protect\citeauthoryear{Ghungrad, Faegh, Gould, Wolff, and Haghighi}{Ghungrad et~al.}{2023}]{Ghungrad_Faegh_Gould_Wolff_Haghighi2023_PIDL}
Ghungrad, S., M.~Faegh, B.~Gould, S.~J. Wolff, and A.~Haghighi (2023).
\newblock Architecture-driven physics-informed deep learning for temperature prediction in laser powder bed fusion additive manufacturing with limited data.
\newblock {\em Journal of Manufacturing Science and Engineering\/}~{\em 145\/}(8), 081007.

\bibitem[\protect\citeauthoryear{Ghungrad, Gould, Soltanalian, Wolff, and Haghighi}{Ghungrad et~al.}{2021}]{Ghungrad_Gould_Soltanalian_Wolff_Haghighi2021}
Ghungrad, S., B.~Gould, M.~Soltanalian, S.~J. Wolff, and A.~Haghighi (2021).
\newblock Model-based deep learning for additive manufacturing: New frontiers and applications.
\newblock {\em Manufacturing Letters\/}~{\em 29}, 94--98.

\bibitem[\protect\citeauthoryear{Ghungrad, Gould, Wolff, and Haghighi}{Ghungrad et~al.}{2022}]{Ghungrad_Gould_Wolff_Haghighi2022_physics_informed_AI}
Ghungrad, S., B.~Gould, S.~Wolff, and A.~Haghighi (2022).
\newblock Physics-informed artificial intelligence for temperature prediction in metal additive manufacturing: A comparative study.
\newblock In {\em International Manufacturing Science and Engineering Conference}, Volume 85802, pp.\  V001T01A008. American Society of Mechanical Engineers.

\bibitem[\protect\citeauthoryear{Gibson, Rosen, Stucker, Khorasani, Rosen, Stucker, and Khorasani}{Gibson et~al.}{2021}]{Gibson_Rosen_Stucker_2021}
Gibson, I., D.~Rosen, B.~Stucker, M.~Khorasani, D.~Rosen, B.~Stucker, and M.~Khorasani (2021).
\newblock {\em Additive manufacturing technologies}, Volume~17.
\newblock Springer.

\bibitem[\protect\citeauthoryear{Glasder, Fabbri, Aschwanden, Bambach, and Wegener}{Glasder et~al.}{2023}]{Glasder_Fabbri__Wegener2023_footprint_weld_bead_prediction_using_transfer_learning_WAMM_CNN}
Glasder, M., M.~Fabbri, I.~Aschwanden, M.~Bambach, and K.~Wegener (2023).
\newblock Towards a general and numerically efficient deposition model for wire-arc directed energy deposition.
\newblock {\em Additive Manufacturing\/}~{\em 78}, 103832.

\bibitem[\protect\citeauthoryear{Goh, Sing, and Yeong}{Goh et~al.}{2021}]{Goh_Sing_Yeong_2021_review}
Goh, G.~D., S.~L. Sing, and W.~Y. Yeong (2021).
\newblock A review on machine learning in 3d printing: applications, potential, and challenges.
\newblock {\em Artificial Intelligence Review\/}~{\em 54\/}(1), 63--94.

\bibitem[\protect\citeauthoryear{Goodfellow, Bengio, and Courville}{Goodfellow et~al.}{2016}]{Goodfellow_Bengio_Courville2016_DL_book}
Goodfellow, I., Y.~Bengio, and A.~Courville (2016).
\newblock {\em Deep learning}.
\newblock MIT press.

\bibitem[\protect\citeauthoryear{Guo, Guo, Bian, and Guo}{Guo et~al.}{2022}]{Guo_Guo_Bian_Guo2022}
Guo, S., W.~Guo, L.~Bian, and Y.~Guo (2022).
\newblock A deep-learning-based surrogate model for thermal signature prediction in laser metal deposition.
\newblock {\em IEEE Transactions on Automation Science and Engineering\/}.

\bibitem[\protect\citeauthoryear{Guo, Ko, and Wang}{Guo et~al.}{2023}]{Guo_Ko_Wang2023_ML_AJP_review}
Guo, S., H.~Ko, and A.~Wang (2023).
\newblock Applications and prospects of machine learning for aerosol jet printing: A review.
\newblock {\em IISE Transactions\/}~(just-accepted), 1--31.

\bibitem[\protect\citeauthoryear{Guo, Lu, and Fuh}{Guo et~al.}{2021}]{Guo_Lu_Fuh2021}
Guo, Y., W.~F. Lu, and J.~Y.~H. Fuh (2021).
\newblock Semi-supervised deep learning based framework for assessing manufacturability of cellular structures in direct metal laser sintering process.
\newblock {\em Journal of Intelligent Manufacturing\/}~{\em 32}, 347--359.

\bibitem[\protect\citeauthoryear{Ha, Yao, Xu, Liu, Liu, Elkins, Kile, Deshpande, Kong, Bauchy, et~al.}{Ha et~al.}{2023}]{Ha_Yao__Zheng2023rapid_inverse_design}
Ha, C.~S., D.~Yao, Z.~Xu, C.~Liu, H.~Liu, D.~Elkins, M.~Kile, V.~Deshpande, Z.~Kong, M.~Bauchy, et~al. (2023).
\newblock Rapid inverse design of metamaterials based on prescribed mechanical behavior through machine learning.
\newblock {\em Nature Communications\/}~{\em 14\/}(1), 5765.

\bibitem[\protect\citeauthoryear{Hamrani, Agarwal, Allouhi, and McDaniel}{Hamrani et~al.}{2023}]{Hamrani_Agarwal__McDaniel2023_Ml_AM_review}
Hamrani, A., A.~Agarwal, A.~Allouhi, and D.~McDaniel (2023).
\newblock Applying machine learning to wire arc additive manufacturing: a systematic data-driven literature review.
\newblock {\em Journal of Intelligent Manufacturing\/}, 1--33.

\bibitem[\protect\citeauthoryear{He, Yuan, Mu, Ros, Ding, Pan, and Li}{He et~al.}{2023}]{He_Yuan_Mu_Ros__Li2023_review_AI_AM}
He, F., L.~Yuan, H.~Mu, M.~Ros, D.~Ding, Z.~Pan, and H.~Li (2023).
\newblock Research and application of artificial intelligence techniques for wire arc additive manufacturing: a state-of-the-art review.
\newblock {\em Robotics and Computer-Integrated Manufacturing\/}~{\em 82}, 102525.

\bibitem[\protect\citeauthoryear{He, Zhang, Ren, and Sun}{He et~al.}{2016}]{He_Zhang_Ren_Sun2016_ResNet_Supp_paper}
He, K., X.~Zhang, S.~Ren, and J.~Sun (2016).
\newblock Deep residual learning for image recognition.
\newblock In {\em Proceedings of the IEEE conference on computer vision and pattern recognition}, pp.\  770--778.

\bibitem[\protect\citeauthoryear{Hemmasian, Ogoke, Akbari, Malen, Beuth, and Farimani}{Hemmasian et~al.}{2023}]{Hemmasian_Ogoke__Beuth_Farimani2023}
Hemmasian, A., F.~Ogoke, P.~Akbari, J.~Malen, J.~Beuth, and A.~B. Farimani (2023).
\newblock Surrogate modeling of melt pool temperature field using deep learning.
\newblock {\em Additive Manufacturing Letters\/}~{\em 5}, 100123.

\bibitem[\protect\citeauthoryear{Herbeaux, Aboleinein, Villani, Maurice, Bergheau, and Kl{\"o}cker}{Herbeaux et~al.}{2024}]{Herbeaux_Aboleinein__Klocker2024_microstructure_modeling_from_SEM_EBSD_images_CNN_WAAM_or_CMT}
Herbeaux, A., H.~Aboleinein, A.~Villani, C.~Maurice, J.-M. Bergheau, and H.~Kl{\"o}cker (2024).
\newblock Combining phase field modeling and deep learning for accurate modeling of grain structure in solidification.
\newblock {\em Additive Manufacturing\/}~{\em 81}, 103994.

\bibitem[\protect\citeauthoryear{Herriott and Spear}{Herriott and Spear}{2020}]{Herriott_Spear2020_property_prediction}
Herriott, C. and A.~D. Spear (2020).
\newblock Predicting microstructure-dependent mechanical properties in additively manufactured metals with machine-and deep-learning methods.
\newblock {\em Computational Materials Science\/}~{\em 175}, 109599.

\bibitem[\protect\citeauthoryear{Hertlein, Buskohl, Gillman, Vemaganti, and Anand}{Hertlein et~al.}{2021}]{Hertlein_Buskohl_Gillman_Vemaganti_Anand2021}
Hertlein, N., P.~R. Buskohl, A.~Gillman, K.~Vemaganti, and S.~Anand (2021).
\newblock Generative adversarial network for early-stage design flexibility in topology optimization for additive manufacturing.
\newblock {\em Journal of Manufacturing Systems\/}~{\em 59}, 675--685.

\bibitem[\protect\citeauthoryear{Hespeler, Dehghan-Niri, Juhasz, Luo, and Halliday}{Hespeler et~al.}{2022}]{Hespeler_Dehghan-Niri__Halliday2022_MDPI_paper}
Hespeler, S., E.~Dehghan-Niri, M.~Juhasz, K.~Luo, and H.~S. Halliday (2022).
\newblock Deep learning for in-situ layer quality monitoring during laser-based directed energy deposition (lb-ded) additive manufacturing process.
\newblock {\em Applied Sciences\/}~{\em 12\/}(18), 8974.

\bibitem[\protect\citeauthoryear{Hilbig, Vogt, Holtzhausen, and Paetzold}{Hilbig et~al.}{2023}]{Hilbig_Vogt_Holtzhausen_Paetzold2023_geometric_feature_recognition}
Hilbig, A., L.~Vogt, S.~Holtzhausen, and K.~Paetzold (2023).
\newblock Enhancing three-dimensional convolutional neural network-based geometric feature recognition for adaptive additive manufacturing: a signed distance field data approach.
\newblock {\em Journal of Computational Design and Engineering\/}~{\em 10\/}(3), 992--1009.

\bibitem[\protect\citeauthoryear{Ho, Zhang, Young, Buchholz, Al~Jufout, Dajani, Bian, and Mozumdar}{Ho et~al.}{2021}]{Ho_Zhang_Young__Mozumder2021}
Ho, S., W.~Zhang, W.~Young, M.~Buchholz, S.~Al~Jufout, K.~Dajani, L.~Bian, and M.~Mozumdar (2021).
\newblock Dlam: Deep learning based real-time porosity prediction for additive manufacturing using thermal images of the melt pool.
\newblock {\em IEEE Access\/}~{\em 9}, 115100--115114.

\bibitem[\protect\citeauthoryear{Howard, Zhu, Chen, Kalenichenko, Wang, Weyand, Andreetto, and Adam}{Howard et~al.}{2017}]{Howard_Zhu_Chen__Adam2017Mobilenets}
Howard, A.~G., M.~Zhu, B.~Chen, D.~Kalenichenko, W.~Wang, T.~Weyand, M.~Andreetto, and H.~Adam (2017).
\newblock Mobilenets: Efficient convolutional neural networks for mobile vision applications.
\newblock {\em arXiv preprint arXiv:1704.04861\/}.

\bibitem[\protect\citeauthoryear{Hu, Wang, Li, and Wang}{Hu et~al.}{2022}]{Hu_Wang_Li_Wang2022}
Hu, K., Y.~Wang, W.~Li, and L.~Wang (2022).
\newblock Cnn-bilstm enabled prediction on molten pool width for thin-walled part fabrication using laser directed energy deposition.
\newblock {\em Journal of Manufacturing Processes\/}~{\em 78}, 32--45.

\bibitem[\protect\citeauthoryear{Huang, Ma, Chen, Feng, and Murakawa}{Huang et~al.}{2020}]{huang2020_supp_ref_for_criticism_physics_based_model}
Huang, H., N.~Ma, J.~Chen, Z.~Feng, and H.~Murakawa (2020).
\newblock Toward large-scale simulation of residual stress and distortion in wire and arc additive manufacturing.
\newblock {\em Additive manufacturing\/}~{\em 34}, 101248.

\bibitem[\protect\citeauthoryear{Huang, Segura, Wang, Zhao, Sun, and Zhou}{Huang et~al.}{2020}]{Huang_Segura__Zhou2020_droplet_evolution_prediction_inkjet_DRNN}
Huang, J., L.~J. Segura, T.~Wang, G.~Zhao, H.~Sun, and C.~Zhou (2020).
\newblock Unsupervised learning for the droplet evolution prediction and process dynamics understanding in inkjet printing.
\newblock {\em Additive Manufacturing\/}~{\em 35}, 101197.

\bibitem[\protect\citeauthoryear{Huang, Sun, Kwok, Zhou, and Xu}{Huang et~al.}{2020}]{Huang_Sun_Kwok_Zhou_Xu2020}
Huang, J., H.~Sun, T.-H. Kwok, C.~Zhou, and W.~Xu (2020).
\newblock Geometric deep learning for shape correspondence in mass customization by three-dimensional printing.
\newblock {\em Journal of Manufacturing Science and Engineering\/}~{\em 142\/}(6), 061003.

\bibitem[\protect\citeauthoryear{Huang}{Huang}{2016}]{Qiang_huang2016analytical_out_of_plane_supp}
Huang, Q. (2016).
\newblock An analytical foundation for optimal compensation of three-dimensional shape deformation in additive manufacturing.
\newblock {\em Journal of Manufacturing Science and Engineering\/}~{\em 138\/}(6), 061010.

\bibitem[\protect\citeauthoryear{Huang, Nouri, Xu, Chen, Sosina, and Dasgupta}{Huang et~al.}{2014}]{Qiang_huang2014statistical_modeling_in_plane1_supp}
Huang, Q., H.~Nouri, K.~Xu, Y.~Chen, S.~Sosina, and T.~Dasgupta (2014).
\newblock Statistical predictive modeling and compensation of geometric deviations of three-dimensional printed products.
\newblock {\em Journal of Manufacturing Science and Engineering\/}~{\em 136\/}(6), 061008.

\bibitem[\protect\citeauthoryear{Huang, Zhang, Sabbaghi, and Dasgupta}{Huang et~al.}{2015}]{Qiang_huang2015optimal_shrinkage_in_plane2_supp}
Huang, Q., J.~Zhang, A.~Sabbaghi, and T.~Dasgupta (2015).
\newblock Optimal offline compensation of shape shrinkage for three-dimensional printing processes.
\newblock {\em IIE transactions\/}~{\em 47\/}(5), 431--441.

\bibitem[\protect\citeauthoryear{Ibhadode, Zhang, Sixt, Nsiempba, Orakwe, Martinez-Marchese, Ero, Shahabad, Bonakdar, and Toyserkani}{Ibhadode et~al.}{2023}]{Ibhadode_Zhang__Toyserkani2023topology_Supplemenatary_article}
Ibhadode, O., Z.~Zhang, J.~Sixt, K.~M. Nsiempba, J.~Orakwe, A.~Martinez-Marchese, O.~Ero, S.~I. Shahabad, A.~Bonakdar, and E.~Toyserkani (2023).
\newblock Topology optimization for metal additive manufacturing: current trends, challenges, and future outlook.
\newblock {\em Virtual and Physical Prototyping\/}~{\em 18\/}(1), e2181192.

\bibitem[\protect\citeauthoryear{Imani, Chen, Diewald, Reutzel, and Yang}{Imani et~al.}{2019}]{Imani_Chen_Diewald__Yang2019_defect_detection}
Imani, F., R.~Chen, E.~Diewald, E.~Reutzel, and H.~Yang (2019).
\newblock Deep learning of variant geometry in layerwise imaging profiles for additive manufacturing quality control.
\newblock {\em Journal of Manufacturing Science and Engineering\/}~{\em 141\/}(11), 111001.

\bibitem[\protect\citeauthoryear{Inyang-Udoh, Chen, and Mishra}{Inyang-Udoh et~al.}{2022}]{Inyang-Udoh_Chen_Mishra2022}
Inyang-Udoh, U., A.~Chen, and S.~Mishra (2022).
\newblock A learn-and-control strategy for jet-based additive manufacturing.
\newblock {\em IEEE/ASME Transactions on Mechatronics\/}~{\em 27\/}(4), 1946--1954.

\bibitem[\protect\citeauthoryear{Inyang-Udoh, Guo, Peters, Oomen, and Mishra}{Inyang-Udoh et~al.}{2020}]{Inyang-Udoh_Guo_Peters_Oomen_Mishra2020_layer_to_layer_predictive_control}
Inyang-Udoh, U., Y.~Guo, J.~Peters, T.~Oomen, and S.~Mishra (2020).
\newblock Layer-to-layer predictive control of inkjet 3-d printing.
\newblock {\em IEEE/ASME Transactions on Mechatronics\/}~{\em 25\/}(4), 1783--1793.

\bibitem[\protect\citeauthoryear{Inyang-Udoh and Mishra}{Inyang-Udoh and Mishra}{2021}]{Inyang-Udoh_Mishra2021_PINN}
Inyang-Udoh, U. and S.~Mishra (2021).
\newblock A physics-guided neural network dynamical model for droplet-based additive manufacturing.
\newblock {\em IEEE Transactions on Control Systems Technology\/}~{\em 30\/}(5), 1863--1875.

\bibitem[\protect\citeauthoryear{{ISO/ASTM}}{{ISO/ASTM}}{2021}]{ISOASTM52900}
{ISO/ASTM} (2021).
\newblock {ISO/ASTM} 52900:2021 -- additive manufacturing -- general principles -- terminology.
\newblock Available at : \url{https://www.iso.org/obp/ui/#iso:std:iso-astm:52900:ed-2:v1:en} (accessed on June 1, 2024).

\bibitem[\protect\citeauthoryear{Iyer, Mirzendehdel, Raghavan, Jiao, Ulu, Behandish, Nelaturi, and Robinson}{Iyer et~al.}{2021}]{Iyer_Mirzendehdel__Robinson2021pato}
Iyer, N.~S., A.~M. Mirzendehdel, S.~Raghavan, Y.~Jiao, E.~Ulu, M.~Behandish, S.~Nelaturi, and D.~M. Robinson (2021).
\newblock Pato: producibility-aware topology optimization using deep learning for metal additive manufacturing.
\newblock {\em arXiv preprint arXiv:2112.04552\/}.

\bibitem[\protect\citeauthoryear{Jamnikar, Liu, Brice, and Zhang}{Jamnikar et~al.}{2023}]{Jamnikar_Liu_Brice_Zhang2023_process-optimization}
Jamnikar, N., S.~Liu, C.~Brice, and X.~Zhang (2023).
\newblock Comprehensive molten pool condition-process relations modeling using cnn for wire-feed laser additive manufacturing.
\newblock {\em Journal of Manufacturing Processes\/}~{\em 98}, 42--53.

\bibitem[\protect\citeauthoryear{Jiang}{Jiang}{2023}]{Jiang2023_AM_ML_survey}
Jiang, J. (2023).
\newblock A survey of machine learning in additive manufacturing technologies.
\newblock {\em International Journal of Computer Integrated Manufacturing\/}, 1--23.

\bibitem[\protect\citeauthoryear{Jiang, Xiong, Zhang, and Rosen}{Jiang et~al.}{2022}]{Jiang_Xiong_Zhang_Rosen2022_DNN_DfAM}
Jiang, J., Y.~Xiong, Z.~Zhang, and D.~W. Rosen (2022).
\newblock Machine learning integrated design for additive manufacturing.
\newblock {\em Journal of Intelligent Manufacturing\/}~{\em 33\/}(4), 1073--1086.

\bibitem[\protect\citeauthoryear{Jiang, Smith, Yi, Sun, Simonds, and Rollett}{Jiang et~al.}{2024}]{Jiang_Smith__Rollett2024_laser_absorptance_prediction_CNN_AM}
Jiang, R., J.~Smith, Y.-T. Yi, T.~Sun, B.~J. Simonds, and A.~D. Rollett (2024).
\newblock Deep learning approaches for instantaneous laser absorptance prediction in additive manufacturing.
\newblock {\em npj Computational Materials\/}~{\em 10\/}(1), 6.

\bibitem[\protect\citeauthoryear{Jiang, Zhang, Chen, Ma, Yuan, Deng, and Zhang}{Jiang et~al.}{2023}]{Jiang_Zhang__ZHang2023_CNN_layerwise_imagesLPBF_surface_defect_classification}
Jiang, Z., A.~Zhang, Z.~Chen, C.~Ma, Z.~Yuan, Y.~Deng, and Y.~Zhang (2023).
\newblock A deep convolutional network combining layerwise images and defect parameter vectors for laser powder bed fusion process anomalies classification.
\newblock {\em Journal of Intelligent Manufacturing\/}, 1--31.

\bibitem[\protect\citeauthoryear{Jin, Zhang, Demir, and Gu}{Jin et~al.}{2020}]{Jin_Zhang_Demir_Gu_2020_review}
Jin, Z., Z.~Zhang, K.~Demir, and G.~X. Gu (2020).
\newblock Machine learning for advanced additive manufacturing.
\newblock {\em Matter\/}~{\em 3\/}(5), 1541--1556.

\bibitem[\protect\citeauthoryear{Jin, Zhang, and Gu}{Jin et~al.}{2019}]{Jin_Zhang_Gu2019}
Jin, Z., Z.~Zhang, and G.~X. Gu (2019).
\newblock Autonomous in-situ correction of fused deposition modeling printers using computer vision and deep learning.
\newblock {\em Manufacturing Letters\/}~{\em 22}, 11--15.

\bibitem[\protect\citeauthoryear{Johnson, Maestas, Emery, Grigoriu, Smith, and Martinez}{Johnson et~al.}{2022}]{Johnson_Maestas_Martinez2022}
Johnson, K.~L., D.~Maestas, J.~M. Emery, M.~D. Grigoriu, M.~D. Smith, and C.~Martinez (2022).
\newblock Failure classification of porous additively manufactured parts using deep learning.
\newblock {\em Computational Materials Science\/}~{\em 204}, 111098.

\bibitem[\protect\citeauthoryear{Kaji, Nguyen-Huu, Budhwani, Narayanan, Zimny, and Toyserkani}{Kaji et~al.}{2022}]{Kaji_Nguyen-Huu_Toyserkani2022_surface_anomaly_detection_DED_point_cloud}
Kaji, F., H.~Nguyen-Huu, A.~Budhwani, J.~A. Narayanan, M.~Zimny, and E.~Toyserkani (2022).
\newblock A deep-learning-based in-situ surface anomaly detection methodology for laser directed energy deposition via powder feeding.
\newblock {\em Journal of Manufacturing Processes\/}~{\em 81}, 624--637.

\bibitem[\protect\citeauthoryear{Khadilkar, Wang, and Rai}{Khadilkar et~al.}{2019}]{Khadilkar_Wang_Rai2019_stress_prediction_SLA_CNN}
Khadilkar, A., J.~Wang, and R.~Rai (2019).
\newblock Deep learning--based stress prediction for bottom-up sla 3d printing process.
\newblock {\em The International Journal of Advanced Manufacturing Technology\/}~{\em 102}, 2555--2569.

\bibitem[\protect\citeauthoryear{Khusheef, Shahbazi, and Hashemi}{Khusheef et~al.}{2023}]{Khusheef_Shahbazi_Hashemi2023_copare_three_levels_of_fusion_FDM_LSTM_transfer_learning}
Khusheef, A.~S., M.~Shahbazi, and R.~Hashemi (2023).
\newblock Deep learning-based multi-sensor fusion for process monitoring: Application to fused deposition modeling.
\newblock {\em Arabian Journal for Science and Engineering\/}, 1--22.

\bibitem[\protect\citeauthoryear{Kim and Zohdi}{Kim and Zohdi}{2022}]{Kim_Zohdi2022}
Kim, D. and T.~Zohdi (2022).
\newblock Tool path optimization of selective laser sintering processes using deep learning.
\newblock {\em Computational Mechanics\/}~{\em 69\/}(1), 383--401.

\bibitem[\protect\citeauthoryear{Kim, Chong, Mahdi, and Shin}{Kim et~al.}{2024}]{Kim_Chong__Shin2024_WAAM_defect_detection_CNN_transfer_learning}
Kim, D.~B., H.~Chong, M.~M. Mahdi, and S.-J. Shin (2024).
\newblock Data-fused and concatenated-ensemble learning for in-situ anomaly detection in wire and arc-based direct energy deposition.
\newblock {\em Journal of Manufacturing Processes\/}~{\em 112}, 273--289.

\bibitem[\protect\citeauthoryear{Kim, Lee, Seo, Kim, and Shin}{Kim et~al.}{2023}]{Kim_Lee_Seo_Kim_Shin2023_in_Situ_monitoring}
Kim, E.-S., D.-H. Lee, G.-J. Seo, D.-B. Kim, and S.-J. Shin (2023).
\newblock Development of a cnn-based real-time monitoring algorithm for additively manufactured molybdenum.
\newblock {\em Sensors and Actuators A: Physical\/}~{\em 352}, 114205.

\bibitem[\protect\citeauthoryear{Kim, Lee, and Ahn}{Kim et~al.}{2022}]{Kim_Lee_Ahn2022}
Kim, H., H.~Lee, and S.-H. Ahn (2022).
\newblock Systematic deep transfer learning method based on a small image dataset for spaghetti-shape defect monitoring of fused deposition modeling.
\newblock {\em Journal of Manufacturing Systems\/}~{\em 65}, 439--451.

\bibitem[\protect\citeauthoryear{Kim, Lee, Kim, and Ahn}{Kim et~al.}{2020}]{Kim_Lee__Ahn2020_failure_detection_ME_VGGNet_transfer_learning}
Kim, H., H.~Lee, J.-S. Kim, and S.-H. Ahn (2020).
\newblock Image-based failure detection for material extrusion process using a convolutional neural network.
\newblock {\em The International Journal of Advanced Manufacturing Technology\/}~{\em 111\/}(5), 1291--1302.

\bibitem[\protect\citeauthoryear{Kim, Yang, Ko, Cho, and Lu}{Kim et~al.}{2023}]{Kim_Yang_Ko_Cho_Lu2023}
Kim, J., Z.~Yang, H.~Ko, H.~Cho, and Y.~Lu (2023).
\newblock Deep learning-based data registration of melt-pool-monitoring images for laser powder bed fusion additive manufacturing.
\newblock {\em Journal of Manufacturing Systems\/}~{\em 68}, 117--129.

\bibitem[\protect\citeauthoryear{Kim, Oh, Lee, and Kim}{Kim et~al.}{2022}]{Kim_Oh__Kim2022_pore_detection_YOLOv5_pore_image_DED}
Kim, J.-H., W.-J. Oh, C.-M. Lee, and D.-H. Kim (2022).
\newblock Achieving optimal process design for minimizing porosity in additive manufacturing of inconel 718 using a deep learning-based pore detection approach.
\newblock {\em The International Journal of Advanced Manufacturing Technology\/}~{\em 121\/}(3), 2115--2134.

\bibitem[\protect\citeauthoryear{Kim, Lee, Hyun, Coatanea, Mika, Mo, and Yoo}{Kim et~al.}{2023}]{Kim_Lee__Yoo2023_time_series_augmentation_StyleGAN}
Kim, Y., T.~Lee, Y.~Hyun, E.~Coatanea, S.~Mika, J.~Mo, and Y.~Yoo (2023).
\newblock Self-supervised representation learning anomaly detection methodology based on boosting algorithms enhanced by data augmentation using stylegan for manufacturing imbalanced data.
\newblock {\em Computers in Industry\/}~{\em 153}, 104024.

\bibitem[\protect\citeauthoryear{Ko, Kim, Lu, Shin, Yang, and Oh}{Ko et~al.}{2022}]{Ko_Kim_Lu_Shin_Yang_Oh2022_spatial}
Ko, H., J.~Kim, Y.~Lu, D.~Shin, Z.~Yang, and Y.~Oh (2022).
\newblock Spatial-temporal modeling using deep learning for real-time monitoring of additive manufacturing.
\newblock In {\em International Design Engineering Technical Conferences and Computers and Information in Engineering Conference}, Volume 86212, pp.\  V002T02A019. American Society of Mechanical Engineers.

\bibitem[\protect\citeauthoryear{Ko{\c{c}}, Zeybek, K{\i}sas{\"o}z, {\c{C}}al{\i}{\c{s}}kan, and Bulduk}{Ko{\c{c}} et~al.}{2022}]{Koc_Zeybek_Kisasoz__Bulduk2022}
Ko{\c{c}}, E., S.~Zeybek, B.~{\"O}. K{\i}sas{\"o}z, C.~{\.I}. {\c{C}}al{\i}{\c{s}}kan, and M.~E. Bulduk (2022).
\newblock Estimation of surface roughness in selective laser sintering using computational models.
\newblock {\em The International Journal of Advanced Manufacturing Technology\/}, 1--13.

\bibitem[\protect\citeauthoryear{Kouraytem, Li, Tan, Kappes, and Spear}{Kouraytem et~al.}{2021}]{Kouraytem2021modeling_supp_ref}
Kouraytem, N., X.~Li, W.~Tan, B.~Kappes, and A.~D. Spear (2021).
\newblock Modeling process--structure--property relationships in metal additive manufacturing: a review on physics-driven versus data-driven approaches.
\newblock {\em Journal of Physics: Materials\/}~{\em 4\/}(3), 032002.

\bibitem[\protect\citeauthoryear{Krizhevsky, Sutskever, and Hinton}{Krizhevsky et~al.}{2012}]{Krizhevsky_Sutskever_Hinton2012_ImageNet_Classification_with_DCNN}
Krizhevsky, A., I.~Sutskever, and G.~E. Hinton (2012).
\newblock Imagenet classification with deep convolutional neural networks.
\newblock {\em Advances in neural information processing systems\/}~{\em 25}.

\bibitem[\protect\citeauthoryear{Kumar, Gopi, Harikeerthana, Gupta, Gaur, Krolczyk, and Wu}{Kumar et~al.}{2023}]{Kumar_Gopi__Wu2023_review}
Kumar, S., T.~Gopi, N.~Harikeerthana, M.~K. Gupta, V.~Gaur, G.~M. Krolczyk, and C.~Wu (2023).
\newblock Machine learning techniques in additive manufacturing: a state of the art review on design, processes and production control.
\newblock {\em Journal of Intelligent Manufacturing\/}~{\em 34\/}(1), 21--55.

\bibitem[\protect\citeauthoryear{Kwon, Kim, Ham, Kim, Kim, Cho, Kim, and Kim}{Kwon et~al.}{2020}]{Kwon_Kim_Ham__Kim2020}
Kwon, O., H.~G. Kim, M.~J. Ham, W.~Kim, G.-H. Kim, J.-H. Cho, N.~I. Kim, and K.~Kim (2020).
\newblock A deep neural network for classification of melt-pool images in metal additive manufacturing.
\newblock {\em Journal of Intelligent Manufacturing\/}~{\em 31}, 375--386.

\bibitem[\protect\citeauthoryear{Kwon, Kim, Lee, and Lee}{Kwon et~al.}{2023}]{Kwon_Kim__Lee2023_PIDL_inkjet_ConvLSTM_next_image_prediction_from_sequential_images}
Kwon, S.~W., J.~S. Kim, H.~M. Lee, and J.~S. Lee (2023).
\newblock Physics-added neural networks: An image-based deep learning for material printing system.
\newblock {\em Additive Manufacturing\/}~{\em 73}, 103668.

\bibitem[\protect\citeauthoryear{Larsen and Hooper}{Larsen and Hooper}{2022}]{Larsen_Hooper2022}
Larsen, S. and P.~A. Hooper (2022).
\newblock Deep semi-supervised learning of dynamics for anomaly detection in laser powder bed fusion.
\newblock {\em Journal of Intelligent Manufacturing\/}~{\em 33\/}(2), 457--471.

\bibitem[\protect\citeauthoryear{LeCun, Bengio, and Hinton}{LeCun et~al.}{2015}]{LeCun_Bengio_Hinton2015_DL_nature_review}
LeCun, Y., Y.~Bengio, and G.~Hinton (2015).
\newblock Deep learning.
\newblock {\em nature\/}~{\em 521\/}(7553), 436--444.

\bibitem[\protect\citeauthoryear{Lee, Heogh, Yang, Yoon, Park, Ji, and Lee}{Lee et~al.}{2022}]{Lee_Heogh_Yang__Lee2022_Powder_Stream_Fault}
Lee, H., W.~Heogh, J.~Yang, J.~Yoon, J.~Park, S.~Ji, and H.~Lee (2022).
\newblock Deep learning for in-situ powder stream fault detection in directed energy deposition process.
\newblock {\em Journal of Manufacturing Systems\/}~{\em 62}, 575--587.

\bibitem[\protect\citeauthoryear{Lee, Zhang, and Gu}{Lee et~al.}{2022}]{Lee_Zhang_Gu2022_NN_GO_lattice_structures_superior_mechanical_properties}
Lee, S., Z.~Zhang, and G.~X. Gu (2022).
\newblock Generative machine learning algorithm for lattice structures with superior mechanical properties.
\newblock {\em Materials Horizons\/}~{\em 9\/}(3), 952--960.

\bibitem[\protect\citeauthoryear{Lee, Saha, Sarkar, and Giera}{Lee et~al.}{2020}]{Lee_Saha_Sarkar_Giera2020}
Lee, X.~Y., S.~K. Saha, S.~Sarkar, and B.~Giera (2020).
\newblock Automated detection of part quality during two-photon lithography via deep learning.
\newblock {\em Additive Manufacturing\/}~{\em 36}, 101444.

\bibitem[\protect\citeauthoryear{Li, Cao, Liu, Zhou, Zhang, and Li}{Li et~al.}{2023}]{Li_Cao_Liu__LI2023_Imbalanced_data_generation_in_situ_monitoring}
Li, J., L.~Cao, H.~Liu, Q.~Zhou, X.~Zhang, and M.~Li (2023).
\newblock Imbalanced data generation and fusion for in-situ monitoring of laser powder bed fusion.
\newblock {\em Mechanical Systems and Signal Processing\/}~{\em 199}, 110508.

\bibitem[\protect\citeauthoryear{Li, Cao, Zhou, Liu, and Zhang}{Li et~al.}{2023}]{Li_Cao__Zhang2023_Imbalanced_SLM_AE}
Li, J., L.~Cao, Q.~Zhou, H.~Liu, and X.~Zhang (2023).
\newblock Imbalanced quality monitoring of selective laser melting using acoustic and photodiode signals.
\newblock {\em Journal of Manufacturing Processes\/}~{\em 105}, 14--26.

\bibitem[\protect\citeauthoryear{Li, Zhou, Huang, Li, and Cao}{Li et~al.}{2023}]{Li_Zhou_Huang_Li_Cao2023}
Li, J., Q.~Zhou, X.~Huang, M.~Li, and L.~Cao (2023).
\newblock In situ quality inspection with layer-wise visual images based on deep transfer learning during selective laser melting.
\newblock {\em Journal of Intelligent Manufacturing\/}, 1--15.

\bibitem[\protect\citeauthoryear{Li, Huang, Liu, and Tan}{Li et~al.}{2024}]{Li_Huang__Tan_2024)_defect_classification_ViT_FDM}
Li, Q., T.~Huang, J.~Liu, and L.~Tan (2024).
\newblock Time-series vision transformer based on cross space-time attention for fault diagnosis in fused deposition modelling with reconstruction of layer-wise data.
\newblock {\em Journal of Manufacturing Processes\/}~{\em 115}, 240--255.

\bibitem[\protect\citeauthoryear{Li, Lambert-Garcia, Getley, Kim, Bhagavath, Majkut, Rack, Lee, and Leung}{Li et~al.}{2024}]{Li_Lambert-Garcia__Leung2024_AM-SegNet}
Li, W., R.~Lambert-Garcia, A.~C. Getley, K.~Kim, S.~Bhagavath, M.~Majkut, A.~Rack, P.~D. Lee, and C.~L.~A. Leung (2024).
\newblock Am-segnet for additive manufacturing in situ x-ray image segmentation and feature quantification.
\newblock {\em Virtual and Physical Prototyping\/}~{\em 19\/}(1), e2325572.

\bibitem[\protect\citeauthoryear{Li, Zhang, Wang, Xiong, Zhao, Li, and Li}{Li et~al.}{2023}]{Li_Zhang__Li2023_surface_defect_detection_WAAM}
Li, W., H.~Zhang, G.~Wang, G.~Xiong, M.~Zhao, G.~Li, and R.~Li (2023).
\newblock Deep learning based online metallic surface defect detection method for wire and arc additive manufacturing.
\newblock {\em Robotics and Computer-Integrated Manufacturing\/}~{\em 80}, 102470.

\bibitem[\protect\citeauthoryear{Li, Jia, Yang, and Lee}{Li et~al.}{2020}]{Li_Jia_Yang_Lee2020}
Li, X., X.~Jia, Q.~Yang, and J.~Lee (2020).
\newblock Quality analysis in metal additive manufacturing with deep learning.
\newblock {\em Journal of Intelligent Manufacturing\/}~{\em 31}, 2003--2017.

\bibitem[\protect\citeauthoryear{Li, Siahpour, Lee, Wang, and Shi}{Li et~al.}{2020}]{Li_Siahpour__Shi2020_CNN_DED_thermal_images}
Li, X., S.~Siahpour, J.~Lee, Y.~Wang, and J.~Shi (2020).
\newblock Deep learning-based intelligent process monitoring of directed energy deposition in additive manufacturing with thermal images.
\newblock {\em Procedia Manufacturing\/}~{\em 48}, 643--649.

\bibitem[\protect\citeauthoryear{Li, Xiong, Li, Wu, Zhang, Liu, Bian, and Dou}{Li et~al.}{2022}]{Li_Xiong__Dou2022_interpretable_DL_review}
Li, X., H.~Xiong, X.~Li, X.~Wu, X.~Zhang, J.~Liu, J.~Bian, and D.~Dou (2022).
\newblock Interpretable deep learning: Interpretation, interpretability, trustworthiness, and beyond.
\newblock {\em Knowledge and Information Systems\/}~{\em 64\/}(12), 3197--3234.

\bibitem[\protect\citeauthoryear{Li, Zhang, Zhou, Wang, Zhu, Wu, and Zhang}{Li et~al.}{2023}]{Li_Zhang_Zhou__Zhang2023}
Li, X., M.~Zhang, M.~Zhou, J.~Wang, W.~Zhu, C.~Wu, and X.~Zhang (2023).
\newblock Qualify assessment for extrusion-based additive manufacturing with 3d scan and machine learning.
\newblock {\em Journal of Manufacturing Processes\/}~{\em 90}, 274--285.

\bibitem[\protect\citeauthoryear{Li, Shi, and Liu}{Li et~al.}{2023}]{Li_Shi_Liu2023_data_imputation}
Li, Y., Z.~Shi, and C.~Liu (2023).
\newblock Transformer-enabled generative adversarial imputation network with selective generation (sgt-gain) for missing region imputation.
\newblock {\em IISE Transactions\/}, 1--13.

\bibitem[\protect\citeauthoryear{Li, Shi, Liu, Tian, Kong, and Williams}{Li et~al.}{2021}]{Li_Shi__Williams2021_ATR_GAN}
Li, Y., Z.~Shi, C.~Liu, W.~Tian, Z.~Kong, and C.~B. Williams (2021).
\newblock Augmented time regularized generative adversarial network (atr-gan) for data augmentation in online process anomaly detection.
\newblock {\em IEEE Transactions on Automation Science and Engineering\/}~{\em 19\/}(4), 3338--3355.

\bibitem[\protect\citeauthoryear{Li, Segura, Li, Zhou, and Sun}{Li et~al.}{2023}]{Li_Segura__Sun2023_Droplet_Pinch-Off_Behaviors_Identification_Inkjet_DRL_GCN}
Li, Z., L.~J. Segura, Y.~Li, C.~Zhou, and H.~Sun (2023).
\newblock Multiclass reinforced active learning for droplet pinch-off behaviors identification in inkjet printing.
\newblock {\em Journal of Manufacturing Science and Engineering\/}~{\em 145\/}(7), 071002.

\bibitem[\protect\citeauthoryear{Li, Zhang, Zhang, Bai, Qin, Huang, Wang, Huang, Zhang, and Wen}{Li et~al.}{2023}]{Li_Zhang__Wen2023_LPBF_defect_classification_tansfer_learning_AE_Resnet50}
Li, Z., Z.~Zhang, S.~Zhang, Z.~Bai, R.~Qin, J.~Huang, J.~Wang, K.~Huang, Q.~Zhang, and G.~Wen (2023).
\newblock A novel approach of online monitoring for laser powder bed fusion defects: Air-borne acoustic emission and deep transfer learning.
\newblock {\em Journal of Manufacturing Processes\/}~{\em 102}, 579--592.

\bibitem[\protect\citeauthoryear{Liu, Kong, Babu, Joslin, and Ferguson}{Liu et~al.}{2021}]{liu2021integrated}
Liu, C., Z.~Kong, S.~Babu, C.~Joslin, and J.~Ferguson (2021).
\newblock An integrated manifold learning approach for high-dimensional data feature extractions and its applications to online process monitoring of additive manufacturing.
\newblock {\em IISE Transactions\/}~{\em 53\/}(11), 1215--1230.

\bibitem[\protect\citeauthoryear{Liu, Tian, and Kan}{Liu et~al.}{2022}]{Liu_Tian_Kan2022_when_ai_meets_AM_review}
Liu, C., W.~Tian, and C.~Kan (2022).
\newblock When ai meets additive manufacturing: Challenges and emerging opportunities for human-centered products development.
\newblock {\em Journal of Manufacturing Systems\/}~{\em 64}, 648--656.

\bibitem[\protect\citeauthoryear{Liu, Wang, Ho, Kong, Williams, Babu, and Joslin}{Liu et~al.}{2022}]{Liu_Wang_Ho_Kong__Chase2022}
Liu, C., R.~R. Wang, I.~Ho, Z.~J. Kong, C.~Williams, S.~Babu, and C.~Joslin (2022).
\newblock Toward online layer-wise surface morphology measurement in additive manufacturing using a deep learning-based approach.
\newblock {\em Journal of Intelligent Manufacturing\/}, 1--17.

\bibitem[\protect\citeauthoryear{Liu, Smaragdis, and Kim}{Liu et~al.}{2014}]{Liu_Smaragdis_Kim2014_supp_paper}
Liu, D., P.~Smaragdis, and M.~Kim (2014).
\newblock Experiments on deep learning for speech denoising.
\newblock In {\em Fifteenth Annual Conference of the International Speech Communication Association}.

\bibitem[\protect\citeauthoryear{Liu, Yuan, Peng, Wang, and Weiwei}{Liu et~al.}{2022}]{Liu_Yuan__Weiwei2022_meltpool_states_classification_DED_ResNet}
Liu, H., J.~Yuan, S.~Peng, F.~Wang, and L.~Weiwei (2022).
\newblock In-suit monitoring melt pool states in direct energy deposition using resnet.
\newblock {\em Measurement Science and Technology\/}~{\em 33\/}(12), 124007.

\bibitem[\protect\citeauthoryear{Liu, Liu, Bai, Rao, Williams, and Kong}{Liu et~al.}{2019}]{liu2019layer}
Liu, J., C.~Liu, Y.~Bai, P.~Rao, C.~B. Williams, and Z.~Kong (2019).
\newblock Layer-wise spatial modeling of porosity in additive manufacturing.
\newblock {\em IISE Transactions\/}~{\em 51\/}(2), 109--123.

\bibitem[\protect\citeauthoryear{Liu, Ye, Silva~Izquierdo, Vinel, Shamsaei, and Shao}{Liu et~al.}{2022}]{Liu_Ye_Izquierdo__Shao2022_review}
Liu, J., J.~Ye, D.~Silva~Izquierdo, A.~Vinel, N.~Shamsaei, and S.~Shao (2022).
\newblock A review of machine learning techniques for process and performance optimization in laser beam powder bed fusion additive manufacturing.
\newblock {\em Journal of Intelligent Manufacturing\/}, 1--27.

\bibitem[\protect\citeauthoryear{Liu, Wang, Tian, Lauria, and Liu}{Liu et~al.}{2021}]{Liu_Wang_Tian_Lauria_Liu2021}
Liu, W., Z.~Wang, L.~Tian, S.~Lauria, and X.~Liu (2021).
\newblock Melt pool segmentation for additive manufacturing: A generative adversarial network approach.
\newblock {\em Computers \& Electrical Engineering\/}~{\em 92}, 107183.

\bibitem[\protect\citeauthoryear{Lu, He, Shi, Bai, Zhao, and Han}{Lu et~al.}{2021}]{Lu_He_Shi_Bai_Zhao_Han2021}
Lu, J., H.~He, Y.~Shi, L.~Bai, Z.~Zhao, and J.~Han (2021).
\newblock Quantitative prediction for weld reinforcement in arc welding additive manufacturing based on molten pool image and deep residual network.
\newblock {\em Additive Manufacturing\/}~{\em 41}, 101980.

\bibitem[\protect\citeauthoryear{Luo, Ma, Xu, Li, and Cao}{Luo et~al.}{2021}]{Luo_Ma__Cao2021_1DCNN_LSTM_RNN_GRU_spatter_defect_classification_SLM}
Luo, S., X.~Ma, J.~Xu, M.~Li, and L.~Cao (2021).
\newblock Deep learning based monitoring of spatter behavior by the acoustic signal in selective laser melting.
\newblock {\em Sensors\/}~{\em 21\/}(21), 7179.

\bibitem[\protect\citeauthoryear{Lyu, Akhavan Taheri~Boroujeni, and Manoochehri}{Lyu et~al.}{2021}]{Lyu_Akhavan__Monoochehri2021_monitoring_anomaly}
Lyu, J., J.~Akhavan Taheri~Boroujeni, and S.~Manoochehri (2021).
\newblock In-situ laser-based process monitoring and in-plane surface anomaly identification for additive manufacturing using point cloud and machine learning.
\newblock In {\em International Design Engineering Technical Conferences and Computers and Information in Engineering Conference}, Volume 85376, pp.\  V002T02A030. American Society of Mechanical Engineers.

\bibitem[\protect\citeauthoryear{Mahadevan, Nath, and Hu}{Mahadevan et~al.}{2022}]{Mahadevan_Nath_Hu2022_UQ_AM_Review}
Mahadevan, S., P.~Nath, and Z.~Hu (2022).
\newblock Uncertainty quantification for additive manufacturing process improvement: Recent advances.
\newblock {\em ASCE-ASME Journal of Risk and Uncertainty in Engineering Systems, Part B: Mechanical Engineering\/}~{\em 8\/}(1), 010801.

\bibitem[\protect\citeauthoryear{Mahmoud, Magolon, Boer, Elbestawi, and Mohammadi}{Mahmoud et~al.}{2021}]{Mahmoud_Magolon__Mohammadi2021_review}
Mahmoud, D., M.~Magolon, J.~Boer, M.~Elbestawi, and M.~G. Mohammadi (2021).
\newblock Applications of machine learning in process monitoring and controls of l-pbf additive manufacturing: a review.
\newblock {\em Applied Sciences\/}~{\em 11\/}(24), 11910.

\bibitem[\protect\citeauthoryear{Manivannan}{Manivannan}{2023}]{Manivannan2023_powder_bed_defect_detection_SLS_CNN}
Manivannan, S. (2023).
\newblock Automatic quality inspection in additive manufacturing using semi-supervised deep learning.
\newblock {\em Journal of Intelligent Manufacturing\/}~{\em 34\/}(7), 3091--3108.

\bibitem[\protect\citeauthoryear{Mao, Lin, Yu, Frye, Beckett, Anderson, Jacquemetton, Carter, Gao, Liao, et~al.}{Mao et~al.}{2023}]{Mao_Lin_Yu_Frye_Beckett_Anderson___Agarwal2023}
Mao, Y., H.~Lin, C.~X. Yu, R.~Frye, D.~Beckett, K.~Anderson, L.~Jacquemetton, F.~Carter, Z.~Gao, W.-k. Liao, et~al. (2023).
\newblock A deep learning framework for layer-wise porosity prediction in metal powder bed fusion using thermal signatures.
\newblock {\em Journal of Intelligent Manufacturing\/}~{\em 34\/}(1), 315--329.

\bibitem[\protect\citeauthoryear{Mattera, Nele, and Paolella}{Mattera et~al.}{2023}]{Mattera_Nele_Paolella2023}
Mattera, G., L.~Nele, and D.~Paolella (2023).
\newblock Monitoring and control the wire arc additive manufacturing process using artificial intelligence techniques: A review.
\newblock {\em Journal of Intelligent Manufacturing\/}, 1--31.

\bibitem[\protect\citeauthoryear{Maurizi, Gao, and Berto}{Maurizi et~al.}{2022}]{Maurizi_Gao_Berto2022_stress_strain_deformation_prediction}
Maurizi, M., C.~Gao, and F.~Berto (2022).
\newblock Predicting stress, strain and deformation fields in materials and structures with graph neural networks.
\newblock {\em Scientific Reports\/}~{\em 12\/}(1), 21834.

\bibitem[\protect\citeauthoryear{McGowan, Gawade, and Guo}{McGowan et~al.}{2022}]{McGowan_Gawade_Guo2022PICNN_porosity_prediction_LMD}
McGowan, E., V.~Gawade, and W.~Guo (2022).
\newblock A physics-informed convolutional neural network with custom loss functions for porosity prediction in laser metal deposition.
\newblock {\em Sensors\/}~{\em 22\/}(2), 494.

\bibitem[\protect\citeauthoryear{Mehta and Shao}{Mehta and Shao}{2022}]{Mehta_Shao2022_Federated_defect_detection}
Mehta, M. and C.~Shao (2022).
\newblock Federated learning-based semantic segmentation for pixel-wise defect detection in additive manufacturing.
\newblock {\em Journal of Manufacturing Systems\/}~{\em 64}, 197--210.

\bibitem[\protect\citeauthoryear{Meng, McWilliams, Jarosinski, Park, Jung, Lee, and Zhang}{Meng et~al.}{2020}]{Meng_McWilliams__Zhang2020_review}
Meng, L., B.~McWilliams, W.~Jarosinski, H.-Y. Park, Y.-G. Jung, J.~Lee, and J.~Zhang (2020).
\newblock Machine learning in additive manufacturing: a review.
\newblock {\em Jom\/}~{\em 72}, 2363--2377.

\bibitem[\protect\citeauthoryear{Mi, Zhang, Li, Shen, Yang, Song, Zhou, Duan, Lu, and Mai}{Mi et~al.}{2023a}]{Mi_Zhang_Li_Shen__Mai2023}
Mi, J., Y.~Zhang, H.~Li, S.~Shen, Y.~Yang, C.~Song, X.~Zhou, Y.~Duan, J.~Lu, and H.~Mai (2023a).
\newblock In-situ monitoring laser based directed energy deposition process with deep convolutional neural network.
\newblock {\em Journal of Intelligent Manufacturing\/}, 683--693.

\bibitem[\protect\citeauthoryear{Mi, Zhang, Li, Shen, Yang, Song, Zhou, Duan, Lu, and Mai}{Mi et~al.}{2023b}]{Mi_Zhang_Li_Shen__Mai2023_DED_CNN_spatter_detection}
Mi, J., Y.~Zhang, H.~Li, S.~Shen, Y.~Yang, C.~Song, X.~Zhou, Y.~Duan, J.~Lu, and H.~Mai (2023b).
\newblock In-situ monitoring laser based directed energy deposition process with deep convolutional neural network.
\newblock {\em Journal of Intelligent Manufacturing\/}, 1--11.

\bibitem[\protect\citeauthoryear{Min, Ming, Peihong, Yang, Zhang, Lingfeng, Liucheng, Yinghong, and Wanlin}{Min et~al.}{2023}]{Yi_Xue_Cong__Guo2023_review_fatigue_prediction}
Min, Y., X.~Ming, C.~Peihong, S.~Yang, H.~Zhang, W.~Lingfeng, Z.~Liucheng, L.~Yinghong, and G.~Wanlin (2023).
\newblock Machine learning for predicting fatigue properties of additively manufactured materials.
\newblock {\em Chinese Journal of Aeronautics\/}.

\bibitem[\protect\citeauthoryear{Mohammadi, Mahmoud, and Elbestawi}{Mohammadi et~al.}{2021}]{Mohammadi_Mahmoud_Elbestawi2021}
Mohammadi, M.~G., D.~Mahmoud, and M.~Elbestawi (2021).
\newblock On the application of machine learning for defect detection in l-pbf additive manufacturing.
\newblock {\em Optics \& Laser Technology\/}~{\em 143}, 107338.

\bibitem[\protect\citeauthoryear{Mohammed, Almutahhar, Sattar, Alhajeri, Nazir, and Ali}{Mohammed et~al.}{2023}]{Mohammed_Almutahhar__Ali2023_porosity_prediction_LPBF_DNN}
Mohammed, A.~S., M.~Almutahhar, K.~Sattar, A.~Alhajeri, A.~Nazir, and U.~Ali (2023).
\newblock Deep learning based porosity prediction for additively manufactured laser powder-bed fusion parts.
\newblock {\em Journal of Materials Research and Technology\/}~{\em 27}, 7330--7335.

\bibitem[\protect\citeauthoryear{Mozaffar, Ebrahimi, and Cao}{Mozaffar et~al.}{2020}]{Mozaffar_Ebrahimi_Cao2020toolpath}
Mozaffar, M., A.~Ebrahimi, and J.~Cao (2020).
\newblock Toolpath design for additive manufacturing using deep reinforcement learning.
\newblock {\em arXiv preprint arXiv:2009.14365\/}.

\bibitem[\protect\citeauthoryear{Mozaffar, Liao, Lin, Ehmann, and Cao}{Mozaffar et~al.}{2021}]{Mozaffar_Liao_Lin_Ehmann_Cao2021}
Mozaffar, M., S.~Liao, H.~Lin, K.~Ehmann, and J.~Cao (2021).
\newblock Geometry-agnostic data-driven thermal modeling of additive manufacturing processes using graph neural networks.
\newblock {\em Additive Manufacturing\/}~{\em 48}, 102449.

\bibitem[\protect\citeauthoryear{Mozaffar, Paul, Al-Bahrani, Wolff, Choudhary, Agrawal, Ehmann, and Cao}{Mozaffar et~al.}{2018}]{Mozaffar_Paul__Cao2018}
Mozaffar, M., A.~Paul, R.~Al-Bahrani, S.~Wolff, A.~Choudhary, A.~Agrawal, K.~Ehmann, and J.~Cao (2018).
\newblock Data-driven prediction of the high-dimensional thermal history in directed energy deposition processes via recurrent neural networks.
\newblock {\em Manufacturing letters\/}~{\em 18}, 35--39.

\bibitem[\protect\citeauthoryear{Nalajam and Varadarajan}{Nalajam and Varadarajan}{2021}]{Nalajam_Varadarajan2021}
Nalajam, P.~K. and R.~Varadarajan (2021).
\newblock A hybrid deep learning model for layer-wise melt pool temperature forecasting in wire-arc additive manufacturing process.
\newblock {\em IEEE Access\/}~{\em 9}, 100652--100664.

\bibitem[\protect\citeauthoryear{Ogoke and Farimani}{Ogoke and Farimani}{2021}]{Ogoke_Farimani2021}
Ogoke, F. and A.~B. Farimani (2021).
\newblock Thermal control of laser powder bed fusion using deep reinforcement learning.
\newblock {\em Additive Manufacturing\/}~{\em 46}, 102033.

\bibitem[\protect\citeauthoryear{Ogoke, Johnson, Glinsky, Laursen, Kramer, and Farimani}{Ogoke et~al.}{2022}]{Ogoke_Johnson_Glinsky__Farimani2022}
Ogoke, O.~F., K.~Johnson, M.~Glinsky, C.~Laursen, S.~Kramer, and A.~B. Farimani (2022).
\newblock Deep-learned generators of porosity distributions produced during metal additive manufacturing.
\newblock {\em Additive Manufacturing\/}~{\em 60}, 103250.

\bibitem[\protect\citeauthoryear{Oster, Breese, Ulbricht, Mohr, and Altenburg}{Oster et~al.}{2023}]{Oster_Breese__Altenburg2023_defect_prediction_LPBF}
Oster, S., P.~P. Breese, A.~Ulbricht, G.~Mohr, and S.~J. Altenburg (2023).
\newblock A deep learning framework for defect prediction based on thermographic in-situ monitoring in laser powder bed fusion.
\newblock {\em Journal of Intelligent Manufacturing\/}, 1--20.

\bibitem[\protect\citeauthoryear{Page, McKenzie, Bossuyt, Boutron, Hoffmann, Mulrow, Shamseer, Tetzlaff, Akl, Brennan, et~al.}{Page et~al.}{2021}]{page2021prisma_Prisma_Reference}
Page, M.~J., J.~E. McKenzie, P.~M. Bossuyt, I.~Boutron, T.~C. Hoffmann, C.~D. Mulrow, L.~Shamseer, J.~M. Tetzlaff, E.~A. Akl, S.~E. Brennan, et~al. (2021).
\newblock The prisma 2020 statement: an updated guideline for reporting systematic reviews.
\newblock {\em Bmj\/}~{\em 372}.

\bibitem[\protect\citeauthoryear{Pandita, Ghosh, Gupta, Meshkov, and Wang}{Pandita et~al.}{2022}]{Pandita_Ghosh_Gupta__Wang2022_process_modeling}
Pandita, P., S.~Ghosh, V.~K. Gupta, A.~Meshkov, and L.~Wang (2022).
\newblock Application of deep transfer learning and uncertainty quantification for process identification in powder bed fusion.
\newblock {\em ASCE-ASME Journal of Risk and Uncertainty in Engineering Systems, Part B: Mechanical Engineering\/}~{\em 8\/}(1), 011106.

\bibitem[\protect\citeauthoryear{Pandiyan, Cui, Le-Quang, Deshpande, Wasmer, and Shevchik}{Pandiyan et~al.}{2022}]{Pandiyan_Cui__Shevchik2022}
Pandiyan, V., D.~Cui, T.~Le-Quang, P.~Deshpande, K.~Wasmer, and S.~Shevchik (2022).
\newblock In situ quality monitoring in direct energy deposition process using co-axial process zone imaging and deep contrastive learning.
\newblock {\em Journal of Manufacturing Processes\/}~{\em 81}, 1064--1075.

\bibitem[\protect\citeauthoryear{Pandiyan, Cui, Parrilli, Deshpande, Masinelli, Shevchik, and Wasmer}{Pandiyan et~al.}{2022}]{Pandiyan_Cui__Wasmer2022_DED_meltpool_state_prediction_meltpool_image_GAN_CNN}
Pandiyan, V., D.~Cui, A.~Parrilli, P.~Deshpande, G.~Masinelli, S.~Shevchik, and K.~Wasmer (2022).
\newblock Monitoring of direct energy deposition process using manifold learning and co-axial melt pool imaging.
\newblock {\em Manufacturing Letters\/}~{\em 33}, 776--785.

\bibitem[\protect\citeauthoryear{Pandiyan, Cui, Richter, Parrilli, and Leparoux}{Pandiyan et~al.}{2023}]{Pandiyan_Cui__Leparoux2023_LDED_meltpool_image_and_AE_CNN_ViT}
Pandiyan, V., D.~Cui, R.~A. Richter, A.~Parrilli, and M.~Leparoux (2023).
\newblock Real-time monitoring and quality assurance for laser-based directed energy deposition: integrating co-axial imaging and self-supervised deep learning framework.
\newblock {\em Journal of Intelligent Manufacturing\/}, 1--25.

\bibitem[\protect\citeauthoryear{Pandiyan, Drissi-Daoudi, Shevchik, Masinelli, Le-Quang, Log{\'e}, and Wasmer}{Pandiyan et~al.}{2021}]{Pandiyan_Drissi-Daoudi_Shevchik__Wasmer2021_semi_supervised_monitoring}
Pandiyan, V., R.~Drissi-Daoudi, S.~Shevchik, G.~Masinelli, T.~Le-Quang, R.~Log{\'e}, and K.~Wasmer (2021).
\newblock Semi-supervised monitoring of laser powder bed fusion process based on acoustic emissions.
\newblock {\em Virtual and Physical Prototyping\/}~{\em 16\/}(4), 481--497.

\bibitem[\protect\citeauthoryear{Pandiyan, Drissi-Daoudi, Shevchik, Masinelli, Le-Quang, Log{\'e}, and Wasmer}{Pandiyan et~al.}{2022}]{Pandiyan_Drissi-Daoudi_Wasmer2022}
Pandiyan, V., R.~Drissi-Daoudi, S.~Shevchik, G.~Masinelli, T.~Le-Quang, R.~Log{\'e}, and K.~Wasmer (2022).
\newblock Deep transfer learning of additive manufacturing mechanisms across materials in metal-based laser powder bed fusion process.
\newblock {\em Journal of Materials Processing Technology\/}~{\em 303}, 117531.

\bibitem[\protect\citeauthoryear{Pandiyan, Masinelli, Claire, Le-Quang, Hamidi-Nasab, de~Formanoir, Esmaeilzadeh, Goel, Marone, Log{\'e}, et~al.}{Pandiyan et~al.}{2022}]{Pandiyan_Masinelli_Claire__Wasmer2022}
Pandiyan, V., G.~Masinelli, N.~Claire, T.~Le-Quang, M.~Hamidi-Nasab, C.~de~Formanoir, R.~Esmaeilzadeh, S.~Goel, F.~Marone, R.~Log{\'e}, et~al. (2022).
\newblock Deep learning-based monitoring of laser powder bed fusion process on variable time-scales using heterogeneous sensing and operando x-ray radiography guidance.
\newblock {\em Additive Manufacturing\/}~{\em 58}, 103007.

\bibitem[\protect\citeauthoryear{Pandiyan, Wr{\'o}bel, Richter, Leparoux, Leinenbach, and Shevchik}{Pandiyan et~al.}{2024}]{Pandiyan_Wrobel__Shevchik_2024_domainAdaptation_LPBF_CNN_AE}
Pandiyan, V., R.~Wr{\'o}bel, R.~A. Richter, M.~Leparoux, C.~Leinenbach, and S.~Shevchik (2024).
\newblock Monitoring of laser powder bed fusion process by bridging dissimilar process maps using deep learning-based domain adaptation on acoustic emissions.
\newblock {\em Additive Manufacturing\/}~{\em 80}, 103974.

\bibitem[\protect\citeauthoryear{Park, Choi, and Um}{Park et~al.}{2024}]{Park_Choi_Um2024_ConvLSTM_meltpool_prediction_from_images_of_laser_tool_path_strategy_LPBF}
Park, J.~M., M.~Choi, and J.~Um (2024).
\newblock Convolutional lstm based melt-pool prediction from images of laser tool path strategy in laser powder bed fusion for additive manufacturing.
\newblock {\em The International Journal of Advanced Manufacturing Technology\/}~{\em 130\/}(3), 1871--1886.

\bibitem[\protect\citeauthoryear{Park, Choi, and Jhang}{Park et~al.}{2021}]{Park_Choi_Jhang2021}
Park, S.-H., S.~Choi, and K.-Y. Jhang (2021).
\newblock Porosity evaluation of additively manufactured components using deep learning-based ultrasonic nondestructive testing.
\newblock {\em International Journal of Precision Engineering and Manufacturing-Green Technology\/}, 1--13.

\bibitem[\protect\citeauthoryear{Park and Lee}{Park and Lee}{2016}]{Park_Lee2016_supp}
Park, S.~R. and J.~Lee (2016).
\newblock A fully convolutional neural network for speech enhancement.
\newblock {\em arXiv preprint arXiv:1609.07132\/}.

\bibitem[\protect\citeauthoryear{Peng, Liu, Huang, Cao, Liu, and Lu}{Peng et~al.}{2022}]{Peng_Liu_Huang__Lu2022}
Peng, H., A.~Liu, J.~Huang, L.~Cao, J.~Liu, and L.~Lu (2022).
\newblock Ph-net: Parallelepiped microstructure homogenization via 3d convolutional neural networks.
\newblock {\em Additive Manufacturing\/}~{\em 60}, 103237.

\bibitem[\protect\citeauthoryear{Perani, Baraldo, Decker, Vandone, Valente, and Paoli}{Perani et~al.}{2023}]{Perani_Baraldo__Paoli2023}
Perani, M., S.~Baraldo, M.~Decker, A.~Vandone, A.~Valente, and B.~Paoli (2023).
\newblock Track geometry prediction for laser metal deposition based on on-line artificial vision and deep neural networks.
\newblock {\em Robotics and Computer-Integrated Manufacturing\/}~{\em 79}, 102445.

\bibitem[\protect\citeauthoryear{Perani, Jandl, Baraldo, Valente, and Paoli}{Perani et~al.}{2023}]{Perani_Jandl_Baraldo_Valente_Paoli2023_modeling_track_geometry}
Perani, M., R.~Jandl, S.~Baraldo, A.~Valente, and B.~Paoli (2023).
\newblock Long-short term memory networks for modeling track geometry in laser metal deposition.
\newblock {\em Frontiers in Artificial Intelligence\/}~{\em 6}, 1156630.

\bibitem[\protect\citeauthoryear{Perumal, Abueidda, Koric, and Kontsos}{Perumal et~al.}{2023}]{Perumal_Abueidda__Kontsos2023_TCN_data-driven_thermal_prediction_DED}
Perumal, V., D.~Abueidda, S.~Koric, and A.~Kontsos (2023).
\newblock Temporal convolutional networks for data-driven thermal modeling of directed energy deposition.
\newblock {\em Journal of Manufacturing Processes\/}~{\em 85}, 405--416.

\bibitem[\protect\citeauthoryear{Petrik and Bambach}{Petrik and Bambach}{2023}]{Petrik_Bambach2023_path_planning_reinforcement}
Petrik, J. and M.~Bambach (2023).
\newblock Reinforcement learning and optimization based path planning for thin-walled structures in wire arc additive manufacturing.
\newblock {\em Journal of Manufacturing Processes\/}~{\em 93}, 75--89.

\bibitem[\protect\citeauthoryear{Pham, Hoang, Tran, Fetni, Duch{\^e}ne, Tran, and Habraken}{Pham et~al.}{2022}]{Pham_Hoang__Habraken2022_uncertainties_DED}
Pham, T., T.~Hoang, X.~Tran, S.~Fetni, L.~Duch{\^e}ne, H.~S. Tran, and A.~Habraken (2022).
\newblock Characterization, propagation, and sensitivity analysis of uncertainties in the directed energy deposition process using a deep learning-based surrogate model.
\newblock {\em Probabilistic Engineering Mechanics\/}~{\em 69}, 103297.

\bibitem[\protect\citeauthoryear{Pham, Hoang, Van~Tran, Pham, Fetni, Duch{\^e}ne, Tran, and Habraken}{Pham et~al.}{2023}]{Pham_Hoang_Tran_Pham_Fetni_Duchene_Tran_Habraken2023}
Pham, T. Q.~D., T.~V. Hoang, X.~Van~Tran, Q.~T. Pham, S.~Fetni, L.~Duch{\^e}ne, H.~S. Tran, and A.-M. Habraken (2023).
\newblock Fast and accurate prediction of temperature evolutions in additive manufacturing process using deep learning.
\newblock {\em Journal of Intelligent Manufacturing\/}~{\em 34\/}(4), 1701--1719.

\bibitem[\protect\citeauthoryear{Piovarci, Foshey, Xu, Erps, Babaei, Didyk, Rusinkiewicz, Matusik, and Bickel}{Piovarci et~al.}{2022}]{Piovarci2022closedLoop_Control_reinforcement}
Piovarci, M., M.~Foshey, J.~Xu, T.~Erps, V.~Babaei, P.~Didyk, S.~Rusinkiewicz, W.~Matusik, and B.~Bickel (2022).
\newblock Closed-loop control of direct ink writing via reinforcement learning.
\newblock {\em arXiv preprint arXiv:2201.11819\/}.

\bibitem[\protect\citeauthoryear{Prince}{Prince}{2023}]{Prince_2023_Understanding_deep_learning_MIT_press}
Prince, S.~J. (2023).
\newblock {\em Understanding deep learning}.
\newblock MIT press.

\bibitem[\protect\citeauthoryear{Qi, Su, Mo, and Guibas}{Qi et~al.}{2017}]{qi2017_pointnet_supp}
Qi, C.~R., H.~Su, K.~Mo, and L.~J. Guibas (2017).
\newblock Pointnet: Deep learning on point sets for 3d classification and segmentation.
\newblock In {\em Proceedings of the IEEE conference on computer vision and pattern recognition}, pp.\  652--660.

\bibitem[\protect\citeauthoryear{Qi, Yi, Su, and Guibas}{Qi et~al.}{2017}]{qi2017pointnet++}
Qi, C.~R., L.~Yi, H.~Su, and L.~J. Guibas (2017).
\newblock Pointnet++: Deep hierarchical feature learning on point sets in a metric space.
\newblock {\em Advances in neural information processing systems\/}~{\em 30}.

\bibitem[\protect\citeauthoryear{Qi, Chen, Li, Cheng, and Li}{Qi et~al.}{2019}]{Qi_Chen_Li_Cheng_Li2019_review}
Qi, X., G.~Chen, Y.~Li, X.~Cheng, and C.~Li (2019).
\newblock Applying neural-network-based machine learning to additive manufacturing: current applications, challenges, and future perspectives.
\newblock {\em Engineering\/}~{\em 5\/}(4), 721--729.

\bibitem[\protect\citeauthoryear{Qin, Hu, Liu, Witherell, Wang, Rosen, Simpson, Lu, and Tang}{Qin et~al.}{2022}]{Qin_Hu_Liu__Tang2022_review}
Qin, J., F.~Hu, Y.~Liu, P.~Witherell, C.~C. Wang, D.~W. Rosen, T.~W. Simpson, Y.~Lu, and Q.~Tang (2022).
\newblock Research and application of machine learning for additive manufacturing.
\newblock {\em Additive Manufacturing\/}~{\em 52}, 102691.

\bibitem[\protect\citeauthoryear{Qin, Ding, Qu, Song, Wang, and Liao}{Qin et~al.}{2024}]{Qin_Ding__Liao2024_DRL_toolpath_generation_thermal_uniformity_LPBF}
Qin, M., J.~Ding, S.~Qu, X.~Song, C.~C. Wang, and W.-H. Liao (2024).
\newblock Deep reinforcement learning based toolpath generation for thermal uniformity in laser powder bed fusion process.
\newblock {\em Additive Manufacturing\/}~{\em 79}, 103937.

\bibitem[\protect\citeauthoryear{Qin, DeWitt, Radhakrishnan, and Biros}{Qin et~al.}{2023}]{Qin_DeWitt_Radhakrishnan_Biros2023}
Qin, Y., S.~DeWitt, B.~Radhakrishnan, and G.~Biros (2023).
\newblock Grainnn: A neighbor-aware long short-term memory network for predicting microstructure evolution during polycrystalline grain formation.
\newblock {\em Computational Materials Science\/}~{\em 218}, 111927.

\bibitem[\protect\citeauthoryear{Radosavovic, Kosaraju, Girshick, He, and Doll{\'a}r}{Radosavovic et~al.}{2020}]{Radosavovic_Kosaraju_Girshick_He_Dollar2020_designing_network_design_spaces}
Radosavovic, I., R.~P. Kosaraju, R.~Girshick, K.~He, and P.~Doll{\'a}r (2020).
\newblock Designing network design spaces.
\newblock In {\em Proceedings of the IEEE/CVF conference on computer vision and pattern recognition}, pp.\  10428--10436.

\bibitem[\protect\citeauthoryear{Ramlatchan and Li}{Ramlatchan and Li}{2022}]{Ramlatchan_Li2022}
Ramlatchan, A. and Y.~Li (2022).
\newblock Image synthesis using conditional gans for selective laser melting additive manufacturing.
\newblock In {\em 2022 International Joint Conference on Neural Networks (IJCNN)}, pp.\  1--8. IEEE.

\bibitem[\protect\citeauthoryear{Rashed, Ashraf, Mines, and Hazell}{Rashed et~al.}{2016}]{Rashed_Ashraf_Mines_Hazell2016_Supp}
Rashed, M., M.~Ashraf, R.~Mines, and P.~J. Hazell (2016).
\newblock Metallic microlattice materials: A current state of the art on manufacturing, mechanical properties and applications.
\newblock {\em Materials \& Design\/}~{\em 95}, 518--533.

\bibitem[\protect\citeauthoryear{Razvi, Feng, Narayanan, Lee, and Witherell}{Razvi et~al.}{2019}]{Razvi_Feng_Narayanan_Lee_Witherrell2019_review}
Razvi, S.~S., S.~Feng, A.~Narayanan, Y.-T.~T. Lee, and P.~Witherell (2019).
\newblock A review of machine learning applications in additive manufacturing.
\newblock In {\em International design engineering technical conferences and computers and information in engineering conference}, Volume 59179, pp.\  V001T02A040. American Society of Mechanical Engineers.

\bibitem[\protect\citeauthoryear{Redmon, Divvala, Girshick, and Farhadi}{Redmon et~al.}{2016}]{Redmon_Divvala_Girshick_Farhadi2016_YOLO_supp_paper}
Redmon, J., S.~Divvala, R.~Girshick, and A.~Farhadi (2016).
\newblock You only look once: Unified, real-time object detection.
\newblock In {\em Proceedings of the IEEE conference on computer vision and pattern recognition}, pp.\  779--788.

\bibitem[\protect\citeauthoryear{Ren, Chew, Liu, Zhang, Fuh, and Bi}{Ren et~al.}{2021}]{Ren_Chew_Liu__Bi2021_toolpath_planning}
Ren, K., Y.~Chew, N.~Liu, Y.~Zhang, J.~Fuh, and G.~Bi (2021).
\newblock Integrated numerical modelling and deep learning for multi-layer cube deposition planning in laser aided additive manufacturing.
\newblock {\em Virtual and Physical Prototyping\/}~{\em 16\/}(3), 318--332.

\bibitem[\protect\citeauthoryear{Ren, Chew, Zhang, Fuh, and Bi}{Ren et~al.}{2020}]{Ren_Chew_Zhang_Fuh_Bi2020}
Ren, K., Y.~Chew, Y.~Zhang, J.~Fuh, and G.~Bi (2020).
\newblock Thermal field prediction for laser scanning paths in laser aided additive manufacturing by physics-based machine learning.
\newblock {\em Computer Methods in Applied Mechanics and Engineering\/}~{\em 362}, 112734.

\bibitem[\protect\citeauthoryear{Ren, Wen, Zhang, and Mazumder}{Ren et~al.}{2022}]{Ren_Wen_Zhang_Mazumder2022_quality_monitoring}
Ren, W., G.~Wen, Z.~Zhang, and J.~Mazumder (2022).
\newblock Quality monitoring in additive manufacturing using emission spectroscopy and unsupervised deep learning.
\newblock {\em Materials and Manufacturing Processes\/}~{\em 37\/}(11), 1339--1346.

\bibitem[\protect\citeauthoryear{Research}{Research}{2023}]{AM_market_size_reference}
Research, P. (2023).
\newblock 3d printing market size is expanding to usd 98,310 million by 2032.
\newblock Available at : \url{https://finance.yahoo.com/news/3d-printing-market-size-expanding-144000417.html} (accessed on June 1, 2024).

\bibitem[\protect\citeauthoryear{Rezasefat and Hogan}{Rezasefat and Hogan}{2024}]{Rezasefat_Hogan2024_Prediction_4D_stress_evolution_SLM_CNN_voxel}
Rezasefat, M. and J.~D. Hogan (2024).
\newblock Prediction of 4d stress field evolution around additive manufacturing-induced porosity through progressive deep-learning frameworks.
\newblock {\em Machine Learning: Science and Technology\/}.

\bibitem[\protect\citeauthoryear{Romera, Alvarez, Bergasa, and Arroyo}{Romera et~al.}{2017}]{Romera_Alvarez_Bergasa_Arroyo2017_ERFNet}
Romera, E., J.~M. Alvarez, L.~M. Bergasa, and R.~Arroyo (2017).
\newblock Erfnet: Efficient residual factorized convnet for real-time semantic segmentation.
\newblock {\em IEEE Transactions on Intelligent Transportation Systems\/}~{\em 19\/}(1), 263--272.

\bibitem[\protect\citeauthoryear{Ruiz, Jafari, Venkata~Subramanian, Vaneker, Ya, and Huang}{Ruiz et~al.}{2022}]{ruiz2022prediction_decompose_deviation_into_global_local_roughness_supp}
Ruiz, C., D.~Jafari, V.~Venkata~Subramanian, T.~H. Vaneker, W.~Ya, and Q.~Huang (2022).
\newblock Prediction and control of product shape quality for wire and arc additive manufacturing.
\newblock {\em Journal of Manufacturing Science and Engineering\/}~{\em 144\/}(11), 111005.

\bibitem[\protect\citeauthoryear{Saluja, Xie, and Fayazbakhsh}{Saluja et~al.}{2020}]{Saluja_Xie_Fayazbakhsh2020}
Saluja, A., J.~Xie, and K.~Fayazbakhsh (2020).
\newblock A closed-loop in-process warping detection system for fused filament fabrication using convolutional neural networks.
\newblock {\em Journal of Manufacturing Processes\/}~{\em 58}, 407--415.

\bibitem[\protect\citeauthoryear{Sarkon, Safaei, Kenevisi, Arman, and Zeeshan}{Sarkon et~al.}{2022}]{Sarkon_Safaei__Zeeshan_2022_review}
Sarkon, G.~K., B.~Safaei, M.~S. Kenevisi, S.~Arman, and Q.~Zeeshan (2022).
\newblock State-of-the-art review of machine learning applications in additive manufacturing; from design to manufacturing and property control.
\newblock {\em Archives of Computational Methods in Engineering\/}~{\em 29\/}(7), 5663--5721.

\bibitem[\protect\citeauthoryear{Schmid, Krabusch, Schromm, Jieqing, Ziegelmeier, Grosse, and Schleifenbaum}{Schmid et~al.}{2021}]{Schmid_Krabusch_Schromm__Schleifenbaum2021}
Schmid, S., J.~Krabusch, T.~Schromm, S.~Jieqing, S.~Ziegelmeier, C.~U. Grosse, and J.~H. Schleifenbaum (2021).
\newblock A new approach for automated measuring of the melt pool geometry in laser-powder bed fusion.
\newblock {\em Progress in Additive Manufacturing\/}~{\em 6}, 269--279.

\bibitem[\protect\citeauthoryear{Schmidhuber}{Schmidhuber}{2015}]{Schmidhuber2015_DL_in_NN_ref_review}
Schmidhuber, J. (2015).
\newblock Deep learning in neural networks: An overview.
\newblock {\em Neural networks\/}~{\em 61}, 85--117.

\bibitem[\protect\citeauthoryear{Scime and Beuth}{Scime and Beuth}{2018}]{Scime_Beuth2018}
Scime, L. and J.~Beuth (2018).
\newblock A multi-scale convolutional neural network for autonomous anomaly detection and classification in a laser powder bed fusion additive manufacturing process.
\newblock {\em Additive Manufacturing\/}~{\em 24}, 273--286.

\bibitem[\protect\citeauthoryear{Scime, Joslin, Collins, Sprayberry, Singh, Halsey, Duncan, Snow, Dehoff, and Paquit}{Scime et~al.}{2023}]{Scime_Joslin__Paquit2023_Tensile_Property_Prediction_LPBF}
Scime, L., C.~Joslin, D.~A. Collins, M.~Sprayberry, A.~Singh, W.~Halsey, R.~Duncan, Z.~Snow, R.~Dehoff, and V.~Paquit (2023).
\newblock A data-driven framework for direct local tensile property prediction of laser powder bed fusion parts.
\newblock {\em Materials\/}~{\em 16\/}(23), 7293.

\bibitem[\protect\citeauthoryear{Scime, Siddel, Baird, and Paquit}{Scime et~al.}{2020}]{Scime_Siddel_Baird_Paquit2020}
Scime, L., D.~Siddel, S.~Baird, and V.~Paquit (2020).
\newblock Layer-wise anomaly detection and classification for powder bed additive manufacturing processes: A machine-agnostic algorithm for real-time pixel-wise semantic segmentation.
\newblock {\em Additive Manufacturing\/}~{\em 36}, 101453.

\bibitem[\protect\citeauthoryear{Segura, Li, Zhou, and Sun}{Segura et~al.}{2023}]{Segura_Li__Sun2023_Tensor_time_series_TGCN_TRNN_Material_jetting}
Segura, L.~J., Z.~Li, C.~Zhou, and H.~Sun (2023).
\newblock Droplet evolution prediction in material jetting via tensor time series analysis.
\newblock {\em Additive Manufacturing\/}~{\em 66}, 103461.

\bibitem[\protect\citeauthoryear{Senanayaka, Tian, Falls, and Bian}{Senanayaka et~al.}{2023}]{Senanayaka_Tian_Falls_Bian2023_Porosity_prediction}
Senanayaka, A., W.~Tian, T.~Falls, and L.~Bian (2023).
\newblock Understanding the effects of process conditions on thermal--defect relationship: A transfer machine learning approach.
\newblock {\em Journal of Manufacturing Science and Engineering\/}~{\em 145\/}(7), 071010.

\bibitem[\protect\citeauthoryear{Sharma, Raissi, and Guo}{Sharma et~al.}{2023}]{Sharma_Raissi_Guo2023_PIDL_multi-physical_PBF}
Sharma, R., M.~Raissi, and Y.~Guo (2023).
\newblock Physics-informed deep learning of gas flow-melt pool multi-physical dynamics during powder bed fusion.
\newblock {\em CIRP Annals\/}~{\em 72\/}(1), 161--164.

\bibitem[\protect\citeauthoryear{Shen, Shang, Li, Bao, Zhang, Dong, Wan, Xiong, and Wang}{Shen et~al.}{2019}]{Shen_Shang_Li__Wang2019_error_prediction_compensation}
Shen, Z., X.~Shang, Y.~Li, Y.~Bao, X.~Zhang, X.~Dong, L.~Wan, G.~Xiong, and F.-Y. Wang (2019).
\newblock Prednet and compnet: Prediction and high-precision compensation of in-plane shape deformation for additive manufacturing.
\newblock In {\em 2019 IEEE 15th International Conference on Automation Science and Engineering (CASE)}, pp.\  462--467. IEEE.

\bibitem[\protect\citeauthoryear{Shen, Shang, Zhao, Dong, Xiong, and Wang}{Shen et~al.}{2019}]{Shen_Shang_Zhao_Dong_Xiong_Wang2019_error_compensation}
Shen, Z., X.~Shang, M.~Zhao, X.~Dong, G.~Xiong, and F.-Y. Wang (2019).
\newblock A learning-based framework for error compensation in 3d printing.
\newblock {\em IEEE transactions on cybernetics\/}~{\em 49\/}(11), 4042--4050.

\bibitem[\protect\citeauthoryear{Shevchik, Masinelli, Kenel, Leinenbach, and Wasmer}{Shevchik et~al.}{2019}]{Shevchik_Masinelli__WAsmer2019_SCNN_quality_level_classification_PBF}
Shevchik, S.~A., G.~Masinelli, C.~Kenel, C.~Leinenbach, and K.~Wasmer (2019).
\newblock Deep learning for in situ and real-time quality monitoring in additive manufacturing using acoustic emission.
\newblock {\em IEEE Transactions on Industrial Informatics\/}~{\em 15\/}(9), 5194--5203.

\bibitem[\protect\citeauthoryear{Shi, Kan, Tian, and Liu}{Shi et~al.}{2021}]{Shi_Kan_Tian_Liu2021_Blockchain_security}
Shi, Z., C.~Kan, W.~Tian, and C.~Liu (2021).
\newblock A blockchain-based g-code protection approach for cyber-physical security in additive manufacturing.
\newblock {\em Journal of Computing and Information Science in Engineering\/}~{\em 21\/}(4), 041007.

\bibitem[\protect\citeauthoryear{Shi, Mamun, Kan, Tian, and Liu}{Shi et~al.}{2022}]{shi_Mamun__Liu2022_LSTM_autoencoder}
Shi, Z., A.~A. Mamun, C.~Kan, W.~Tian, and C.~Liu (2022).
\newblock An lstm-autoencoder based online side channel monitoring approach for cyber-physical attack detection in additive manufacturing.
\newblock {\em Journal of Intelligent Manufacturing\/}, 1--17.

\bibitem[\protect\citeauthoryear{Shi, Mandal, Harimkar, and Liu}{Shi et~al.}{2022}]{shi2022hybrid}
Shi, Z., S.~Mandal, S.~Harimkar, and C.~Liu (2022).
\newblock Hybrid data-driven feature extraction-enabled surface modeling for metal additive manufacturing.
\newblock {\em The International Journal of Advanced Manufacturing Technology\/}~{\em 121\/}(7-8), 4643--4662.

\bibitem[\protect\citeauthoryear{Shin, Hong, Jadhav, and Kim}{Shin et~al.}{2023}]{Shin_Hong_Jadhav_KIm2023_defect_detection_transfer_learning}
Shin, S.-J., S.-H. Hong, S.~Jadhav, and D.~B. Kim (2023).
\newblock Detecting balling defects using multisource transfer learning in wire arc additive manufacturing.
\newblock {\em Journal of Computational Design and Engineering\/}~{\em 10\/}(4), 1423--1442.

\bibitem[\protect\citeauthoryear{Siegkas}{Siegkas}{2022}]{Petros_Siegkas2022}
Siegkas, P. (2022).
\newblock Generating 3d porous structures using machine learning and additive manufacturing.
\newblock {\em Materials \& Design\/}~{\em 220}, 110858.

\bibitem[\protect\citeauthoryear{Simonyan and Zisserman}{Simonyan and Zisserman}{2014}]{Simonyan_Zisserman2014_VGG16}
Simonyan, K. and A.~Zisserman (2014).
\newblock Very deep convolutional networks for large-scale image recognition.
\newblock {\em arXiv preprint arXiv:1409.1556\/}.

\bibitem[\protect\citeauthoryear{Sing, Kuo, Shih, Ho, and Chua}{Sing et~al.}{2021}]{Sing_Kuo_Shih__Chua2021_ML_LPBF_review}
Sing, S.~L., C.~Kuo, C.~Shih, C.~Ho, and C.~K. Chua (2021).
\newblock Perspectives of using machine learning in laser powder bed fusion for metal additive manufacturing.
\newblock {\em Virtual and Physical Prototyping\/}~{\em 16\/}(3), 372--386.

\bibitem[\protect\citeauthoryear{Snow, Diehl, Reutzel, and Nassar}{Snow et~al.}{2021}]{Snow_Diehl__Nassar2021_in_situ_flaw_detection}
Snow, Z., B.~Diehl, E.~W. Reutzel, and A.~Nassar (2021).
\newblock Toward in-situ flaw detection in laser powder bed fusion additive manufacturing through layerwise imagery and machine learning.
\newblock {\em Journal of Manufacturing Systems\/}~{\em 59}, 12--26.

\bibitem[\protect\citeauthoryear{Snow, Reutzel, and Petrich}{Snow et~al.}{2022}]{Snow_Reutzel_Petrich2022_Correlating_monitoring_data_to_fatigue_performance}
Snow, Z., E.~W. Reutzel, and J.~Petrich (2022).
\newblock Correlating in-situ sensor data to defect locations and part quality for additively manufactured parts using machine learning.
\newblock {\em Journal of Materials Processing Technology\/}~{\em 302}, 117476.

\bibitem[\protect\citeauthoryear{Sofi and Ravani}{Sofi and Ravani}{2023}]{Sofi_Ravani2023}
Sofi, A.~R. and B.~Ravani (2023).
\newblock Sub-second prediction of the heatmap of powder-beds in additive manufacturing using deep encoder--decoder convolutional neural networks.
\newblock {\em Journal of Computing and Information Science in Engineering\/}~{\em 23\/}(2), 021008.

\bibitem[\protect\citeauthoryear{Song, Wang, Gao, Son, and Wu}{Song et~al.}{2023}]{Song_Wang_Gao_Son_Wu2023_pore_prediction}
Song, Z., X.~Wang, Y.~Gao, J.~Son, and J.~Wu (2023).
\newblock A hybrid deep generative network for pore morphology prediction in metal additive manufacturing.
\newblock {\em Journal of Manufacturing Science and Engineering\/}~{\em 145\/}(7), 071005.

\bibitem[\protect\citeauthoryear{Standfield, Gra{\v{c}}anin, Wang, and Kong}{Standfield et~al.}{2022}]{Standfield_Wang__Kong2022_shape_deformation_prediction}
Standfield, B., D.~Gra{\v{c}}anin, R.~Wang, and Z.~Kong (2022).
\newblock High-resolution shape deformation prediction in additive manufacturing using 3d cnn.
\newblock In {\em 2022 Winter Simulation Conference (WSC)}, pp.\  641--652. IEEE.

\bibitem[\protect\citeauthoryear{Surana, Lynch, Nassar, Ojard, Fisher, Corbin, and Overdorff}{Surana et~al.}{2023}]{Surana_Lynch_Nassar__Overdorff2023}
Surana, A., M.~E. Lynch, A.~R. Nassar, G.~C. Ojard, B.~A. Fisher, D.~Corbin, and R.~Overdorff (2023).
\newblock Flaw detection in multi-laser powder bed fusion using in situ coaxial multi-spectral sensing and deep learning.
\newblock {\em Journal of Manufacturing Science and Engineering\/}~{\em 145\/}(5), 051005.

\bibitem[\protect\citeauthoryear{Surovi and Soh}{Surovi and Soh}{2023}]{Surovi_Soh2023_monitoring_Defect_identification}
Surovi, N.~A. and G.~S. Soh (2023).
\newblock Acoustic feature based geometric defect identification in wire arc additive manufacturing.
\newblock {\em Virtual and Physical Prototyping\/}~{\em 18\/}(1), e2210553.

\bibitem[\protect\citeauthoryear{Szegedy, Liu, Jia, Sermanet, Reed, Anguelov, Erhan, Vanhoucke, and Rabinovich}{Szegedy et~al.}{2015}]{Szegedy_Liu_Jia__Rabinovich2015_GoogLeNet_Inception}
Szegedy, C., W.~Liu, Y.~Jia, P.~Sermanet, S.~Reed, D.~Anguelov, D.~Erhan, V.~Vanhoucke, and A.~Rabinovich (2015).
\newblock Going deeper with convolutions.
\newblock In {\em Proceedings of the IEEE conference on computer vision and pattern recognition}, pp.\  1--9.

\bibitem[\protect\citeauthoryear{Tan, Huang, Liu, Li, and Wu}{Tan et~al.}{2023}]{Tan_Huang_Liu_Li_Wu2022}
Tan, L., T.~Huang, J.~Liu, Q.~Li, and X.~Wu (2023).
\newblock Deep adversarial learning system for fault diagnosis in fused deposition modeling with imbalanced data.
\newblock {\em Computers \& Industrial Engineering\/}~{\em 176}, 108887.

\bibitem[\protect\citeauthoryear{Tan and Le}{Tan and Le}{2019}]{Tan_Le2019_Efficientnet_supp_paper}
Tan, M. and Q.~Le (2019).
\newblock Efficientnet: Rethinking model scaling for convolutional neural networks.
\newblock In {\em International conference on machine learning}, pp.\  6105--6114. PMLR.

\bibitem[\protect\citeauthoryear{Tan, Jin, Nettekoven, Chen, Yue, Topcu, and Sangiovanni-Vincentelli}{Tan et~al.}{2019}]{Jin_Tan__Sangiovanni-Vincentelli2019}
Tan, Y., B.~Jin, A.~Nettekoven, Y.~Chen, Y.~Yue, U.~Topcu, and A.~Sangiovanni-Vincentelli (2019).
\newblock An encoder-decoder based approach for anomaly detection with application in additive manufacturing.
\newblock In {\em 2019 18th IEEE international conference on machine learning and applications (ICMLA)}, pp.\  1008--1015. IEEE.

\bibitem[\protect\citeauthoryear{Tan, Fang, Li, Liu, Zhu, and Yang}{Tan et~al.}{2020}]{Tan_Fang_Li__Yang2020}
Tan, Z., Q.~Fang, H.~Li, S.~Liu, W.~Zhu, and D.~Yang (2020).
\newblock Neural network based image segmentation for spatter extraction during laser-based powder bed fusion processing.
\newblock {\em Optics \& Laser Technology\/}~{\em 130}, 106347.

\bibitem[\protect\citeauthoryear{Tang, Dehaghani, and Wang}{Tang et~al.}{2022}]{Tang_Dehaghani_Wang_2023_Review_transfer_learning}
Tang, Y., M.~R. Dehaghani, and G.~G. Wang (2022).
\newblock Review of transfer learning in modeling additive manufacturing processes.
\newblock {\em Additive Manufacturing\/}, 103357.

\bibitem[\protect\citeauthoryear{Tian, Guo, Guo, et~al.}{Tian et~al.}{2020}]{Guo_Tian_Guo_Guo2020}
Tian, Q., S.~Guo, Y.~Guo, et~al. (2020).
\newblock A physics-driven deep learning model for process-porosity causal relationship and porosity prediction with interpretability in laser metal deposition.
\newblock {\em CIRP Annals\/}~{\em 69\/}(1), 205--208.

\bibitem[\protect\citeauthoryear{Tian, Guo, Melder, Bian, and Guo}{Tian et~al.}{2021}]{Tian_Guo_Melder_Bian_Guo2021_data_fusion_porosity_detection}
Tian, Q., S.~Guo, E.~Melder, L.~Bian, and W.~â. Guo (2021).
\newblock Deep learning-based data fusion method for in situ porosity detection in laser-based additive manufacturing.
\newblock {\em Journal of Manufacturing Science and Engineering\/}~{\em 143\/}(4), 041011.

\bibitem[\protect\citeauthoryear{Tu, Liu, Carneiro, Ryan, Parnell, Leen, and Harrison}{Tu et~al.}{2022}]{Tu_Liu_Carneiro__Harrison2022}
Tu, Y., Z.~Liu, L.~Carneiro, C.~M. Ryan, A.~C. Parnell, S.~B. Leen, and N.~M. Harrison (2022).
\newblock Towards an instant structure-property prediction quality control tool for additive manufactured steel using a crystal plasticity trained deep learning surrogate.
\newblock {\em Materials \& Design\/}~{\em 213}, 110345.

\bibitem[\protect\citeauthoryear{Valizadeh and Wolff}{Valizadeh and Wolff}{2022}]{Valizadeh_Wolff2022}
Valizadeh, M. and S.~J. Wolff (2022).
\newblock Convolutional neural network applications in additive manufacturing: A review.
\newblock {\em Advances in Industrial and Manufacturing Engineering\/}, 100072.

\bibitem[\protect\citeauthoryear{Voigt and Moeckel}{Voigt and Moeckel}{2022}]{Voigt_Moeckel2022}
Voigt, J. and M.~Moeckel (2022).
\newblock Benchmarking a multi-layer approach and neural network architectures for defect detection in pbf-lb/m.
\newblock {\em Materials Today Communications\/}~{\em 33}, 104878.

\bibitem[\protect\citeauthoryear{Wang, Chandra, Huang, Tor, and Tan}{Wang et~al.}{2023}]{Wang_Chandra_Huang_Tor_Tan2023}
Wang, C., S.~Chandra, S.~Huang, S.~B. Tor, and X.~Tan (2023).
\newblock Unraveling process-microstructure-property correlations in powder-bed fusion additive manufacturing through information-rich surface features with deep learning.
\newblock {\em Journal of Materials Processing Technology\/}~{\em 311}, 117804.

\bibitem[\protect\citeauthoryear{Wang, Tan, Tor, and Lim}{Wang et~al.}{2020}]{Wang_Tan_Tor_Lim_2020_review}
Wang, C., X.~Tan, S.~B. Tor, and C.~Lim (2020).
\newblock Machine learning in additive manufacturing: State-of-the-art and perspectives.
\newblock {\em Additive Manufacturing\/}~{\em 36}, 101538.

\bibitem[\protect\citeauthoryear{Wang, Shraida, and Yu}{Wang et~al.}{2023}]{Wang_Shraida_Jin2023predictive_cloud}
Wang, H., H.~A.~A. Shraida, and J.~Yu (2023).
\newblock Predictive modeling of out-of-plane deviation for the quality improvement of additive manufacturing.
\newblock In {\em Materials Science Forum}, Volume 1086, pp.\  79--83. Trans Tech Publ.

\bibitem[\protect\citeauthoryear{Wang and Cheung}{Wang and Cheung}{2022}]{Wang_Cheung2022}
Wang, R. and C.~F. Cheung (2022).
\newblock Centernet-based defect detection for additive manufacturing.
\newblock {\em Expert Systems with Applications\/}~{\em 188}, 116000.

\bibitem[\protect\citeauthoryear{Wang, Cheung, and Wang}{Wang et~al.}{2023}]{Wang_Cheung_wang2023_surface_Defect_Segmentation_SLM_CNN_with_attention}
Wang, R., C.~F. Cheung, and C.~Wang (2023).
\newblock Unsupervised defect segmentation in selective laser melting.
\newblock {\em IEEE Transactions on Instrumentation and Measurement\/}.

\bibitem[\protect\citeauthoryear{Wang, Wang, Zhang, Chen, Wang, Lu, Chen, Liu, and Li}{Wang et~al.}{2023}]{Wang_Wang__Li2023_suface_Defect_Detection_AM_YOLOv8}
Wang, W., P.~Wang, H.~Zhang, X.~Chen, G.~Wang, Y.~Lu, M.~Chen, H.~Liu, and J.~Li (2023).
\newblock A real-time defect detection strategy for additive manufacturing processes based on deep learning and machine vision technologies.
\newblock {\em Micromachines\/}~{\em 15\/}(1), 28.

\bibitem[\protect\citeauthoryear{Wang, Hu, Li, and Wang}{Wang et~al.}{2023}]{Wang_Hu_Li_Wang2023_meltpool_width_layerHeight_prediction}
Wang, Y., K.~Hu, W.~Li, and L.~Wang (2023).
\newblock Prediction of melt pool width and layer height for laser directed energy deposition enabled by physics-driven temporal convolutional network.
\newblock {\em Journal of Manufacturing Systems\/}~{\em 69}, 1--17.

\bibitem[\protect\citeauthoryear{Wang, Lu, Zhao, Deng, Han, Bai, Yang, and Yao}{Wang et~al.}{2021}]{Wang_Lu_Zhao_Deng__Yao2021}
Wang, Y., J.~Lu, Z.~Zhao, W.~Deng, J.~Han, L.~Bai, X.~Yang, and J.~Yao (2021).
\newblock Active disturbance rejection control of layer width in wire arc additive manufacturing based on deep learning.
\newblock {\em Journal of Manufacturing Processes\/}~{\em 67}, 364--375.

\bibitem[\protect\citeauthoryear{Wang, Sun, Jin, Kong, and Yue}{Wang et~al.}{2023}]{Wang_Sun_Jin_Kong_Yue2022_MVGCN_defect_identification}
Wang, Y., W.~Sun, J.~Jin, Z.~Kong, and X.~Yue (2023).
\newblock Mvgcn: Multi-view graph convolutional neural network for surface defect identification using three-dimensional point cloud.
\newblock {\em Journal of Manufacturing Science and Engineering\/}~{\em 145\/}(3), 031004.

\bibitem[\protect\citeauthoryear{Wang, Sun, Liu, Sarma, Bronstein, and Solomon}{Wang et~al.}{2019}]{Wang_Sun__Solomon2019_DGCNN_ref}
Wang, Y., Y.~Sun, Z.~Liu, S.~E. Sarma, M.~M. Bronstein, and J.~M. Solomon (2019).
\newblock Dynamic graph cnn for learning on point clouds.
\newblock {\em ACM Transactions on Graphics (tog)\/}~{\em 38\/}(5), 1--12.

\bibitem[\protect\citeauthoryear{Wang, Xu, Zhao, Deng, Han, Bai, Liang, and Yao}{Wang et~al.}{2021}]{Wang_Xu__Yao2021_deposited_layer_width_and_reinforcement_prediction_WAAM}
Wang, Y., X.~Xu, Z.~Zhao, W.~Deng, J.~Han, L.~Bai, X.~Liang, and J.~Yao (2021).
\newblock Coordinated monitoring and control method of deposited layer width and reinforcement in waam process.
\newblock {\em Journal of Manufacturing Processes\/}~{\em 71}, 306--316.

\bibitem[\protect\citeauthoryear{Wang, Zhang, Lu, Bai, Zhao, and Han}{Wang et~al.}{2020}]{Wang_Zhang__Han2020_Weld_reinforcement_prediction_molten_pool_image_WAAM_LSTM}
Wang, Y., C.~Zhang, J.~Lu, L.~Bai, Z.~Zhao, and J.~Han (2020).
\newblock Weld reinforcement analysis based on long-term prediction of molten pool image in additive manufacturing.
\newblock {\em IEEE Access\/}~{\em 8}, 69908--69918.

\bibitem[\protect\citeauthoryear{Wang, Yang, Liu, Zhao, Liu, Wu, Banu, and Chen}{Wang et~al.}{2022}]{Wang_Yang_Liu__Chen2022_Data_driven_AM_modeling_review}
Wang, Z., W.~Yang, Q.~Liu, Y.~Zhao, P.~Liu, D.~Wu, M.~Banu, and L.~Chen (2022).
\newblock Data-driven modeling of process, structure and property in additive manufacturing: A review and future directions.
\newblock {\em Journal of Manufacturing Processes\/}~{\em 77}, 13--31.

\bibitem[\protect\citeauthoryear{Westphal and Seitz}{Westphal and Seitz}{2021}]{Westphal_Seitz2021}
Westphal, E. and H.~Seitz (2021).
\newblock A machine learning method for defect detection and visualization in selective laser sintering based on convolutional neural networks.
\newblock {\em Additive Manufacturing\/}~{\em 41}, 101965.

\bibitem[\protect\citeauthoryear{Williams, Meisel, Simpson, and McComb}{Williams et~al.}{2019}]{Williams_Meisel_Simpson_McComb2019design_repository_manufacturability}
Williams, G., N.~A. Meisel, T.~W. Simpson, and C.~McComb (2019).
\newblock Design repository effectiveness for 3d convolutional neural networks: Application to additive manufacturing.
\newblock {\em Journal of Mechanical Design\/}~{\em 141\/}(11).

\bibitem[\protect\citeauthoryear{Williams, Dryburgh, Clare, Rao, and Samal}{Williams et~al.}{2018}]{Williams_Dryburgh_Clare_Rao_Samal2018}
Williams, J., P.~Dryburgh, A.~Clare, P.~Rao, and A.~Samal (2018).
\newblock Defect detection and monitoring in metal additive manufactured parts through deep learning of spatially resolved acoustic spectroscopy signals.
\newblock {\em Smart and Sustainable Manufacturing Systems\/}~{\em 2\/}(1).

\bibitem[\protect\citeauthoryear{Williams and Sing}{Williams and Sing}{2024}]{Williams_Sing2024_Spatiotemporal_PBF_meltpool_monitoring_videos_convRNN}
Williams, R.~J. and S.~L. Sing (2024).
\newblock Spatiotemporal analysis of powder bed fusion melt pool monitoring videos using deep learning.
\newblock {\em Journal of Intelligent Manufacturing\/}, 1--14.

\bibitem[\protect\citeauthoryear{Wright, Darville, Celik, Koerner, and Celik}{Wright et~al.}{2022}]{Wright_Darville_Celik__Celik2022}
Wright, W.~J., J.~Darville, N.~Celik, H.~Koerner, and E.~Celik (2022).
\newblock In-situ optimization of thermoset composite additive manufacturing via deep learning and computer vision.
\newblock {\em Additive Manufacturing\/}~{\em 58}, 102985.

\bibitem[\protect\citeauthoryear{Xames, Torsha, and Sarwar}{Xames et~al.}{2022}]{Xames_Torsha_Sarwar2022_review}
Xames, M.~D., F.~K. Torsha, and F.~Sarwar (2022).
\newblock A systematic literature review on recent trends of machine learning applications in additive manufacturing.
\newblock {\em Journal of Intelligent Manufacturing\/}, 1--27.

\bibitem[\protect\citeauthoryear{Xia, Pan, Li, Chen, and Li}{Xia et~al.}{2022}]{Xia_Pan_Li_Chen_Li2022}
Xia, C., Z.~Pan, Y.~Li, J.~Chen, and H.~Li (2022).
\newblock Vision-based melt pool monitoring for wire-arc additive manufacturing using deep learning method.
\newblock {\em The International Journal of Advanced Manufacturing Technology\/}~{\em 120\/}(1-2), 551--562.

\bibitem[\protect\citeauthoryear{Xie, Jiang, and Chen}{Xie et~al.}{2022}]{Xie_Jiang_Chen2022}
Xie, J., T.~Jiang, and X.~Chen (2022).
\newblock An image segmentation framework for in-situ monitoring in laser powder bed fusion additive manufacturing.
\newblock {\em IFAC-PapersOnLine\/}~{\em 55\/}(37), 800--806.

\bibitem[\protect\citeauthoryear{Yang, Wang, Li, Qi, Wang, Lei, and Fang}{Yang et~al.}{2022}]{Yang_Wang__Fang2022_UNet_XCT_image_reconstruction_prediction_compression_lattice_structures}
Yang, H., W.~Wang, C.~Li, J.~Qi, P.~Wang, H.~Lei, and D.~Fang (2022).
\newblock Deep learning-based x-ray computed tomography image reconstruction and prediction of compression behavior of 3d printed lattice structures.
\newblock {\em Additive Manufacturing\/}~{\em 54}, 102774.

\bibitem[\protect\citeauthoryear{Yang, Qiu, Liu, Qiu, and Bai}{Yang et~al.}{2023}]{Yang_Qiu__Bai2023_Defect_classification_LPBF_simulated_meltpool_images_and_thermal_images_CNN}
Yang, W., Y.~Qiu, W.~Liu, X.~Qiu, and Q.~Bai (2023).
\newblock Defect prediction in laser powder bed fusion with the combination of simulated melt pool images and thermal images.
\newblock {\em Journal of Manufacturing Processes\/}~{\em 106}, 214--222.

\bibitem[\protect\citeauthoryear{Yang and Kan}{Yang and Kan}{2023}]{Yang_Kan2023_3D_Geometry_Representation_AM}
Yang, Y. and C.~Kan (2023).
\newblock Recurrence network-based 3d geometry representation learning for quality control in additive manufacturing of metamaterials.
\newblock {\em Journal of Manufacturing Science and Engineering\/}~{\em 145\/}(11), 111006.

\bibitem[\protect\citeauthoryear{Yangue, Ye, Kan, and Liu}{Yangue et~al.}{2023}]{Yangue2023_online_surface_prediction}
Yangue, E., Z.~Ye, C.~Kan, and C.~Liu (2023).
\newblock Integrated deep learning-based online layer-wise surface prediction of additive manufacturing.
\newblock {\em Manufacturing Letters\/}~{\em 35}, 760--769.

\bibitem[\protect\citeauthoryear{Ye, Fuh, Zhang, Hong, and Zhu}{Ye et~al.}{2018}]{Ye_Fuh_Zhang_Hong_Zhu2018}
Ye, D., J.~Y.~H. Fuh, Y.~Zhang, G.~S. Hong, and K.~Zhu (2018).
\newblock In situ monitoring of selective laser melting using plume and spatter signatures by deep belief networks.
\newblock {\em ISA transactions\/}~{\em 81}, 96--104.

\bibitem[\protect\citeauthoryear{Ye, Hong, Zhang, Zhu, and Fuh}{Ye et~al.}{2018}]{Ye_Hong_Zhang_Zhu_Fuh2018}
Ye, D., G.~S. Hong, Y.~Zhang, K.~Zhu, and J.~Y.~H. Fuh (2018).
\newblock Defect detection in selective laser melting technology by acoustic signals with deep belief networks.
\newblock {\em The International Journal of Advanced Manufacturing Technology\/}~{\em 96}, 2791--2801.

\bibitem[\protect\citeauthoryear{Ye, Liu, Tian, and Kan}{Ye et~al.}{2020}]{Ye_Liu__Tian_Kan2020_monitoring_point_cloud}
Ye, Z., C.~Liu, W.~Tian, and C.~Kan (2020).
\newblock A deep learning approach for the identification of small process shifts in additive manufacturing using 3d point clouds.
\newblock {\em Procedia Manufacturing\/}~{\em 48}, 770--775.

\bibitem[\protect\citeauthoryear{Ye, Liu, Tian, and Kan}{Ye et~al.}{2021}]{Ye_Liu__Kan2021_point_cloud_fusion}
Ye, Z., C.~Liu, W.~Tian, and C.~Kan (2021).
\newblock In-situ point cloud fusion for layer-wise monitoring of additive manufacturing.
\newblock {\em Journal of Manufacturing Systems\/}~{\em 61}, 210--222.

\bibitem[\protect\citeauthoryear{Yuan, Giera, Guss, Matthews, and Mcmains}{Yuan et~al.}{2019}]{Yuan_Giera_Guss__Mcmains2019}
Yuan, B., B.~Giera, G.~Guss, I.~Matthews, and S.~Mcmains (2019).
\newblock Semi-supervised convolutional neural networks for in-situ video monitoring of selective laser melting.
\newblock In {\em 2019 IEEE winter conference on applications of computer vision (WACV)}, pp.\  744--753. IEEE.

\bibitem[\protect\citeauthoryear{Yue, Chen, Li, Li, Zhu, and Yin}{Yue et~al.}{2023}]{Yue_Chen_LI_Yin2023_control_droplet_volume_inkjet_printing_DRL}
Yue, X., J.~Chen, Y.~Li, X.~Li, H.~Zhu, and Z.~Yin (2023).
\newblock Intelligent control system for droplet volume in inkjet printing based on stochastic state transition soft actor--critic drl algorithm.
\newblock {\em Journal of Manufacturing Systems\/}~{\em 68}, 455--464.

\bibitem[\protect\citeauthoryear{Zamiela, Jiang, Stokes, Tian, Netchaev, Dickerson, Tian, and Bian}{Zamiela et~al.}{2023}]{Zamiela_Jiang__Bian2023_porosity_detection}
Zamiela, C., Z.~Jiang, R.~Stokes, Z.~Tian, A.~Netchaev, C.~Dickerson, W.~Tian, and L.~Bian (2023).
\newblock Deep multi-modal u-net fusion methodology of thermal and ultrasonic images for porosity detection in additive manufacturing.
\newblock {\em Journal of Manufacturing Science and Engineering\/}~{\em 145\/}(6), 061009.

\bibitem[\protect\citeauthoryear{Zhang, Liu, and Shin}{Zhang et~al.}{2019}]{Zhang_Liu_Shin2019}
Zhang, B., S.~Liu, and Y.~C. Shin (2019).
\newblock In-process monitoring of porosity during laser additive manufacturing process.
\newblock {\em Additive Manufacturing\/}~{\em 28}, 497--505.

\bibitem[\protect\citeauthoryear{Zhang, Vallabh, and Zhao}{Zhang et~al.}{2022}]{Zhang_Vallabh__Zhao2022_meltpool_temperature_and_morphology_monitoring_LSTM_LPBF}
Zhang, H., C.~K.~P. Vallabh, and X.~Zhao (2022).
\newblock Registration and fusion of large-scale melt pool temperature and morphology monitoring data demonstrated for surface topography prediction in lpbf.
\newblock {\em Additive Manufacturing\/}~{\em 58}, 103075.

\bibitem[\protect\citeauthoryear{Zhang, Wang, and Gao}{Zhang et~al.}{2018}]{Zhang_Wang_Gao2018}
Zhang, J., P.~Wang, and R.~X. Gao (2018).
\newblock Modeling of layer-wise additive manufacturing for part quality prediction.
\newblock {\em Procedia Manufacturing\/}~{\em 16}, 155--162.

\bibitem[\protect\citeauthoryear{Zhang, Wang, and Gao}{Zhang et~al.}{2019}]{Zhang_Wang_Gao2019}
Zhang, J., P.~Wang, and R.~X. Gao (2019).
\newblock Deep learning-based tensile strength prediction in fused deposition modeling.
\newblock {\em Computers in industry\/}~{\em 107}, 11--21.

\bibitem[\protect\citeauthoryear{Zhang, Fu, Jahn, Collet, and Schleifenbaum}{Zhang et~al.}{2023}]{Zhang_Fu__Schleifenbaum2023_image_quality_enhancement_Unet_LPBF}
Zhang, S., T.~Fu, A.~Jahn, A.~Collet, and J.~H. Schleifenbaum (2023).
\newblock Towards deep-learning-based image enhancement for optical camera-based monitoring system of laser powder bed fusion process.
\newblock {\em International Journal of Computer Integrated Manufacturing\/}~{\em 36\/}(9), 1281--1294.

\bibitem[\protect\citeauthoryear{Zhang, Xu, Cheng, Chen, Wang, and Wang}{Zhang et~al.}{2023}]{Zhang_Xu__Wang2023_WAAM_detect_surface_oxidation_defects_Transformer_time_series_voltage_data}
Zhang, T., C.~Xu, J.~Cheng, Z.~Chen, L.~Wang, and K.~Wang (2023).
\newblock Research of surface oxidation defects in copper alloy wire arc additive manufacturing based on time-frequency analysis and deep learning method.
\newblock {\em Journal of Materials Research and Technology\/}~{\em 25}, 511--521.

\bibitem[\protect\citeauthoryear{Zhang, Safdar, Xie, Li, Sage, and Zhao}{Zhang et~al.}{2022}]{Zhang_Safdar_Xie_Li_Sage_Zhao2022_review}
Zhang, Y., M.~Safdar, J.~Xie, J.~Li, M.~Sage, and Y.~F. Zhao (2022).
\newblock A systematic review on data of additive manufacturing for machine learning applications: the data quality, type, preprocessing, and management.
\newblock {\em Journal of Intelligent Manufacturing\/}, 1--36.

\bibitem[\protect\citeauthoryear{Zhang, Yang, Dong, and Zhao}{Zhang et~al.}{2021}]{Zhang_Yang_Dong_Zhao2021_predictive_manufacturability}
Zhang, Y., S.~Yang, G.~Dong, and Y.~F. Zhao (2021).
\newblock Predictive manufacturability assessment system for laser powder bed fusion based on a hybrid machine learning model.
\newblock {\em Additive Manufacturing\/}~{\em 41}, 101946.

\bibitem[\protect\citeauthoryear{Zhang and Zhao}{Zhang and Zhao}{2022}]{Zhang_Zhao2022}
Zhang, Y. and Y.~F. Zhao (2022).
\newblock Hybrid sparse convolutional neural networks for predicting manufacturability of visual defects of laser powder bed fusion processes.
\newblock {\em Journal of Manufacturing Systems\/}~{\em 62}, 835--845.

\bibitem[\protect\citeauthoryear{Zhang, Sahu, Singh, Rai, Yang, and Lu}{Zhang et~al.}{2023}]{Zhang_Sahu_Singh__Lu2023_meltpool_morphology_prediction}
Zhang, Z., C.~K. Sahu, S.~K. Singh, R.~Rai, Z.~Yang, and Y.~Lu (2023).
\newblock Machine learning based prediction of melt pool morphology in a laser-based powder bed fusion additive manufacturing process.
\newblock {\em International Journal of Production Research\/}, 1--15.

\bibitem[\protect\citeauthoryear{Zhang, Wen, and Chen}{Zhang et~al.}{2019}]{Zhang_Wen_Chen2019_Weld_image_defect_detection_CNN_Robotic_arc_welding}
Zhang, Z., G.~Wen, and S.~Chen (2019).
\newblock Weld image deep learning-based on-line defects detection using convolutional neural networks for al alloy in robotic arc welding.
\newblock {\em Journal of Manufacturing Processes\/}~{\em 45}, 208--216.

\bibitem[\protect\citeauthoryear{Zhao, Jiang, Jia, Torr, and Koltun}{Zhao et~al.}{2021}]{Zhao2021_Point_Transformer_ref_paper}
Zhao, H., L.~Jiang, J.~Jia, P.~H. Torr, and V.~Koltun (2021).
\newblock Point transformer.
\newblock In {\em Proceedings of the IEEE/CVF international conference on computer vision}, pp.\  16259--16268.

\bibitem[\protect\citeauthoryear{Zhao, Wei, Mao, Zhang, Liu, and Liao}{Zhao et~al.}{2023}]{Zhao_Wei_Mao__Liao2023_prediction_params_melt_pool_dimensions_PIDL}
Zhao, M., H.~Wei, Y.~Mao, C.~Zhang, T.~Liu, and W.~Liao (2023).
\newblock Predictions of additive manufacturing process parameters and molten pool dimensions with a physics-informed deep learning model.
\newblock {\em Engineering\/}~{\em 23}, 181--195.

\bibitem[\protect\citeauthoryear{Zhao, Xiong, Shang, Liu, Shen, and Wu}{Zhao et~al.}{2019}]{Zhao_Xiong_Shang__Wu2019nonlinear_error_prediction}
Zhao, M., G.~Xiong, X.~Shang, C.~Liu, Z.~Shen, and H.~Wu (2019).
\newblock Nonlinear deformation prediction and compensation for 3d printing based on cae neural networks.
\newblock In {\em 2019 IEEE 15th International Conference on Automation Science and Engineering (CASE)}, pp.\  667--672. IEEE.

\bibitem[\protect\citeauthoryear{Zhao, Xiong, Wang, Fang, Shen, Wan, and Zhu}{Zhao et~al.}{2022}]{Zhao_Xiong___Zhu2022point_deformation_prediction}
Zhao, M., G.~Xiong, W.~Wang, Q.~Fang, Z.~Shen, L.~Wan, and F.~Zhu (2022).
\newblock A point-based neural network for real-scenario deformation prediction in additive manufacturing.
\newblock In {\em 2022 IEEE 18th International Conference on Automation Science and Engineering (CASE)}, pp.\  1656--1661. IEEE.

\bibitem[\protect\citeauthoryear{Zhou, Shen, Lin, Liu, and Sheng}{Zhou et~al.}{2022}]{Zhou_Shen__Sheng2022_tool-path_planning_optimizing_thermo-mechanical_properties_WAAM_DNN_RNN_LSTM}
Zhou, Z., H.~Shen, J.~Lin, B.~Liu, and X.~Sheng (2022).
\newblock Continuous tool-path planning for optimizing thermo-mechanical properties in wire-arc additive manufacturing: An evolutional method.
\newblock {\em Journal of Manufacturing Processes\/}~{\em 83}, 354--373.

\bibitem[\protect\citeauthoryear{Zhou, Shen, Liu, Du, and Jin}{Zhou et~al.}{2021}]{Zhou_Shen_Liu_Du_Jin2021_Thermal_field_prediction}
Zhou, Z., H.~Shen, B.~Liu, W.~Du, and J.~Jin (2021).
\newblock Thermal field prediction for welding paths in multi-layer gas metal arc welding-based additive manufacturing: A machine learning approach.
\newblock {\em Journal of Manufacturing Processes\/}~{\em 64}, 960--971.

\bibitem[\protect\citeauthoryear{Zhu, Liu, and Yan}{Zhu et~al.}{2021}]{Zhu_Liu_Yan2021}
Zhu, Q., Z.~Liu, and J.~Yan (2021).
\newblock Machine learning for metal additive manufacturing: predicting temperature and melt pool fluid dynamics using physics-informed neural networks.
\newblock {\em Computational Mechanics\/}~{\em 67}, 619--635.

\bibitem[\protect\citeauthoryear{Zhu, Jiang, Guo, Xu, Wang, and Jiang}{Zhu et~al.}{2023}]{Zhu_Jiang_Guo_Xu__Jiang2023_surface_morphology_inspection_TL}
Zhu, X., F.~Jiang, C.~Guo, D.~Xu, Z.~Wang, and G.~Jiang (2023).
\newblock Surface morphology inspection for directed energy deposition using small dataset with transfer learning.
\newblock {\em Journal of Manufacturing Processes\/}~{\em 93}, 101--115.

\bibitem[\protect\citeauthoryear{Zhu, Ferreira, Anwer, Mathieu, Guo, and Qiao}{Zhu et~al.}{2020}]{Zhu_Ferreira_Anwar__Qiao2020}
Zhu, Z., K.~Ferreira, N.~Anwer, L.~Mathieu, K.~Guo, and L.~Qiao (2020).
\newblock Convolutional neural network for geometric deviation prediction in additive manufacturing.
\newblock {\em Procedia Cirp\/}~{\em 91}, 534--539.

\bibitem[\protect\citeauthoryear{Ziabari, Venkatakrishnan, Snow, Lisovich, Sprayberry, Brackman, Frederick, Bhattad, Graham, Bingham, et~al.}{Ziabari et~al.}{2023}]{Ziabari_Venkatakrishnan__Paquit2023_part_reconstruction_from_XCT_images_MAM_CNN}
Ziabari, A., S.~V. Venkatakrishnan, Z.~Snow, A.~Lisovich, M.~Sprayberry, P.~Brackman, C.~Frederick, P.~Bhattad, S.~Graham, P.~Bingham, et~al. (2023).
\newblock Enabling rapid x-ray ct characterisation for additive manufacturing using cad models and deep learning-based reconstruction.
\newblock {\em npj Computational Materials\/}~{\em 9\/}(1), 91.

\end{thebibliography}
	
\end{document}